\documentclass{article}
\usepackage[margin=1in]{geometry}

\usepackage{amsmath,amsfonts,bm}

\def\eqref#1{equation~\ref{#1}}
\def\1{\bm{1}}

\DeclareMathAlphabet{\mathsfit}{\encodingdefault}{\sfdefault}{m}{sl}
\SetMathAlphabet{\mathsfit}{bold}{\encodingdefault}{\sfdefault}{bx}{n}

\newcommand{\E}{\mathbb{E}}

\newcommand{\R}{\mathbb{R}}

\DeclareMathOperator*{\argmin}{arg\,min}

\usepackage[T1]{fontenc}    % use 8-bit T1 fonts
\usepackage{hyperref}       % hyperlinks
\usepackage{url}            % simple URL typesetting
\usepackage{booktabs}       % professional-quality tables
\usepackage{amsfonts}       % blackboard math symbols
\usepackage{nicefrac}       % compact symbols for 1/2, etc.
\usepackage{microtype}      % microtypography
\usepackage{graphicx}
\usepackage{subcaption}
\usepackage{booktabs} % for professional tables
\usepackage{tikz}

\usepackage[sort, numbers]{natbib}

\usepackage{tikz}
\usetikzlibrary{shapes,decorations}

\usepackage{amssymb}
\usepackage{nicefrac}       % compact symbols for 1/2, etc.
\usepackage{amsfonts}       % blackboard math symbols
\usepackage{amsmath}
\usepackage{amsthm}

\usepackage{mathtools}

\usepackage{siunitx}
\newcommand{\abs}[1]{\left\lvert#1\right\rvert}

\DeclareMathOperator{\Fc}{\mathcal{F}}

\DeclareMathOperator{\Lc}{\mathcal{L}}

\usepackage{mathtools}

\newtheorem{theorem}{Theorem}
\newtheorem{corollary}{Corollary}[theorem]

\theoremstyle{definition}

\usepackage{authblk}

\title{Dropout as a Regularizer of Interaction Effects}

\author[1]{Ben Lengerich\thanks{\texttt{blengeri@cs.cmu.edu}}}
\author[1,2]{Eric P. Xing\thanks{\texttt{epxing@cs.cmu.edu}}}
\author[3]{Rich Caruana\thanks{\texttt{rcaruana@microsoft.com}}}
\affil[1]{Carnegie Mellon University}
\affil[2]{Petuum, Inc.}
\affil[3]{Microsoft Research}

\begin{document}

\maketitle

\begin{abstract}
We examine Dropout through the perspective of \emph{interactions}. This view provides a symmetry to explain Dropout: given $N$ variables, there are ${N \choose k}$ possible sets of $k$ variables to form an interaction (i.e. $\mathcal{O}(N^k)$); conversely, the probability an interaction of $k$ variables survives Dropout at rate $p$ is $(1-p)^k$ (decaying with $k$). These rates effectively cancel, and so Dropout regularizes against higher-order interactions. We prove this perspective analytically and empirically. This perspective of Dropout as a regularizer against interaction effects has several practical implications: (1) higher Dropout rates should be used when we need stronger regularization against spurious high-order interactions, (2) caution should be exercised when interpreting Dropout-based explanations and uncertainty measures, and (3) networks trained with Input Dropout are biased estimators. We also compare Dropout to other regularizers and find that it is difficult to obtain the same selective pressure against high-order interactions.
\end{abstract}

\section{Introduction}
\label{sec:introduction}

We examine Dropout through the perspective of \emph{interactions}: effects that require multiple variables. 
Given $N$ variables, there are ${N \choose k}$ possible sets of $k$ variables ($N$ univariate effects, $\mathcal{O}(N^2)$ pairwise interactions, $\mathcal{O}(N^3)$ 3-way interactions); we can thus imagine that models with large representational capacity could be dominated by high-order interactions.
In this paper, we show that Dropout contributes a regularization effect which helps neural networks (NNs) explore functions of lower-order interactions before considering functions of higher-order interactions. 
Dropout imposes this regularization by reducing the effective learning rate of higher-order interactions. % according to the number of variables in the interaction. 
As a result, Dropout encourages models to learn lower-order functions of additive components. 
This understanding of Dropout has implications for choosing Dropout rates: higher Dropout rates should be used when we need stronger regularization against interactions. 
This perspective also issues caution against using Dropout to measure term salience because Dropout regularizes against high-order interactions. 
Finally, this view of Dropout as a regularizer of interactions provides insight into the varying effectiveness of Dropout across architectures and datasets. 
We also compare Dropout to weight decay and early stopping and find that it is difficult to obtain the same regularization with these alternatives.

\paragraph{Why Interaction Effects?}
Dropout was originally motivated to prevent ``complex co-adaptations in which a feature detector is only helpful in the context of several other specific feature detectors" \citep{hinton2012improving, JMLR:v15:srivastava14a}. Most "complex co-adaptations" are interaction effects, so it is natural to quantify the effect of Dropout through the lens of interaction effects. 
This perspective is valuable because (1) %modern NNs have so many weights that understanding networks by looking at their weights is infeasible, but interactions are far more tractable because interaction effects live in function space, not weight space, 
modern NNs, containing intractable numbers of parameters, are more suitable to analysis via nonparametric functional analysis than parametric analysis,
(2) interaction effects can be calculated identifiably, and (3) this perspective has practical implications on choosing Dropout rates. 
To preview the results, when NNs are trained on data without important interactions, the optimal Dropout rate is high, but when NNs are trained on data with important interaction effects, the optimal Dropout rate is lower.

\section{Preliminaries and Related Work}
\label{sec:prelim}

\subsection{Functional ANOVA and Pure Interaction Effects}
%\label{sec:prelim:intx}

In this paper, we use the concept of \emph{pure interaction effects} from \cite{lengerich2019purifying}: a pure interaction effect is variance explained by a group of variables $u$ that \emph{cannot} be explained by any group of variables $u'$ where $\abs{u'} < \abs{u}$ (e.g. any subset of $u$). 
Multiplicative terms like $X_1X_2$ are often used to encode ``interaction effects". They are, however, only \emph{pure} interaction effects if $X_1$ and $X_2$ are uncorrelated and have mean zero; otherwise, some portion of the variance in the outcome $X_1X_2$ could be explained by main effects of each individual variable. %(e.g. if $X_1$ and $X_2$ are perfectly correlated, $X_1X_2 = X_1^2$). 
Correlation between two variables does not imply an interaction effect on the outcome, and an interaction effect of two variables on the outcome does not imply correlation between the variables. 

This definition of pure interaction effects is equivalent to the functional ANOVA (fANOVA) decomposition: %of $F$: 
Given a density $w(X)$ and $\Fc^u \subset \Lc^2(\R^u)$ the family of allowable functions for variable set $u$, the weighted fANOVA \citep{hooker2004diagnostics,hooker2007generalized,cuevas2004anova} decomposition of $F(X)$ is:
\begin{subequations}
\begin{align}
    \{f_u(X_u)|u  \subseteq [d] \} = 
    \argmin_{\{g_u \in \Fc^u\}_{u\in [d]}} \int \Big(\sum_{u \subseteq [d]}g_u(X_u) - F(X) \Big)^2w(X)dX, %\nonumber
\end{align}
where $[d]$ indicates the power set of $d$ features, such that
\begin{equation}
    \forall~v \subseteq~u,\quad \int f_u(X_u)g_v(X_v)w(X)dX = 0 \quad~\forall~g_v,
    \label{eq:orthogonality}
\end{equation}
\end{subequations}
i.e., each member $f_u$ is orthogonal to the members which operate on any subset of $u$.\footnote{
The fANOVA decomposition describes a unique decomposition for a given data distribution; thus, pure interaction effects are defined in conjunction with a data distribution. 
An example of this interplay between is shown in Figure~\ref{fig:toy_intx}. 
As \cite{lengerich2019purifying} describe, the correct distribution to use is the data-generating distribution $p(x)$. 
Estimating $p(x)$ is one of the central challenges of machine learning; for this paper, we mainly use simulation data for which we know $p(x)$.% and can precisely study the effects of Dropout. 
}
%Given this decomposition, we have a set of functions $f_u$ which can be analyzed individually. 
An interaction effect $f_u$ is of \emph{order} $k$ if $|u| = k$. 
Given $N$ variables, there are ${N \choose k}$ possible sets of size $k$, so we say that there are ${N \choose k}$ interaction effects of order $k$. %$$\mathcal{O}(N)$ possible effects of individual variables, $O(N^2)$ possible pairwise interactions, $\mathcal{O}(N^3)$ possible 3-way interactions, i.e. $\mathcal{O}(N^k)$ possible interactions of order $k$.

\subsection{Related Work}
\label{sec:related}
\citeauthor{hinton2012improving} proposed Dropout to prevent spurious co-adaptation (i.e., spurious interactions), and it has proved an extremely effective regularizer of deep models. 
However, questions remain.
For example: Is the expectation of the output of a NN trained with Dropout the same as for a NN trained without Dropout? Does Dropout change the trajectory of learning during optimization even in the asymptotic limit of infinite training data? Should Dropout be used at run-time when querying a NN to see what it has learned?
These questions are important because Dropout has been used as a method for Bayesian uncertainty \citep{gal2016dropout,NIPS2017_6949,chang2017dropout,chang2017interpreting}. % which implicitly assume that Dropout does not bias the model's output. 
The use of Dropout for uncertainty quantification has been questioned due to its failure to separate aleotoric and epistemic sources of uncertainty \citep{osband2016risk} (i.e., the uncertainty does not decrease even as more data is gathered). 
In this paper we ask a separate yet related question: Does Dropout treat functions equivalently?

Significant work has focused on the effect of Dropout as a weight regularizer \citep{baldi2013understanding,warde2013empirical,cavazza2018dropout,mianjy2018implicit,zunino2018excitation}, including its properties of structured shrinkage \citep{nalisnick2018dropout} or adaptive regularization \citep{wager2013dropout}. 
However, weight regularization is of limited utility for modern-scale NNs. 
Instead of focusing on the influence of Dropout on parameters, we take a nonparametric view of NNs as function approximators. %and query the input-output change. 
Thus, our work is similar in spirit to \cite{pmlr-v28-wan13}, which showed a linear relationship between keep probability and Rademacher complexity, and \cite{duvenaud2014avoiding}, which showed that Dropout can be viewed as a mixture of models which each depend on only a subset of the input variables. 
Our work crystallizes these observations into a description of Dropoutas a regularizer of interaction effects, resulting in models that generalize better by down-weighting high-order interaction effects. % that are difficult to learn from limited training data.

\section{Analysis: Dropout Regularizes Interaction Effects}
\label{sec:dropout_regularizes}
Dropout operates by probabilistically setting values to zero (i.e. multiplying by a Bernoulli mask). 
For clarity, we call this operation ``Input Dropout'' if the perturbed values are input variables, and ``Activation Dropout'' if the perturbed values are activations of hidden nodes. 

\paragraph{Input Dropout Shrinks Interaction Effects}
First, Input Dropout replaces the training dataset with samples drawn from a perturbed distribution: 
\begin{theorem}
Let $\E[Y|X] = F(X) =  \sum_{u \in [d]}f_u(X_u)$ and $\tilde{Y} = F(X \odot M)$, where $M_p \sim \text{Bernoulli}(p)$ is the Input Dropout mask and $\odot$ is element-wise multiplication. Then
\begin{align}
    \E[\tilde{Y}|X] = \sum_{u \in [d]}(1-p)^{|u|}f_u(X) + a_u(X_u).
\end{align} 
{\footnotesize with} $a_u(X_u) = \sum_{v \subseteq u}p^{|u|-|v|}(1-p)^{|v|}f_u(X_{u\backslash v}, X_v=0)$. 
\end{theorem}
This theorem shows that Input Dropout shrinks the conditional expectation of $\tilde{Y}|X$ by preferentially targeting high-order interactions: the scaling factor $(1-p)^{|u|}$ shrinks exponentially with $|u|$. 
For multiplicative interaction effects, we can simplify $a_u(X_u)$:

\begin{corollary}
Let $f_u(X_u)$ be a multiplicative interaction effect such that $f_u(X_u) = \prod_{u' \in u}g_{u'}(X_{u'})$ with $g_{u'}(0) = 0$ for each $u'$. Then
\begin{align}
    \E[\tilde{Y}|X] = \sum_{u \in [d]}(1-p)^{|u|}f_u(X) %a_u(X_u) = 0
\end{align}
\end{corollary}

\begin{figure}[h]
    \centering
    \begin{subfigure}[b]{0.22\textwidth}
        \centering
        \includegraphics[width=\textwidth]{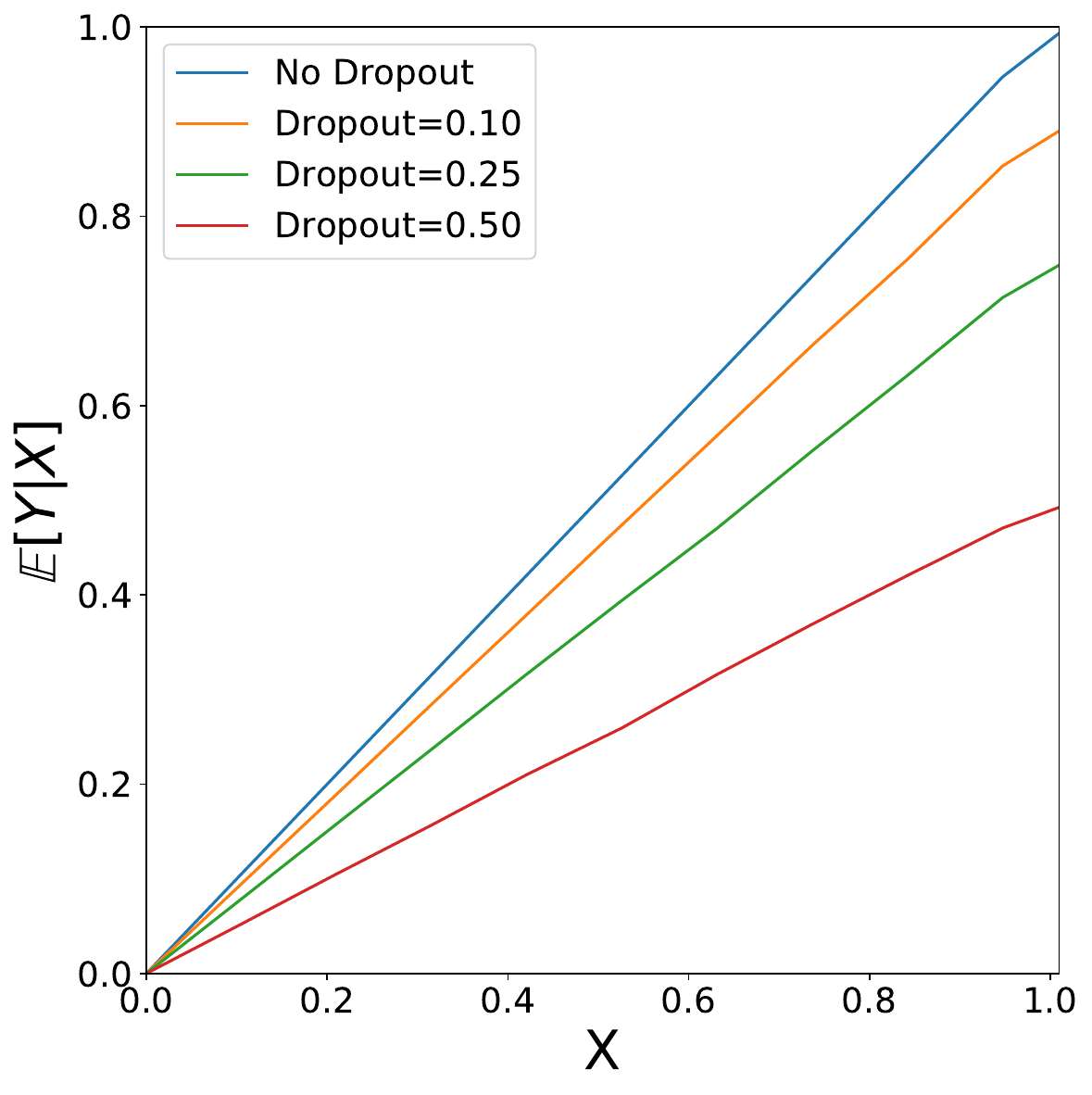}
        \caption{$k=1$}
    \end{subfigure}
    ~
    \begin{subfigure}[b]{0.22\textwidth}
        \centering
        \includegraphics[width=\textwidth]{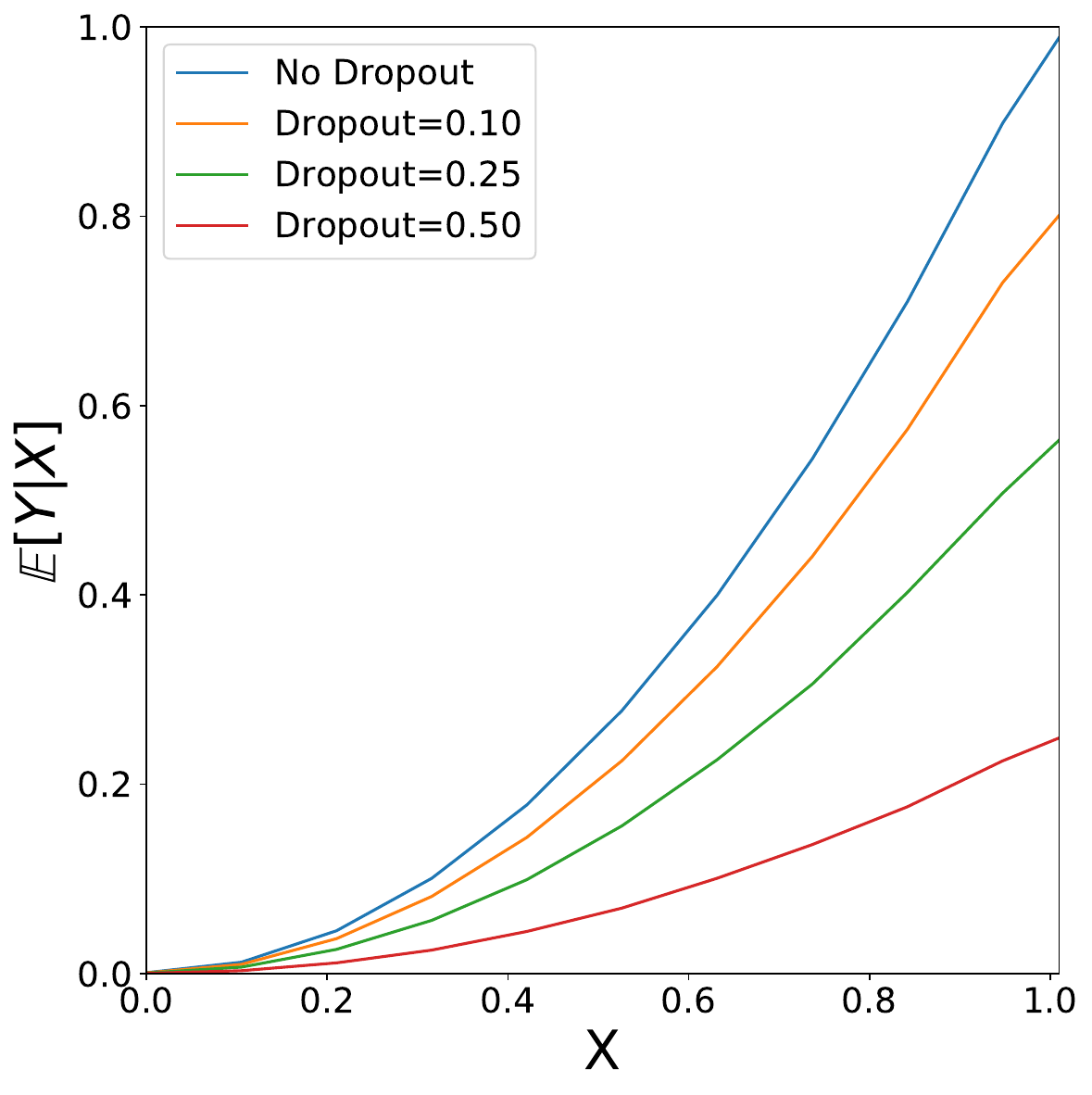}
        \caption{$k=2$}
    \end{subfigure}
    ~
    \begin{subfigure}[b]{0.22\textwidth}
        \centering
        \includegraphics[width=\textwidth]{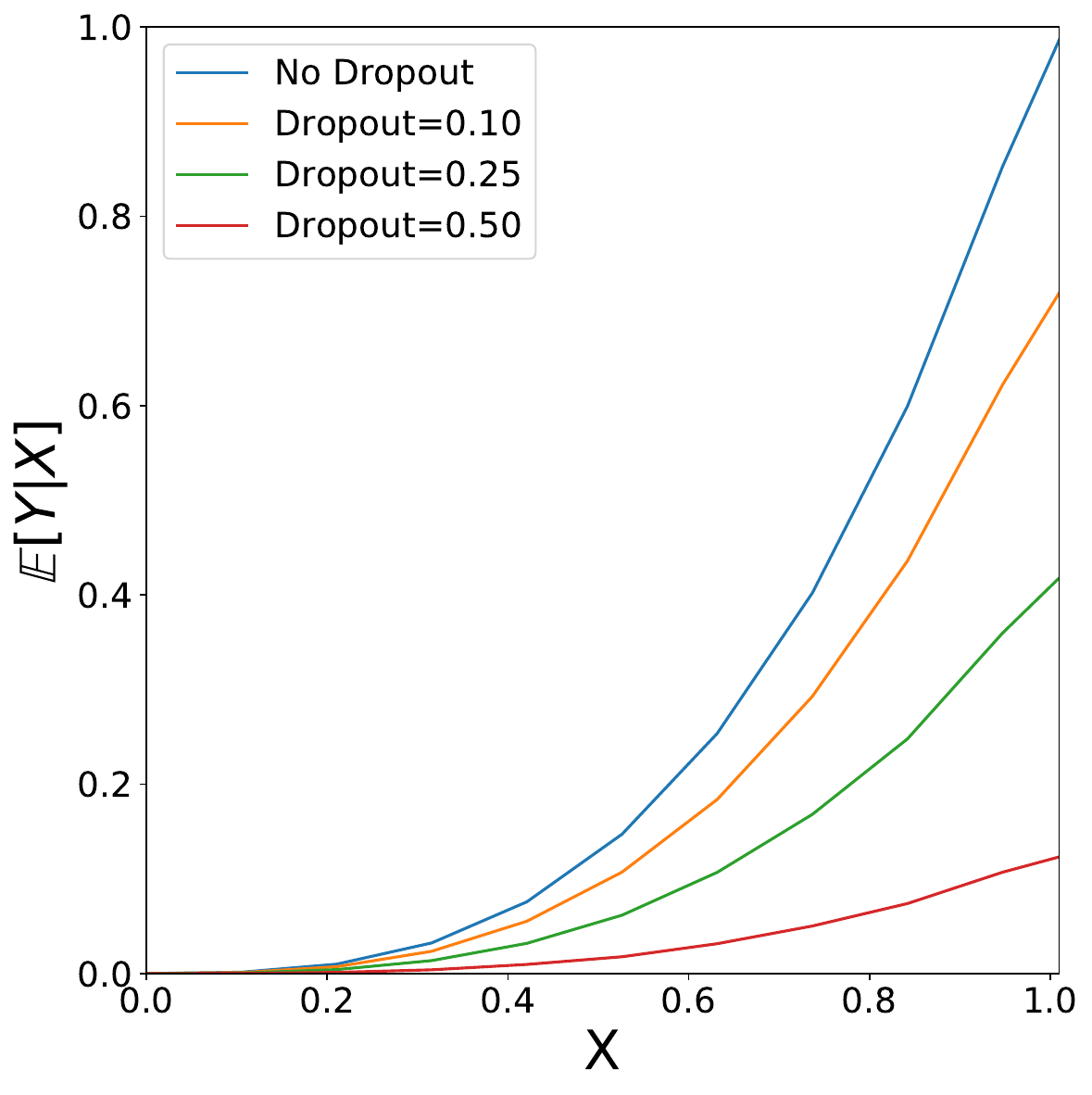}
        \caption{$k=3$}
    \end{subfigure}
    ~
    \begin{subfigure}[b]{0.22\textwidth}
        \centering
        \includegraphics[width=\textwidth]{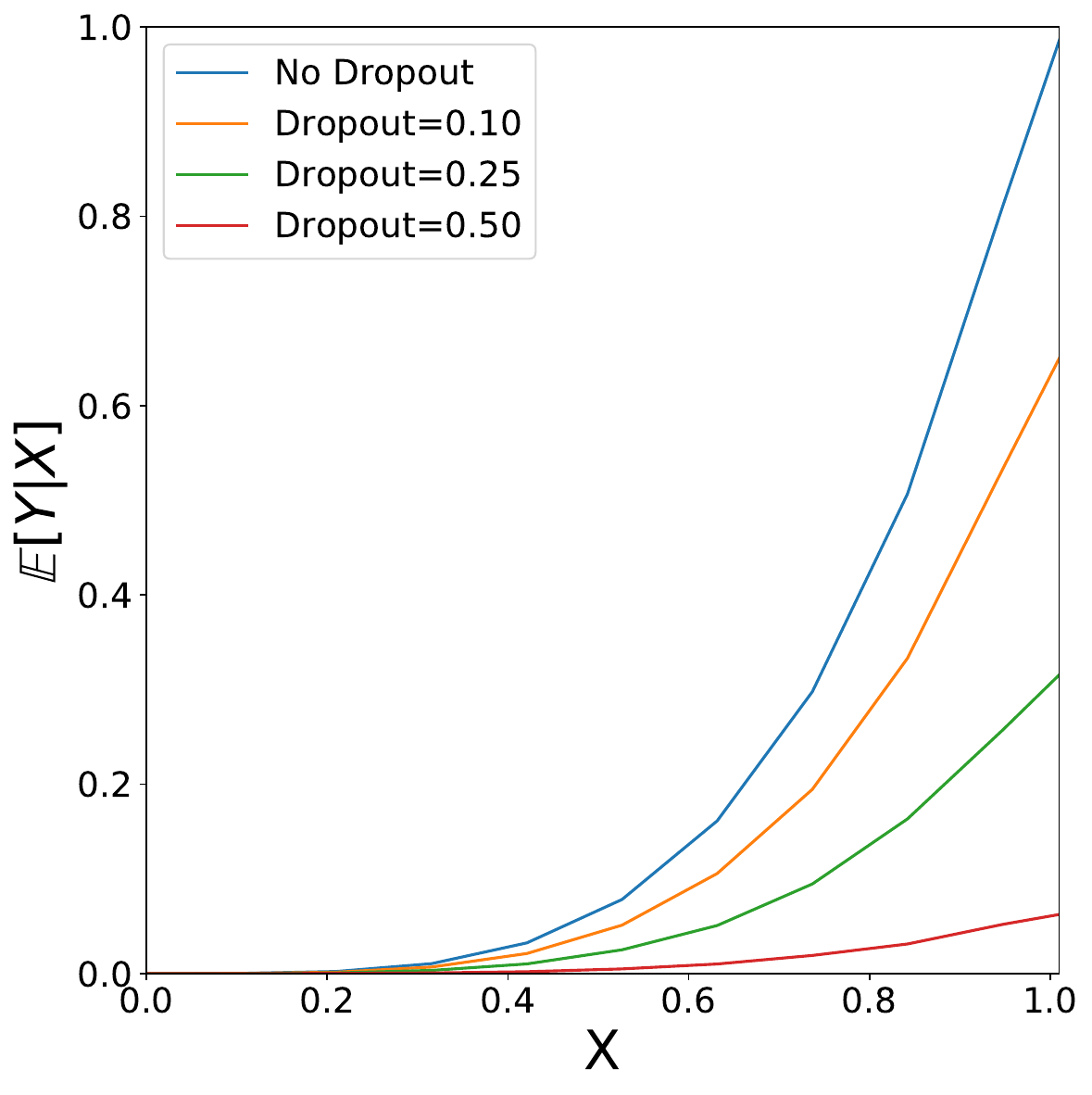}
        \caption{$k=4$}
    \end{subfigure}
    \caption{Visualizing the impact of Input Dropout on observed conditional means. we visualize the outcome which is a fixed multiplicative interaction effect of order $k$. Each colored line provides the conditional expectation of the outcome observed under a particular frequency of Input Dropout. As $k$ increases, the shrinkage effect of Input Dropout increases.}
    \label{fig:vis_input}
\end{figure}

This theorem implies that for multiplicative interaction effects, the shrinkage factor is exact. 
We can visualize this order-specific shrinkage of Input Dropout by empirically calculating $\E[\tilde{Y}|X]$ for $\tilde{Y}=f_u(X_u\odot M)$. 
In Figure~\ref{fig:vis_input}, we examine four multiplicative interaction effects: $f_u(X) = X_1$ for $k=1$, $f_u(X)=X_1X_2$ for $k=2$, $f_u(X)=X_1X_2X_3$ for $k=3$, and $f_u(X)=X_1X_2X_3X_4$ for $k=4$. 
For each of these interaction orders, we measure $\E[\tilde{Y}|X]$ for various input Dropout rates. 
We see that the shrinkage effect of Input Dropout is strongest for $k=4$ (e.g. the conditional mean observed under Input Dropout of $p=0.5$ is 12.5\% of the conditional mean observed without Input Dropout), and weakest for $k=1$ (e.g. the conditional mean observed under Input Dropout of $p=0.5$ is 50\% of the conditional mean observed without input Dropout).

Thus, the conditional expectation of the observed outcome is changed by applying Input Dropout, and the change is dependent on the order of the interaction effect. Practical implications include: (1) Input Dropout changes the distribution of model predictions, so even NNs trained for more epochs or with large sample size cannot overcome the bias introduced by Input Dropout and will converge to different optima based on the Input Dropout level. This is unlike L1 or L2 weight regularization which can be overcome by increasing the size of the training set and are affected by the downstream net architecture. (2) Input Dropout affects higher-order interactions more than lower-order interactions, biasing the prediction of any model (regardless of model training procedure).

\paragraph{Dropout Shrinks Gradients}
Next, we examine how Dropout affects training: 
\begin{corollary}
Let $G(X, Y, \theta) = \nabla_{\theta}(f_{\theta}(X), Y)$ be the gradient update for parameters $\theta$ on data $X,Y$. Let $G(X,Y,\theta) = \sum_{u}g_{uvw}(X_u, Y_v, \theta_w)$ be the fANOVA decomposition of $G(X, Y, \theta)$. Then for mask $M_p \sim \text{Bern}(p)$
\begin{align}
    \E[g_{uvw}(X_u \odot M_p, Y_v, \theta_w)] = 
    (1-p)^{|u|}g_{uvw}(X_u, Y_v, \theta_w) + a_{uvw}(X_u, Y_v, \theta_w) \nonumber
\end{align}
where $\E_{X.Y,\theta}[a_{uvw}(X_u, Y_v, \theta_w)] = 0$.
\end{corollary}

\begin{figure}[h]
    \centering
    \begin{subfigure}[b]{0.22\textwidth}
        \centering
        \includegraphics[width=\textwidth]{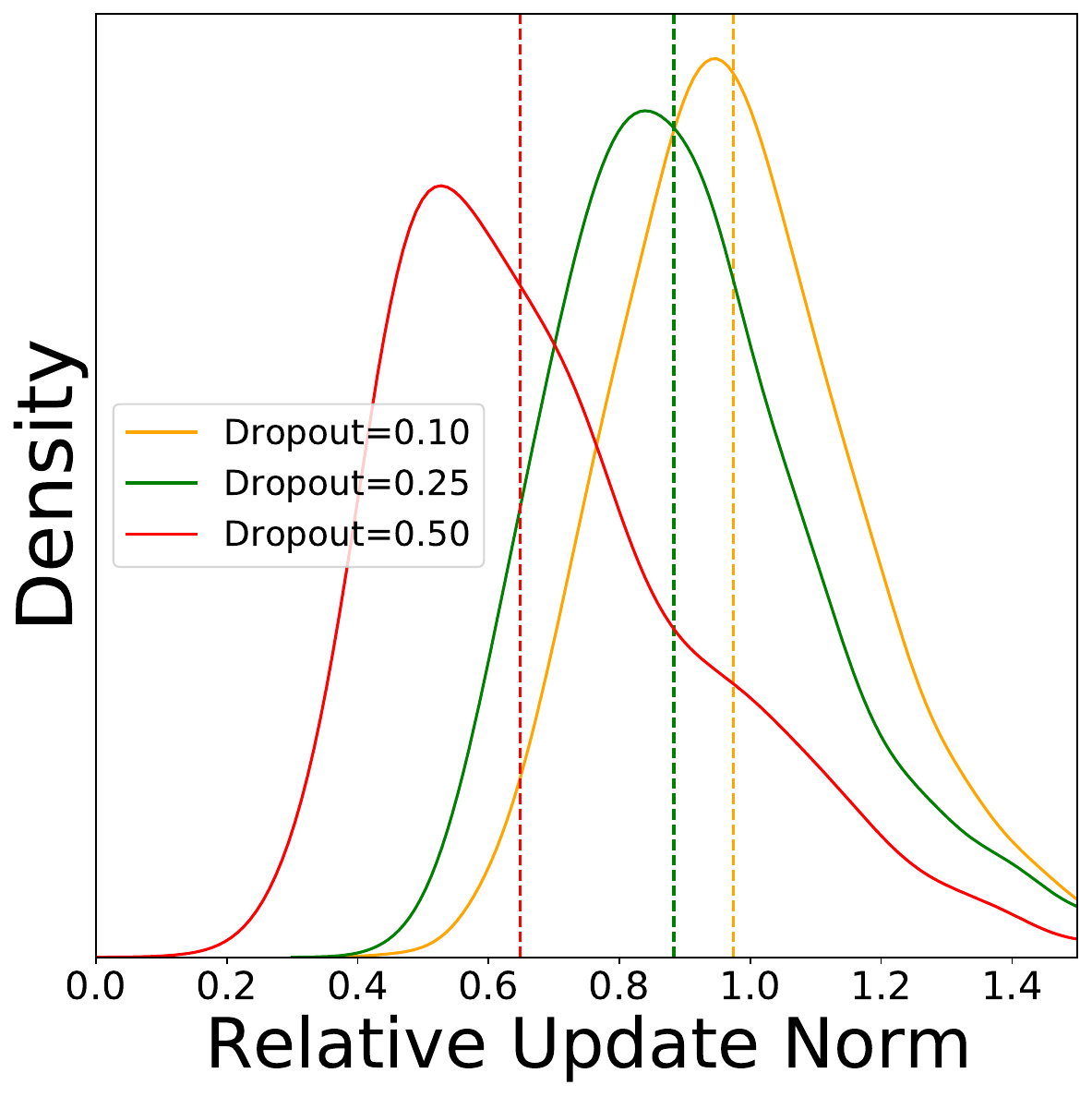}
        \caption{$k=1$}
    \end{subfigure}
    ~
    \begin{subfigure}[b]{0.22\textwidth}
        \centering
        \includegraphics[width=\textwidth]{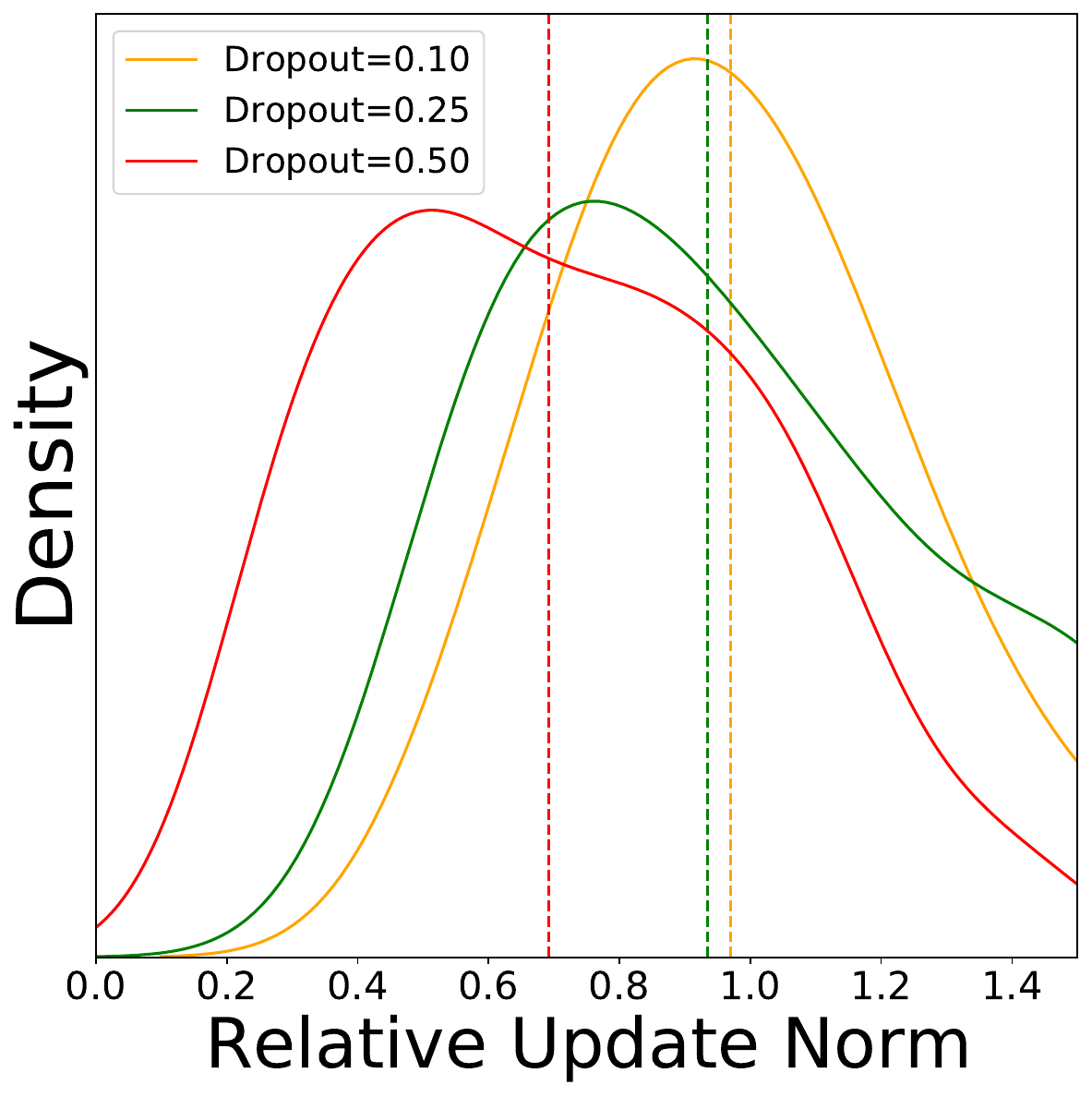}
        \caption{$k=2$}
    \end{subfigure}
    ~
    \begin{subfigure}[b]{0.22\textwidth}
        \centering
        \includegraphics[width=\textwidth]{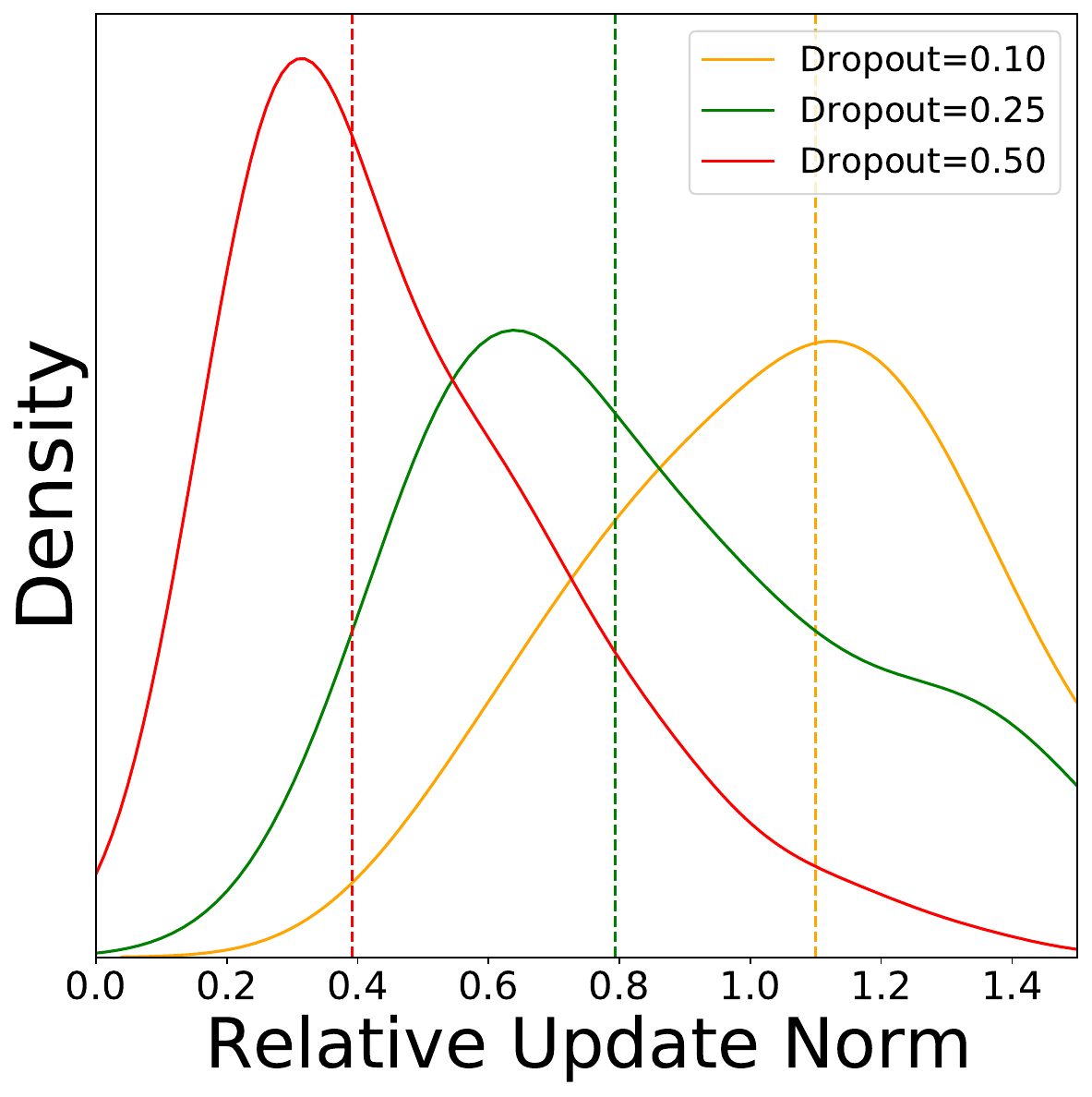}
        \caption{$k=3$}
    \end{subfigure}
    ~
    \begin{subfigure}[b]{0.22\textwidth}
        \centering
        \includegraphics[width=\textwidth]{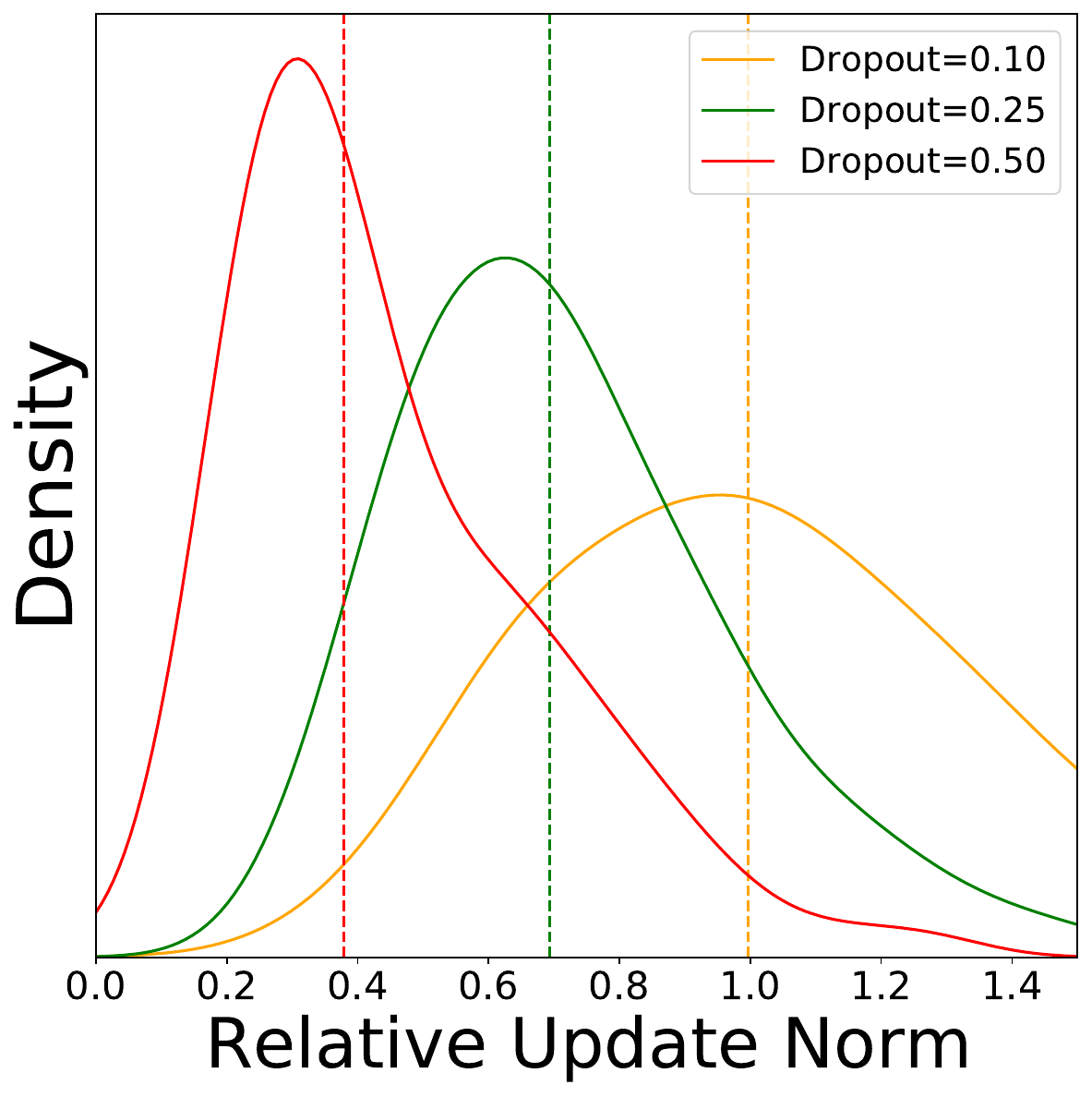}
        \caption{$k=4$}
    \end{subfigure}
    \caption{Distribution of gradient sizes based on training on various orders of interaction effect ($k=1,2,3,$ or $4$) and with various levels of Dropout. We normalize the gradient sizes by comparing to the gradient induced by training for the same training points and initialization without any Dropout. Dashed vertical lines indicate the median of each distribution (best viewed in color). The gradient norms are down-weighted for higher-interaction effects and for higher Dropout rates.
    \label{fig:vis_grads}}
\end{figure}

That is, Input Dropout shrinks gradient updates according to the order of the interaction effect which produced the gradient updates. 
We can visualize this effect by calculating the gradient at initialization induced by NNs trained with various levels of input Dropout. 
In Figure.~\ref{fig:vis_grads}, we show the distribution of the $\ell_1$-norm norm of these gradients normalized by the $\ell_1$-norm of the gradients for a NN without input Dropout. We see that for NNs fitting to an interaction effect of order $k=1$ (Fig.~\ref{fig:vis_grads}a), the gradients are shrunk only slightly under large Dropout, while for NNs fitting to an interaction effect of order $k=4$ (Fig.~\ref{fig:vis_grads}d), the gradients are shrunk severely under large Dropout. 

To describe Activation Dropout, we modify $G$ to act on activations of a particular layer:
\begin{corollary}
Let $G_i(A_i, Y, \theta) = \nabla_{\theta}(f_{\theta}(A_i), Y)$ be the gradient update for parameters $\theta$ from activation $A_i$ at layer $i$ with target outcome $Y$. Let $G_i(A_i,Y,\theta) = \sum_{u}g_{iuvw}(A_{iu}, Y_v, \theta_w)$ be the fANOVA decomposition of $G_i(A_i, Y, \theta)$. Then for mask $M_p \sim Bernoulli(p)$,
\begin{align}
    \E[g_{iuvw}(A_{iu} \odot M_p, Y_v, \theta_w)] = 
    (1-p)^{|u|}g_{iuvw}(A_{iu}, Y_v, \theta_w) + a_{uvw}(A_{iu}, Y_v, \theta_w) \nonumber
\end{align}
where $\E_{A_i.Y,\theta}[a_{uvw}(A_{iu}, Y_v, \theta_w)] = 0$.
\end{corollary}

Activation Dropout thus shrinks the gradient update according to the number of hidden nodes in layer $i$ which interact to form the gradient update. 
Thus, both Input Dropout and Activation Dropout correspond to order-specific \emph{effective learning rates} $r_p(k) = (1-p)^{k}$ %which decay exponentially in the interaction order $k$ 
(the distinction being whether $k$ counts the \emph{input features} in the interaction or counts the \emph{hidden activations} in the interaction)\footnote{This explains why Dropout tends to produce hidden units which are ``specialized" \citep{NIPS2019_8460}.}.

\begin{figure}[htb]
    \centering
    \begin{subfigure}[t]{0.45\columnwidth}
        \centering
        \includegraphics[width=\columnwidth]{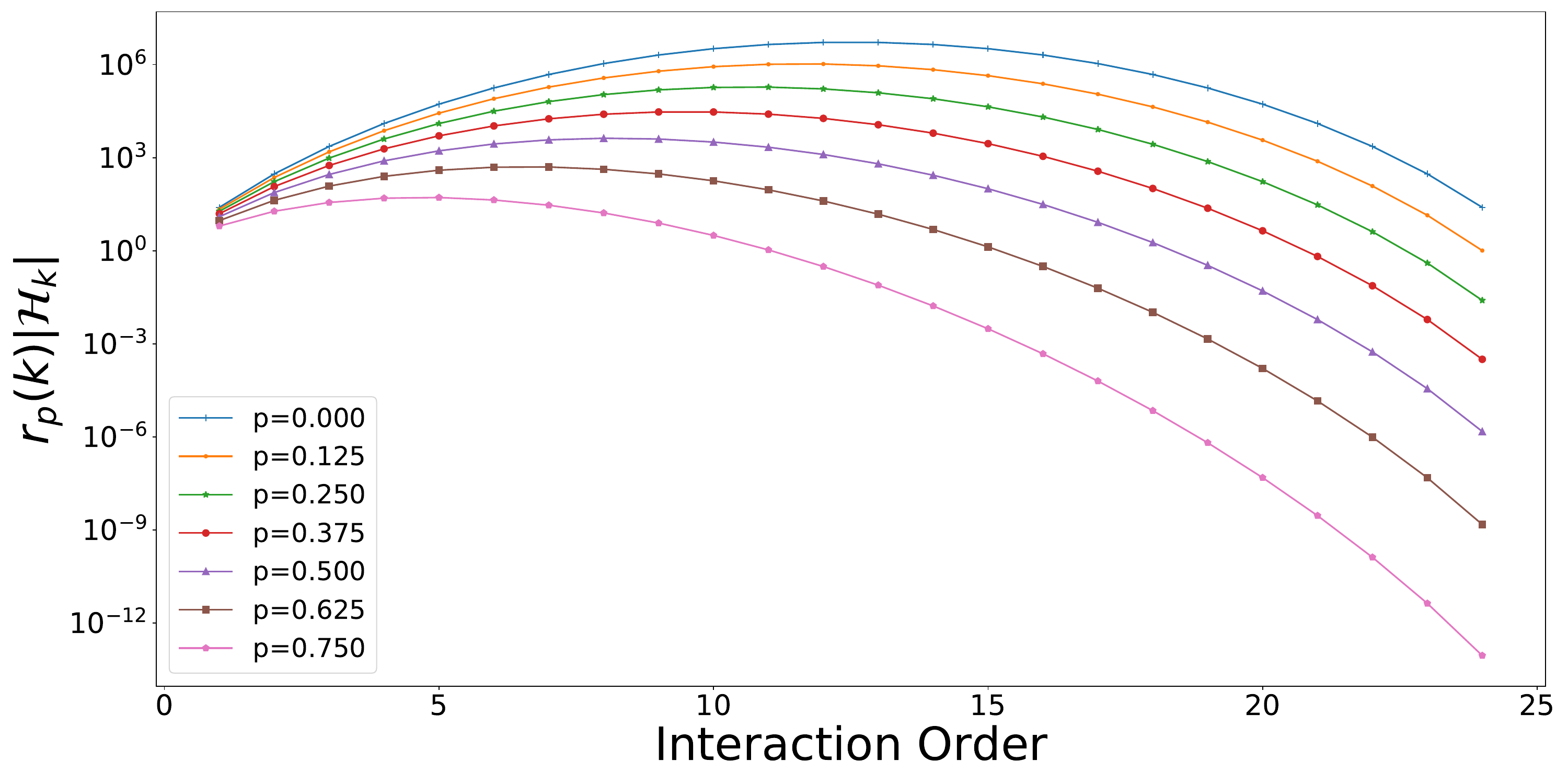}
        \caption{Full space. \label{fig:hypo:full}}
    \end{subfigure}
    ~
    \begin{subfigure}[t]{0.45\columnwidth}
        \centering
        \includegraphics[width=\columnwidth]{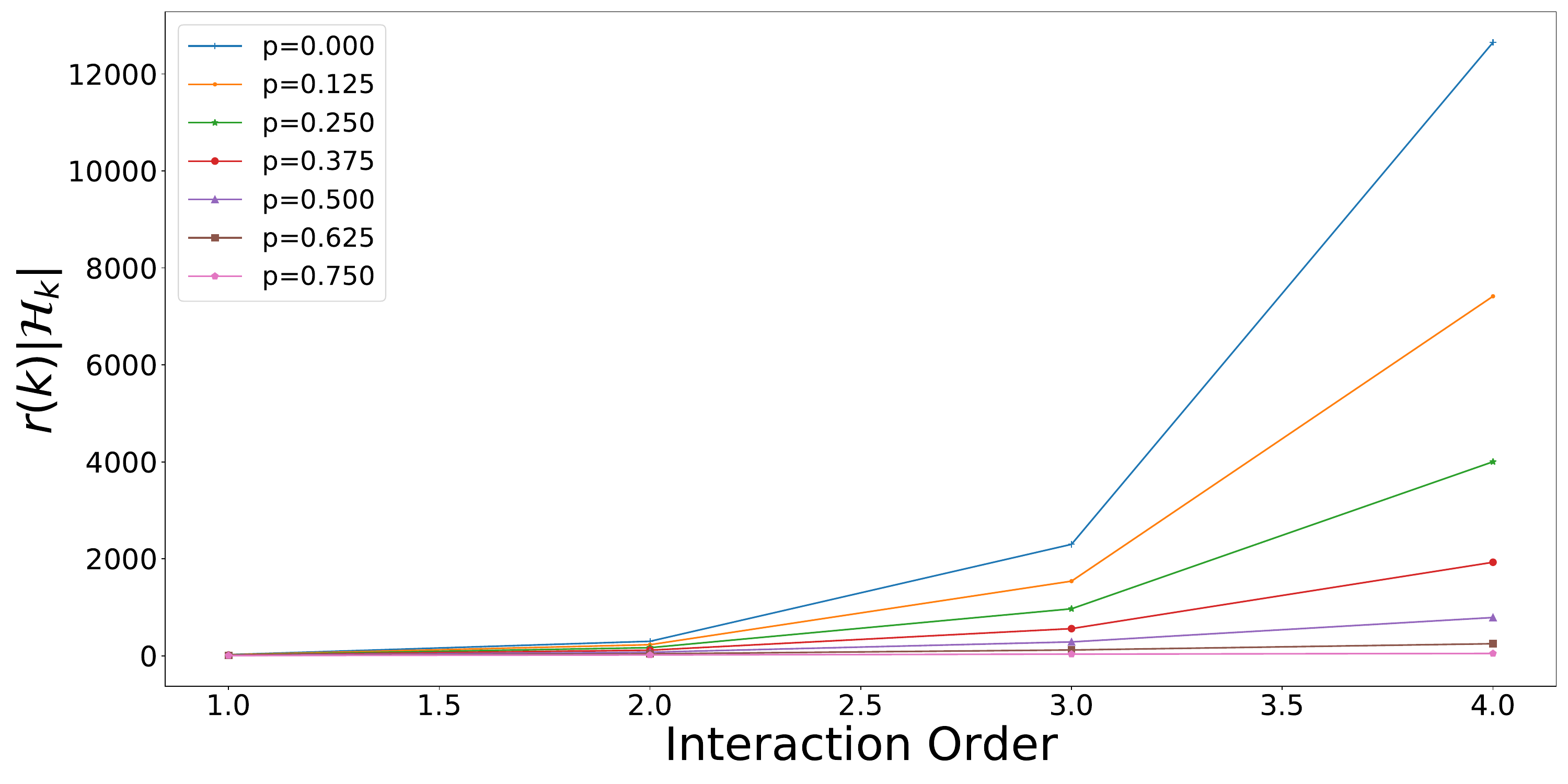}
        \caption{First four orders. \label{fig:hypo:first}}
    \end{subfigure}
    \caption{The growing hypothesis space of interaction effects is balanced against the effective learning rate imposed by Dropout. 
    In this figure, we plot the product of the effective learning rate ($r_p(k)$) and the number of potential interaction effects of order $k$ ($|\mathcal{H}_k|$). % for a variety of Dropout rates $p$. 
    In \subref{fig:hypo:full}, we plot these values on a log scale for the entire range of interaction orders for an input of $N=25$ features. In \subref{fig:hypo:first}, we plot up to order 4.% on a linear scale. 
    \label{fig:hypo}}
\end{figure}

\paragraph{Symmetry Between Dropout Strength and Number of Interaction Effects}
\label{sec:symmetry}
The effective learning rate $r_p(k) = (1-p)^k$ of $k$-order interactions decays exponentially with $k$. 
This is a symmetry with $\binom{N}{k}$. % (which is $\approx N^k$ for small $k$). %$\leq N^k$ for all $k$ and $\approx N^k$ for small $k$). 
As shown in Fig~\ref{fig:hypo}, the exponential growth of the hypothesis space $|\mathcal{H}_k| = \binom{N}{k}$ %with interaction order 
is balanced by the exponential decay of the effective learning rate.%, providing strong regularization against high-order interaction effects.

\section{Experiments}
\label{sec:experiments}
As shown above, Dropout exerts regularization against interaction effects.
Here, we decompose NNs to measure the strength of learned interaction effects.

\subsection{Measuring Interaction Effects in NNs}
\label{sec:intx_nn:measuring}

As with any function, the $\hat{F}(X)$ learned by a NN can be decomposed as:
%\begin{align}
$    \hat{F}(X) = \sum_{u \in [d]} \hat{f}_u(X_u)$
%\end{align}
by fANOVA (Eq.~\ref{eq:orthogonality}). 
%We use this decomposition to measure the interaction effects implicit in $\hat{F}$ estimated by NNs. 
To calculate this decomposition, we apply model distillation \cite{bucilua2006model,hinton2015distilling} using the \texttt{XGBoost} software package \cite{Chen2016XGBoostAS} to train boosted decision trees with maximum depth $k$ to distill the interaction effects of order $k$. 
By successively increasing $k$ and training on the residuals of the shallower trees, % of the previous model, 
the estimated effects are orthogonal and hence satisfy fANOVA requirements.

\paragraph{Accuracy of Decomposition}
How accurate is this distillation procedure for calculating the fANOVA decomposition of a NN? 
To empirically validate this procedure, we test distillation using simulation data. 
Our goal in this experiment is to %accurately fit a NN to a known function so that we can 
measure the error of distillation on NNs representing known functions. % against the known function encoded in the NN.
\footnote{
We do not claim that this distillation is always suitable as general-purpose explanations of NNs; in this context, however, we care about only a single aspect of the compressed models: approximation error. 
For example, from the NN $\hat{F}(X)$, we estimate an additive model $\hat{f}_1(X) \in \mathcal{S}(\hat{F}, \mathcal{F}_1) = \argmin_{f \in \mathcal{F}_1} \mathbb{E}_X [(f(X) - \hat{F}(X))^2 ]$ where $\mathcal{F}_1$ is the class of additive models. % and $\mathcal{L}$ is squared loss. 
The set of possible explanations $\mathcal{S}(\hat{F}, \mathcal{F}_1)$ may have more than one member; however, all of these explanations must have the same compression loss. % $\mathbb{E}_X \left[\mathcal{L}(f(X), \hat{F}(X)) \right]$. 
As the only metric we are reporting about these models is the compression loss, 
all members of $\mathcal{S}(\hat{F}, \mathcal{F}_1)$ are equivalent. 
}
For each run, we generate data according to $X \sim \text{Unif}(-1, 1)^5$, and train a NN to fit a pure $k$-order interaction (a multiplication of $k$ uncorrelated features of $X$). 
A perfect distillation procedure would assign 100\% of the variance to interactions of order $k$. 

\begin{figure}[htb]
    \centering
    \begin{subfigure}[b]{0.29\columnwidth}
        \centering
        \includegraphics[width=\textwidth]{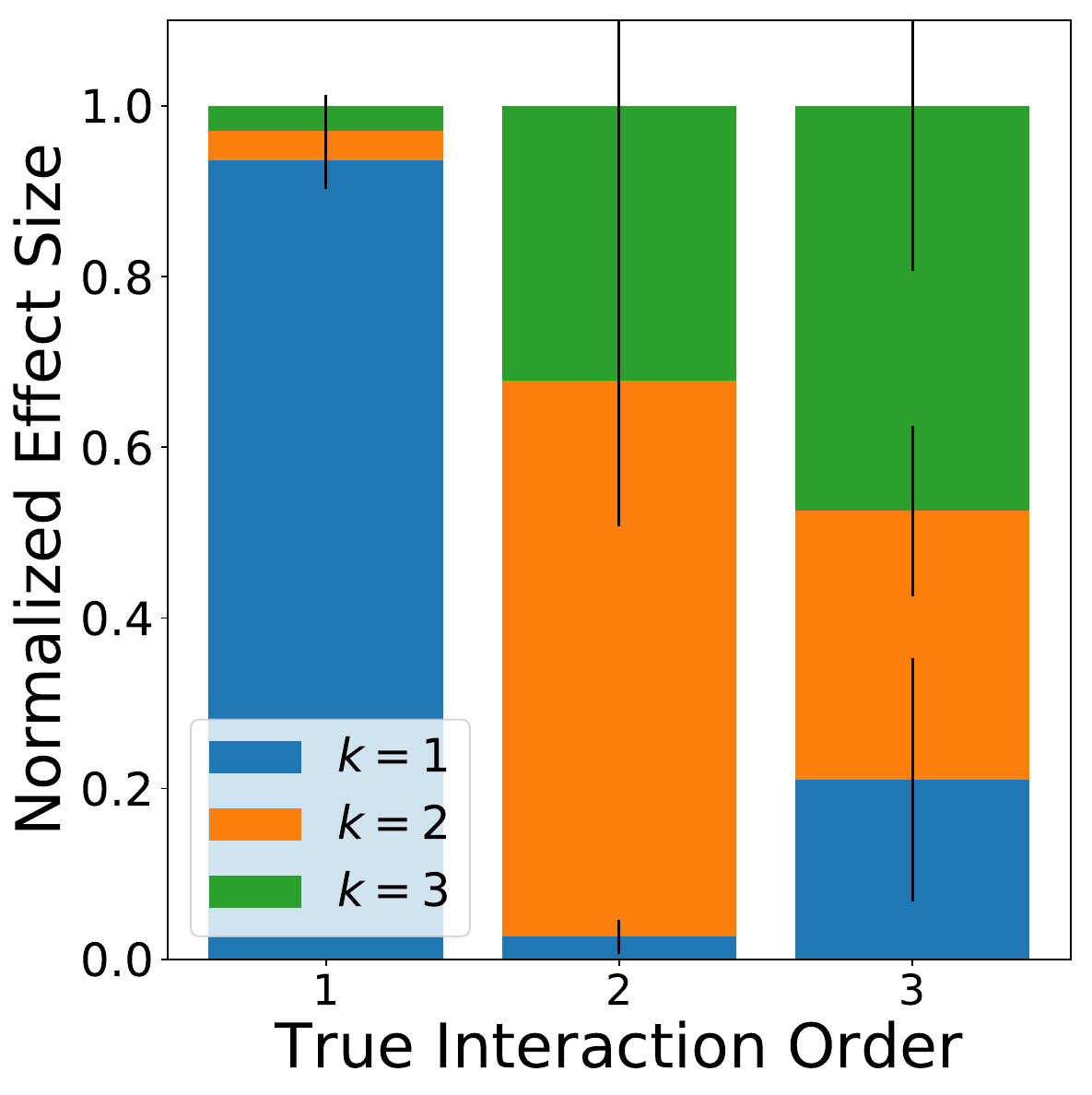}
        \caption{100 Samples\label{fig:simulation:100}}
    \end{subfigure}
    ~
    \begin{subfigure}[b]{0.29\columnwidth}
        \centering
        \includegraphics[width=\textwidth]{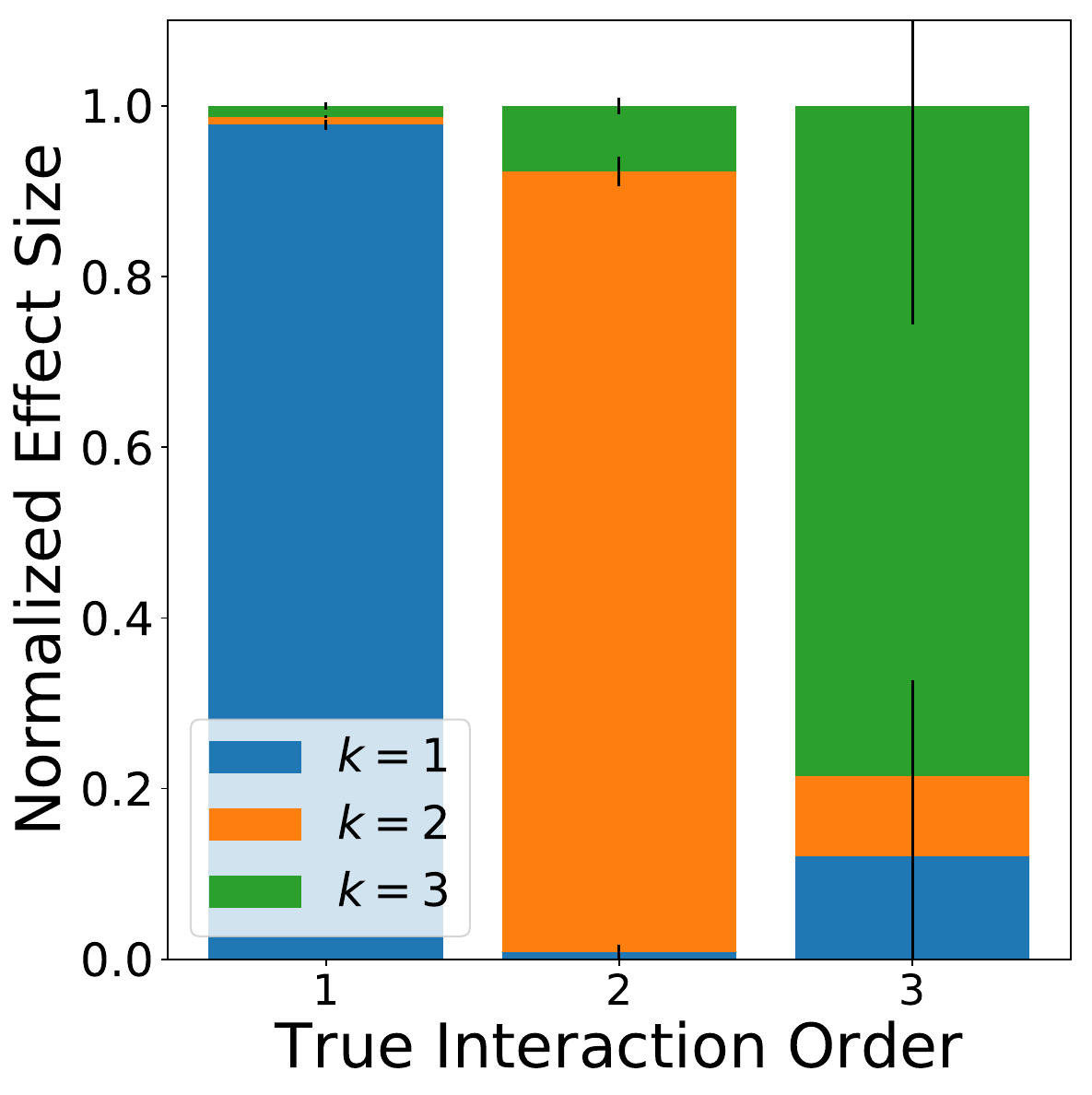}
        \caption{1000 Samples\label{fig:simulation:1000}}
    \end{subfigure}
    ~
    \begin{subfigure}[b]{0.32\columnwidth}
        \centering
        \includegraphics[width=0.93\textwidth]{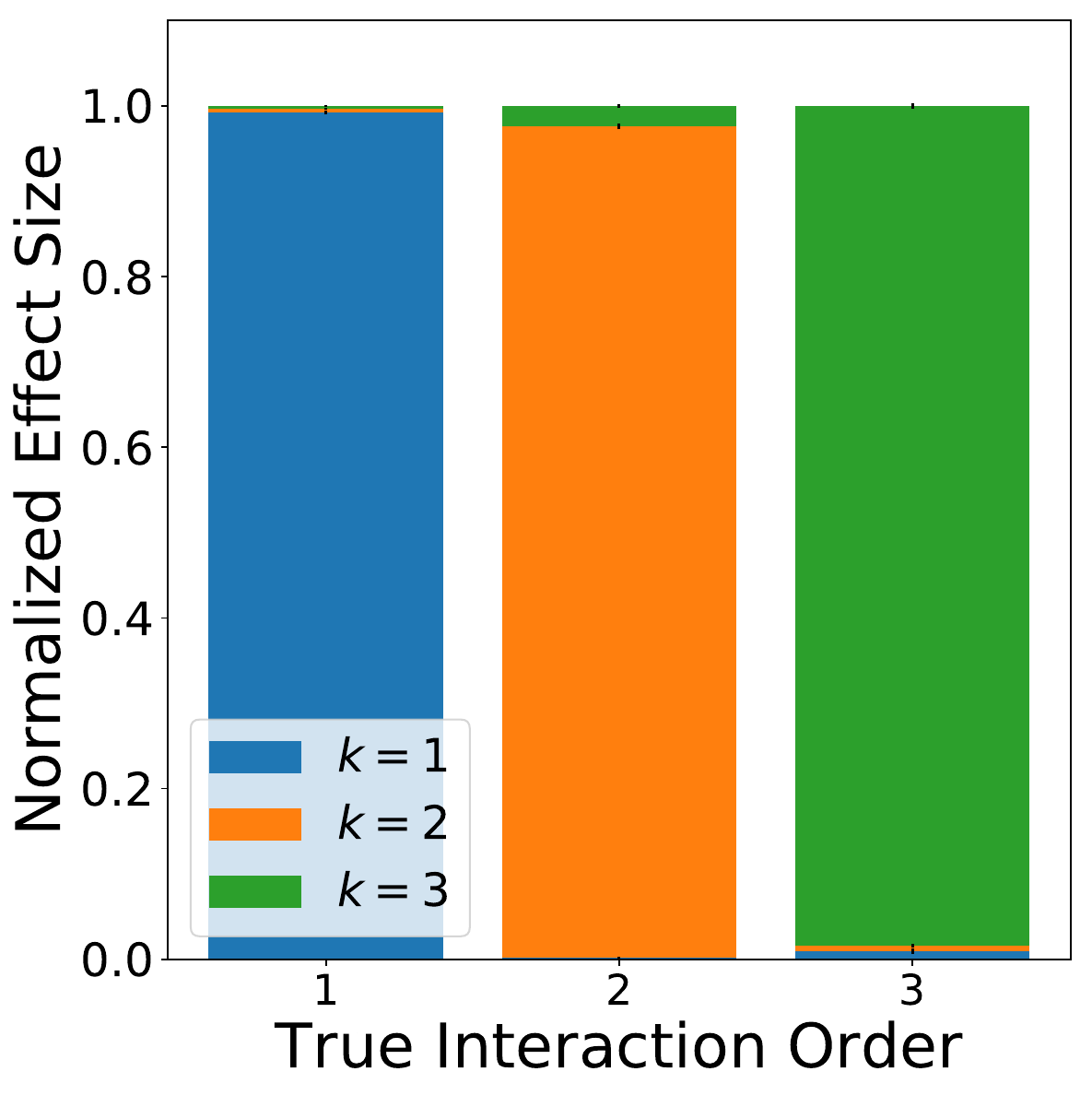}
        \caption{10000 Samples \label{fig:simulation:10000}}
    \end{subfigure}
    \caption{Distilled effect sizes (mean $\pm$ var over 10 runs) of the interactions in NNs representing interactions of order 1, 2, or 3. The size of effect $f_X$ is defined as $\text{Var}_X(f_X(X))$. Each pane shows results for a number of samples used for distillation. 
    \label{fig:simulation}}
\end{figure}

Results are shown in Fig.~\ref{fig:simulation}. 
In each pane, there are 3 bars which each represent an order. 
The height of the bars (and the corresponding colors) represent the normalized effect size estimated by the distillation procedure. % for each of these underlying interaction effects. 
In Fig.~\ref{fig:simulation:100}, only 100 samples are used for distillation; as a result, the low-order models underfit the NN and exaggerate the effects of high-order interactions.  
When the number of samples is increased to 1000 (Fig.~\ref{fig:simulation:1000}) or to 10000 (Fig.~\ref{fig:simulation:10000}), the distillation procedure is increasingly accurate at recovering the true interaction breakdown in the NN.
Importantly, none of the distillations over-estimated the influence of low-order interaction effects.

\subsection{Dropout Regularizes Spurious Interactions}
\label{sec:noise_expt}
\begin{figure*}[t]
    \centering
    \begin{subfigure}[t]{0.25\textwidth}
        \centering
        \includegraphics[width=\columnwidth]{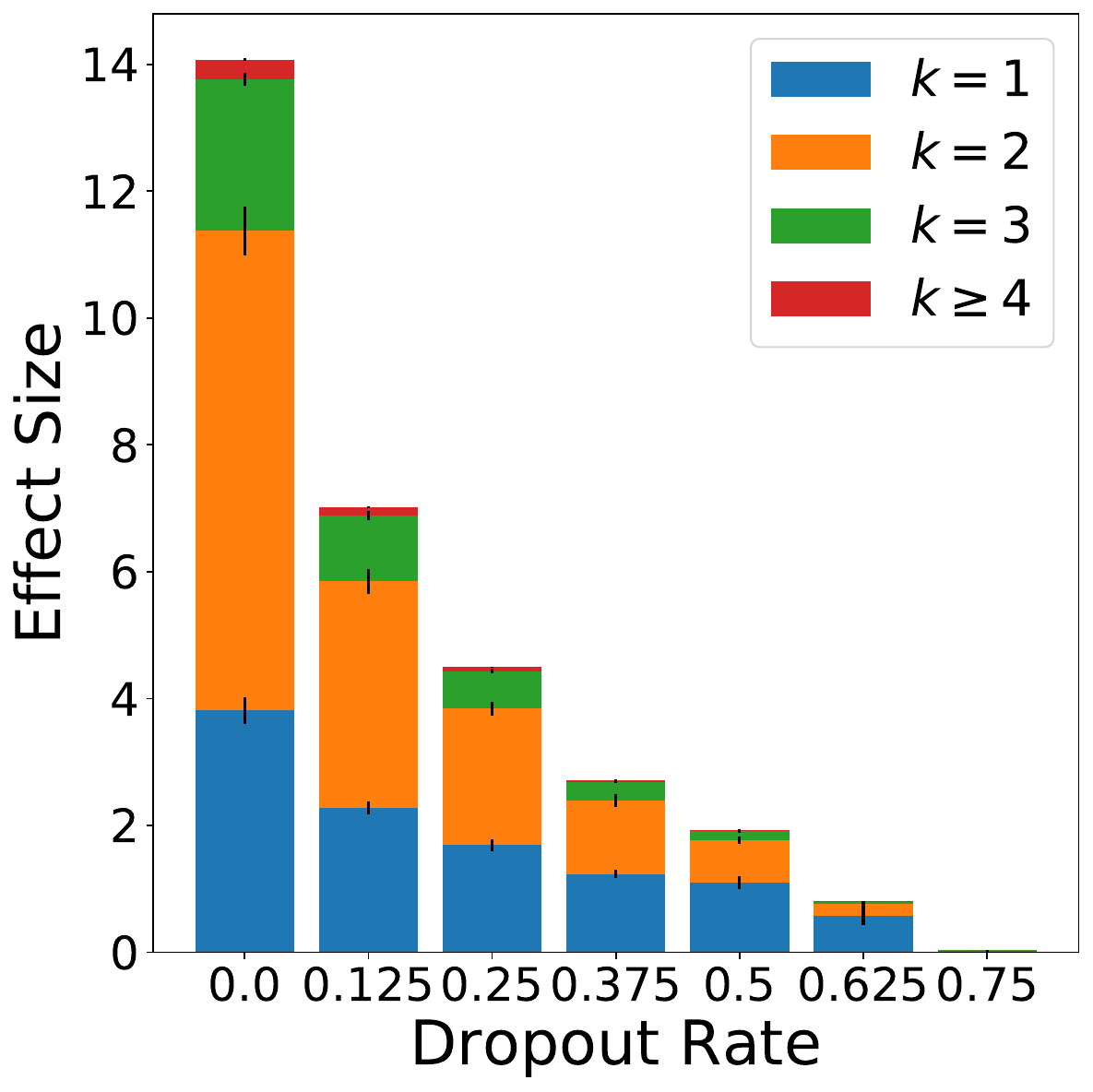}
        \caption{Total: Activation 
        \label{fig:converged_32:total_activation}}
    \end{subfigure}
    ~
    \begin{subfigure}[t]{0.25\textwidth}
        \centering
        \includegraphics[width=\columnwidth]{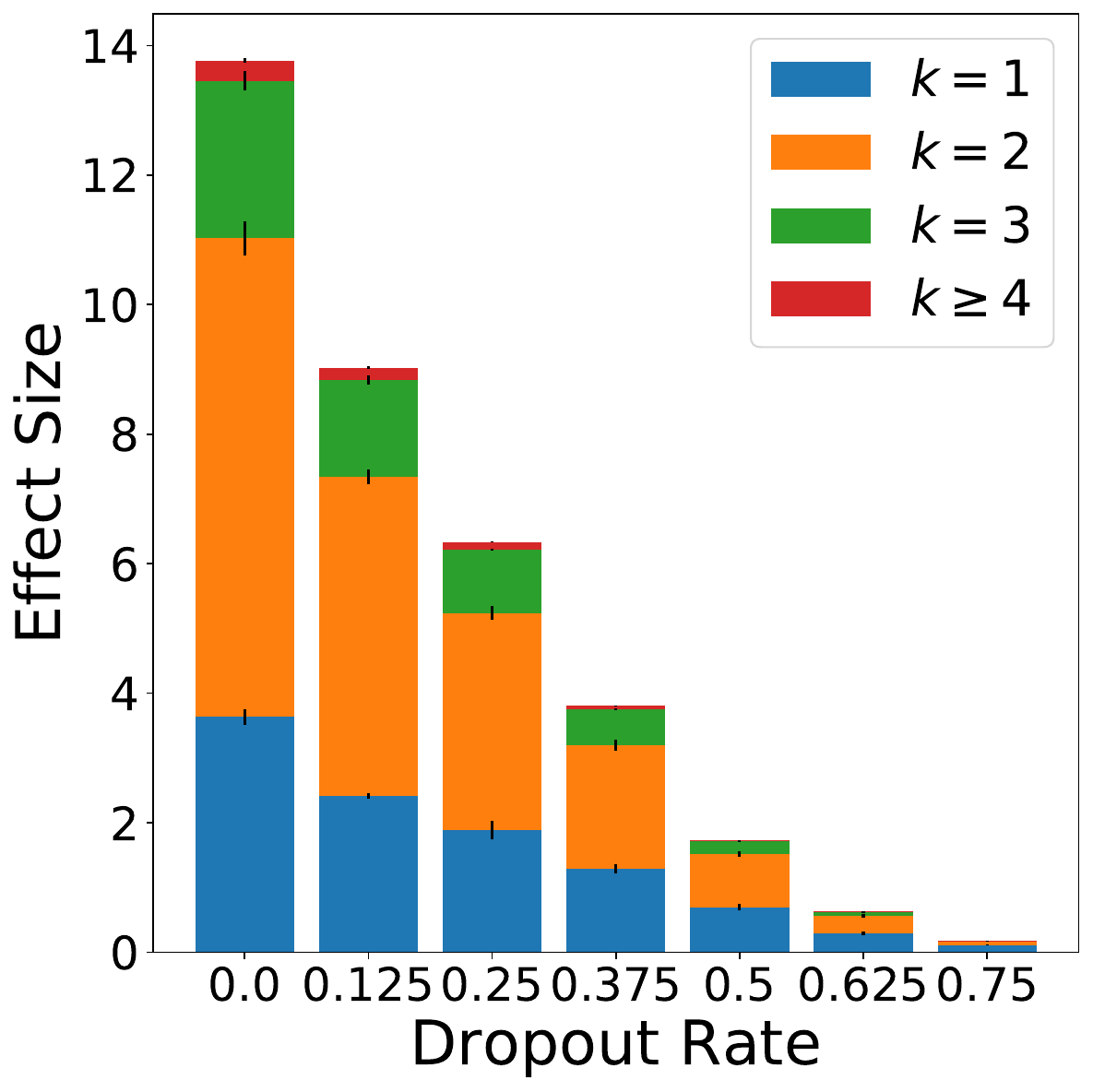}
        \caption{Total: Input
        \label{fig:converged_32:total_input}}
    \end{subfigure}
    ~
    \begin{subfigure}[t]{0.25\textwidth}
        \centering
        \includegraphics[width=\columnwidth]{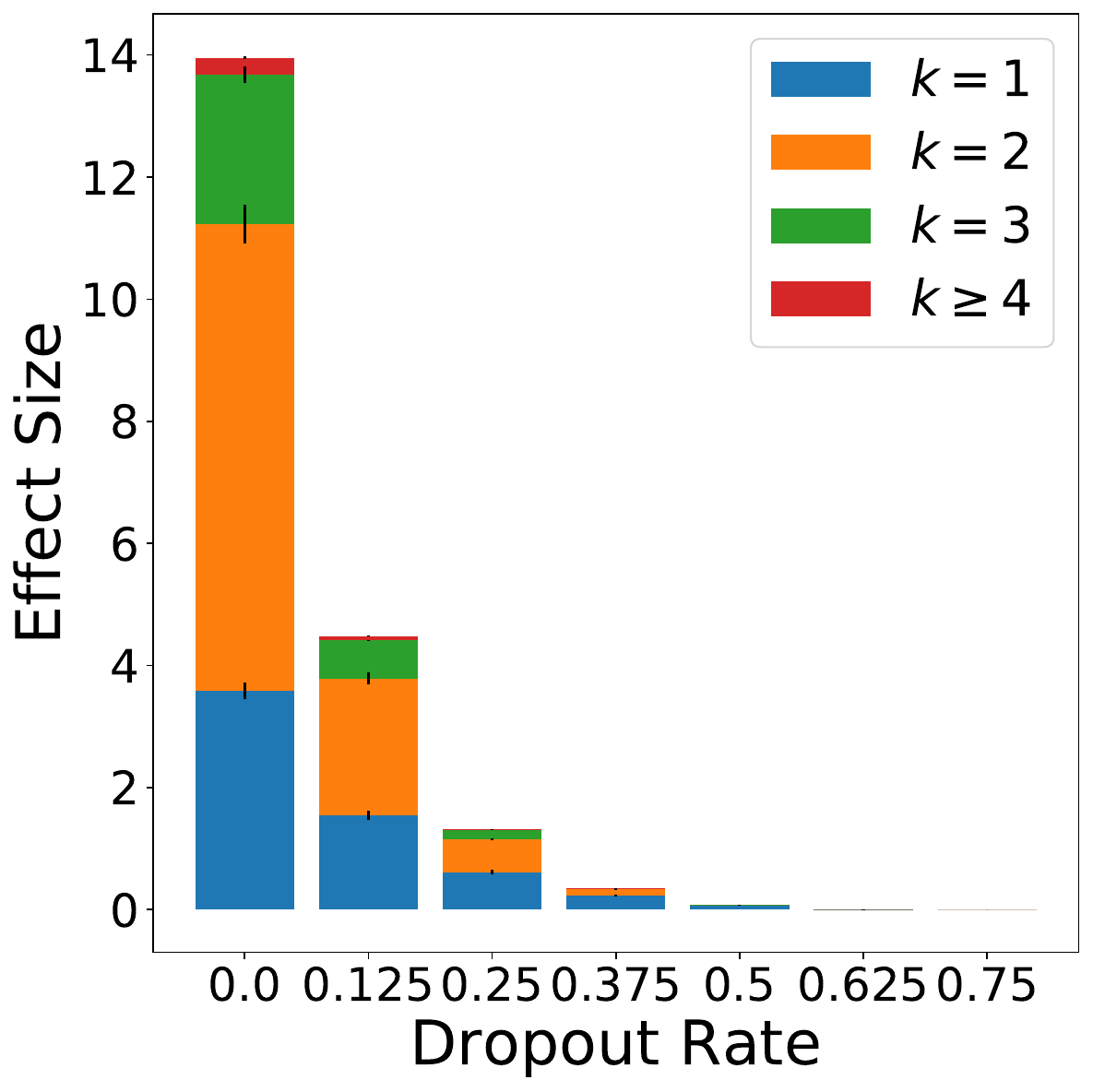}
        \caption{\footnotesize{Total: Input + Act.}
        \label{fig:converged_32:total_both}}
    \end{subfigure}
    \\
    \begin{subfigure}[t]{0.25\textwidth}
    \centering
        \includegraphics[width=\columnwidth]{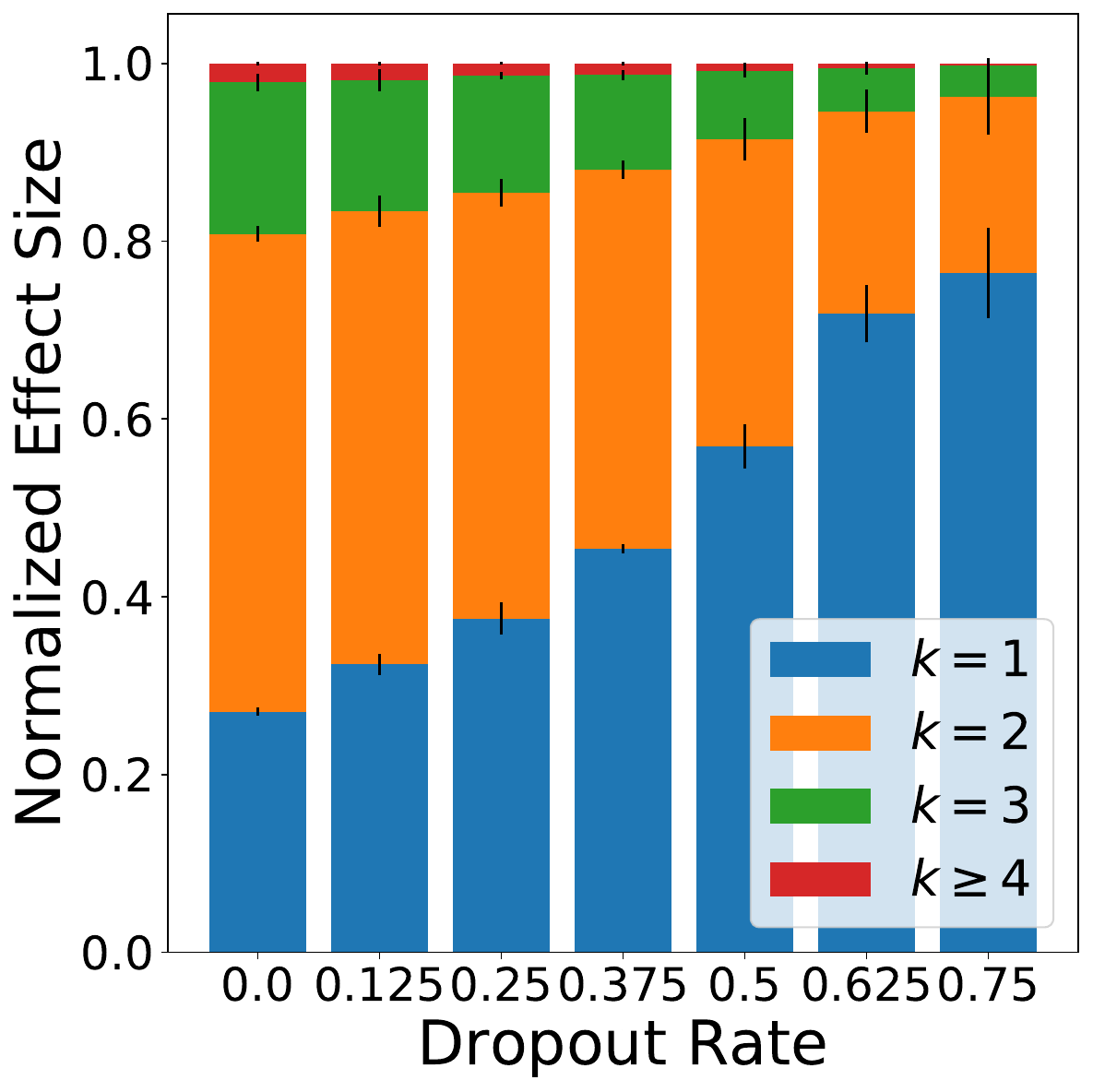}
        \caption{Normalized: Activation
        \label{fig:converged_32:norm_activation}}
    \end{subfigure}
    ~
    \begin{subfigure}[t]{0.25\textwidth}
    \centering
        \includegraphics[width=\columnwidth]{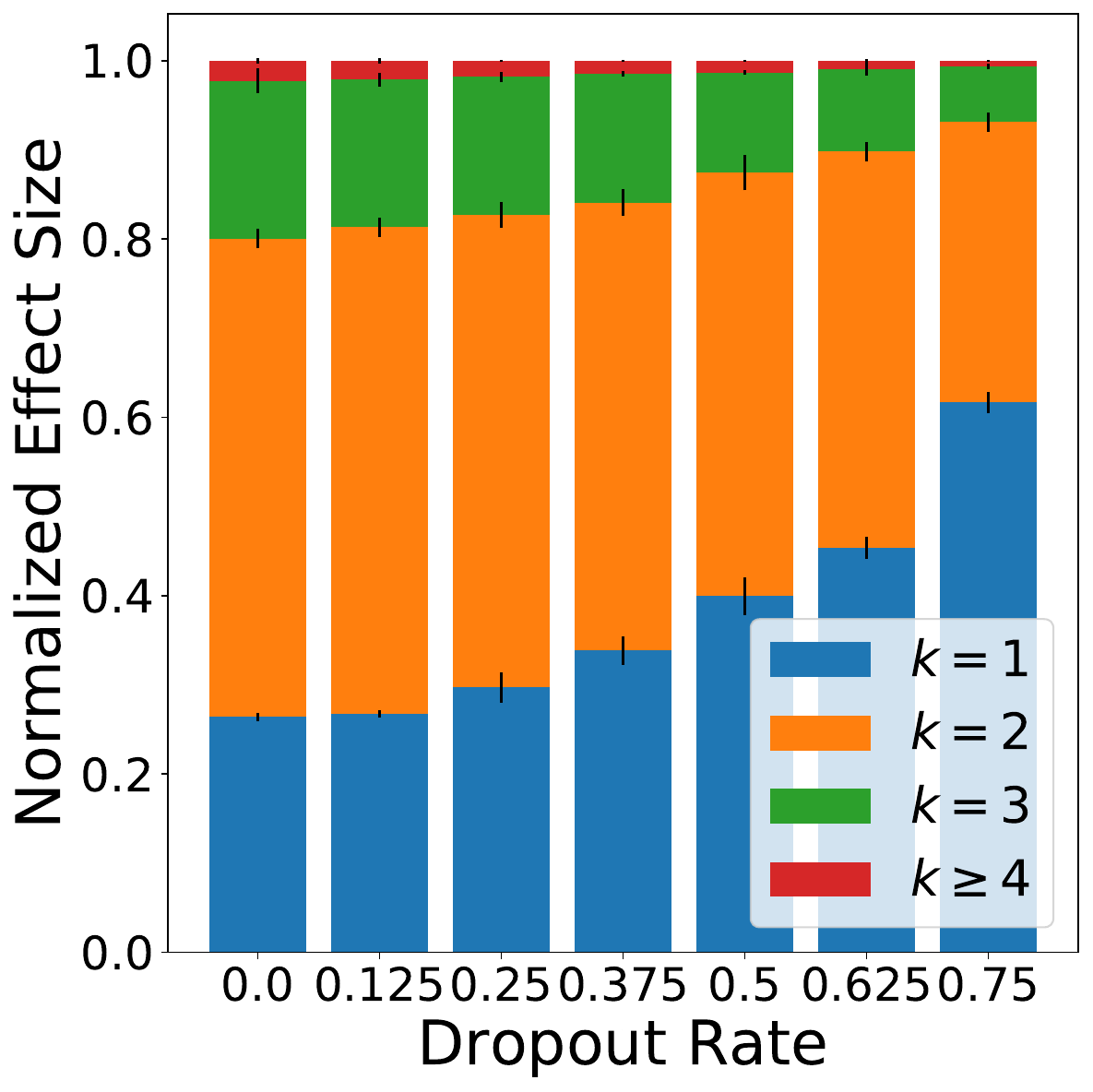}
        \caption{Normalized: Input
        \label{fig:converged_32:norm_input}}
    \end{subfigure}
    ~
    \begin{subfigure}[t]{0.25\textwidth}
    \centering
        \includegraphics[width=\columnwidth]{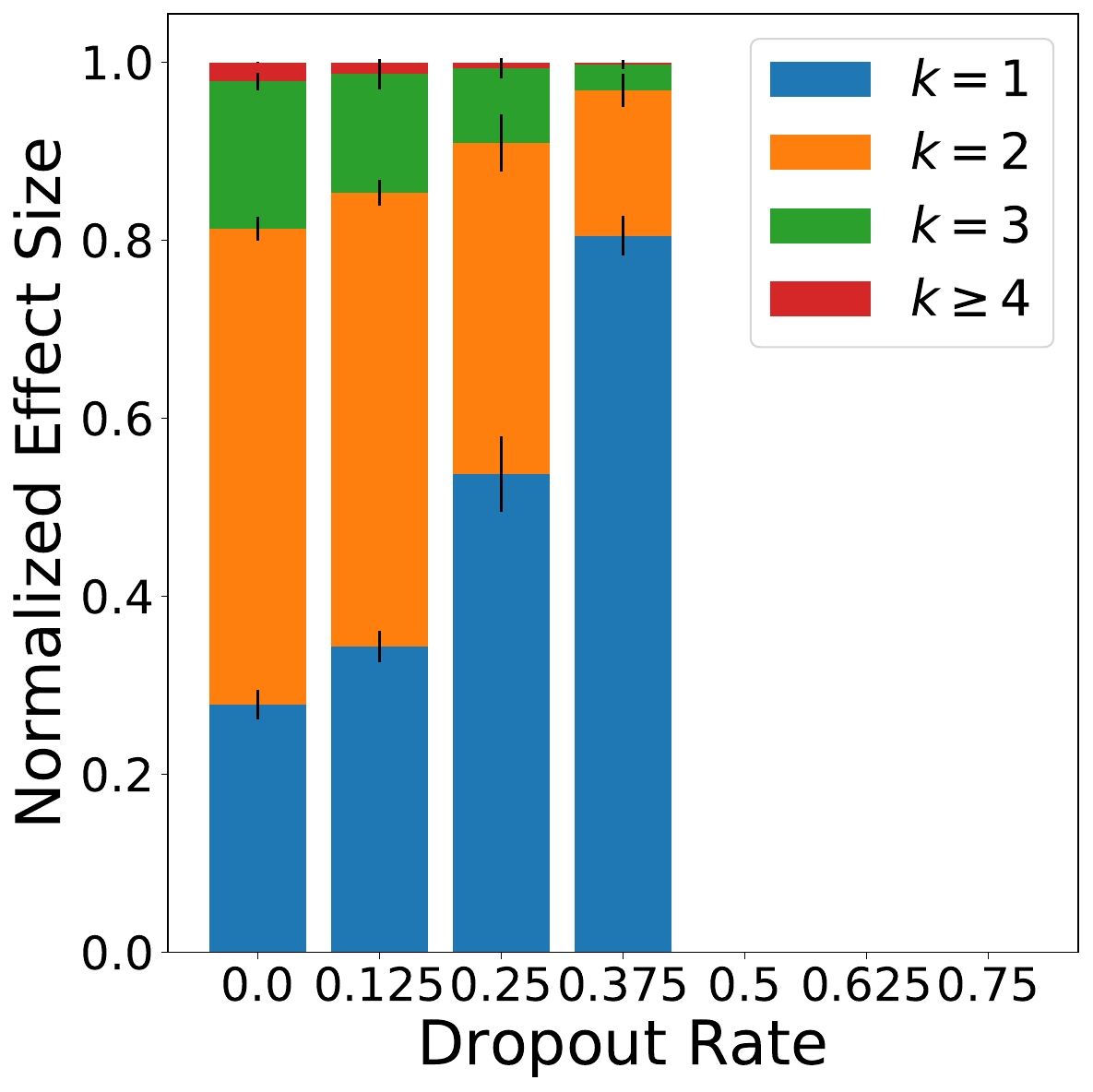}
        \caption{Normalized: Input + Act.
        \label{fig:converged_32:norm_both}}
    \end{subfigure}
    % \\
    % \begin{subfigure}[t]{0.32\textwidth}
    %     \centering
    %     \includegraphics[width=\columnwidth]{fig/converged/intro_fig_decay_32_True_False.pdf}
    %     \caption{Shrinkage: Act. Dropout
    %     \label{fig:converged_32:decay_activation}}
    % \end{subfigure}
    % ~
    % \begin{subfigure}[t]{0.32\textwidth}
    %     \centering
    %     \includegraphics[width=\columnwidth]{fig/converged/intro_fig_decay_32_False_True.pdf}
    %     \caption{Shrinkage: Input Dropout
    %     \label{fig:converged_32:decay_input}}
    % \end{subfigure}
    % ~
    % \begin{subfigure}[t]{0.32\textwidth}
    %     \centering
    %     \includegraphics[width=\columnwidth]{fig/converged/intro_fig_decay_32_True_True.pdf}
    %     \caption{Shrinkage: Input + Act.
    %     \label{fig:converged_32:decay_both}}
    % \end{subfigure}
    \caption{
    NNs trained on pure noise (details in Sec.~\ref{sec:noise_expt}). 
    Displayed values are the mean $\pm$ std. over 10 runs of the effect size for each interaction order. % in the trained model%'s variance explained by each order of interaction. 
    Activation and Input Dropout both reduce the effect sizes of the learned high-order interactions. 
    The top row (a--c) shows absolute effect sizes (which decrease as Dropout increases), % for different values of Dropout --- as Dropout grows, overfitting is reduced and the variance of the predictions converges towards zero. 
    while the middle row (d--f) shows the relative effect sizes% of interactions. 
    , making it easier to see how the Dropout rate affects each order.  %The bottom row (g--i) shows the
    %effects normalized by their strength in the unregularized model to visualize the shrinkage effect of Dropout for each degree of interaction.
    \label{fig:converged_32}}
\end{figure*}

In this experiment, we use a simulation setting in which there is no signal (so any estimated effects are spurious). 
This gives us a testbench to easily see the regularization strength of different levels of Dropout.
Specially, we generate 1500 samples of 25 input features where $X_i \sim \text{Unif}(-1, 1)$ and $Y \sim N(0, 1)$. 
We optimize NNs with 3 hidden layers and ReLU nonlinearities and measure effect sizes as described in Sec.~\ref{sec:intx_nn:measuring}. 
In Fig.~\ref{fig:converged_32}, we see the results for NNs with 32 units in each hidden layer. 
For this small network, both Activation and Input Dropout have strong regularizing effects on a NN. 
Not only do they reduce the overall estimated effect size, both Activation and Input Dropout preferentially target higher-order interactions (e.g., the proportion of variance explained by low-order interactions monotonically increases as the Dropout Rate is increased for Figs.~\ref{fig:converged_32:norm_activation},\ref{fig:converged_32:norm_input}, and \ref{fig:converged_32:norm_both}. 
In Fig.~\ref{fig:converged_128}, we see results from the same experiment on NNs with 128 units in each hidden layer; 
as our analysis predicts, Input Dropout is just as strong for this network (Fig.~\ref{fig:converged_128:norm_input}).

\subsection{Optimal Dropout Rate Depends On True Interactions}
\label{sec:experiments:optimal_rate}
A natural application of this perspective is that Dropout should be used at higher rates where we need to regularize against interaction effects.
To test this guideline, we perform two experiments. 

\paragraph{Modified 20-NewsGroups Data}
We use the 20-NewsGroups dataset \footnote{\url{http://qwone.com/~jason/20Newsgroups/}}, which is a classification task on documents from 20 news organizations. 
We modify this dataset by adding $k$ new features (each feature is IID $\text{Unif}(0,1)$) and a 21st class which is the correct label if all of the $k$ new features take on a value greater than $0.5$. 
This modified dataset then has a strong $k$-way interaction effect, and as $k$ grows, we would expect the optimal Dropout rate to be lower. 
As predicted by our understanding of Dropout, indeed the optimal Dropout rate is lower for larger $k$; with optimal rates of $0.375$ for $k=1$, $0.25$ for $k=2$, and $0.125$ for $k=3$ (full results are shown in Table~\ref{tab:newsgroups}).

\paragraph{BikeShare} The New York City BikeShare dataset\footnote{\url{https://www.citibikenyc.com/system-data}} (preprocessing from \footnote{\url{https://www.kaggle.com/akkithetechie/new-york-city-bike-share-dataset}}) is a large dataset designed to help predict the demand of Citi Bikes in New York City. 
Because bicyclists base travel plans on hourly, daily, and weekly cycles, there are real interaction effects in this dataset \citep{tan2018learning}. 
As predicted by the interaction view of Dropout, the optimal Dropout rate Dropout is 0 (full results in Fig.~\ref{fig:dropout_bikeshare}).

\subsection{Do Other Regularizers Penalize Interaction Effects?}
\label{sec:other_reg}

Here, we examine early stopping and weight decay as potential regularizers of interaction effects.
We find that neither of these regularization techniques specifically target interaction effects. 
However, because Dropout changes the effective learning rate of interaction effects, it can act in concert with early stopping to magnify the regularization against interaction effects. 

\begin{figure*}[t]
    \centering
    \begin{subfigure}[t]{0.23\textwidth}
        \centering
        \includegraphics[width=\textwidth]{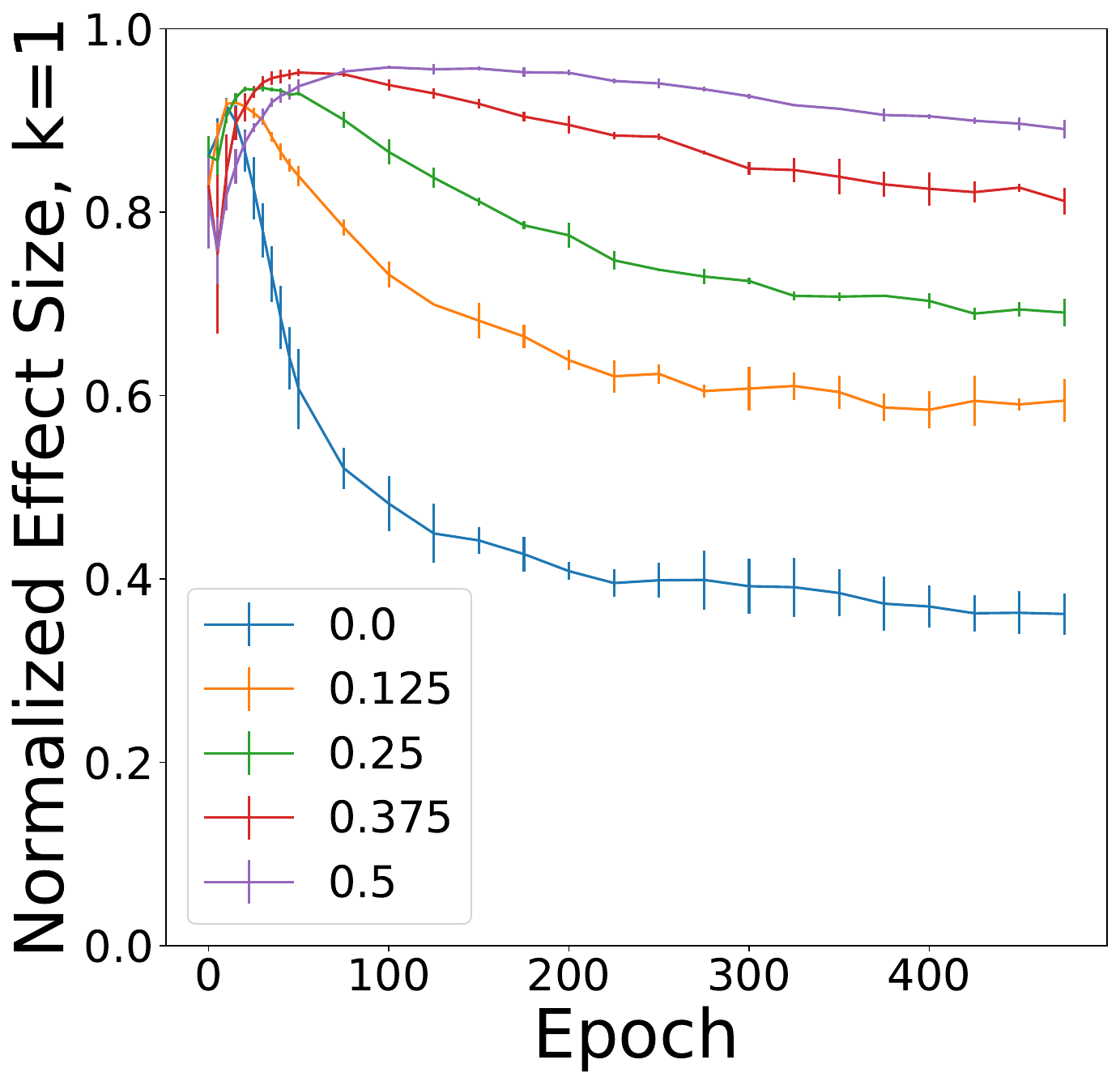}
        \caption{1-way intxs, gen 1}
    \end{subfigure}
    ~
    \begin{subfigure}[t]{0.23\textwidth}
        \centering
        \includegraphics[width=\textwidth]{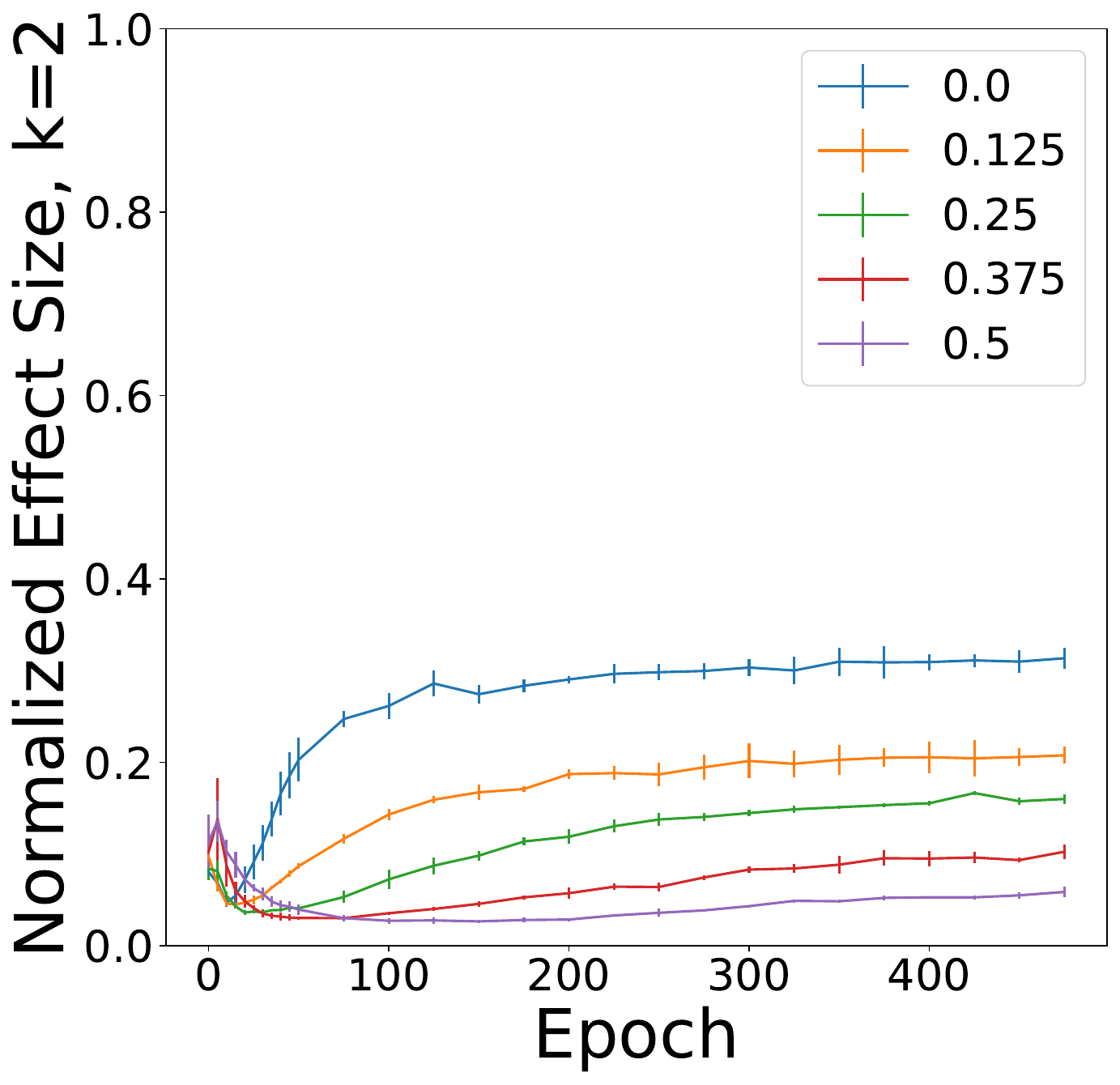}
        \caption{2-way intxs, gen 1}
    \end{subfigure}
    ~
    \begin{subfigure}[t]{0.23\textwidth}
        \centering
        \includegraphics[width=\textwidth]{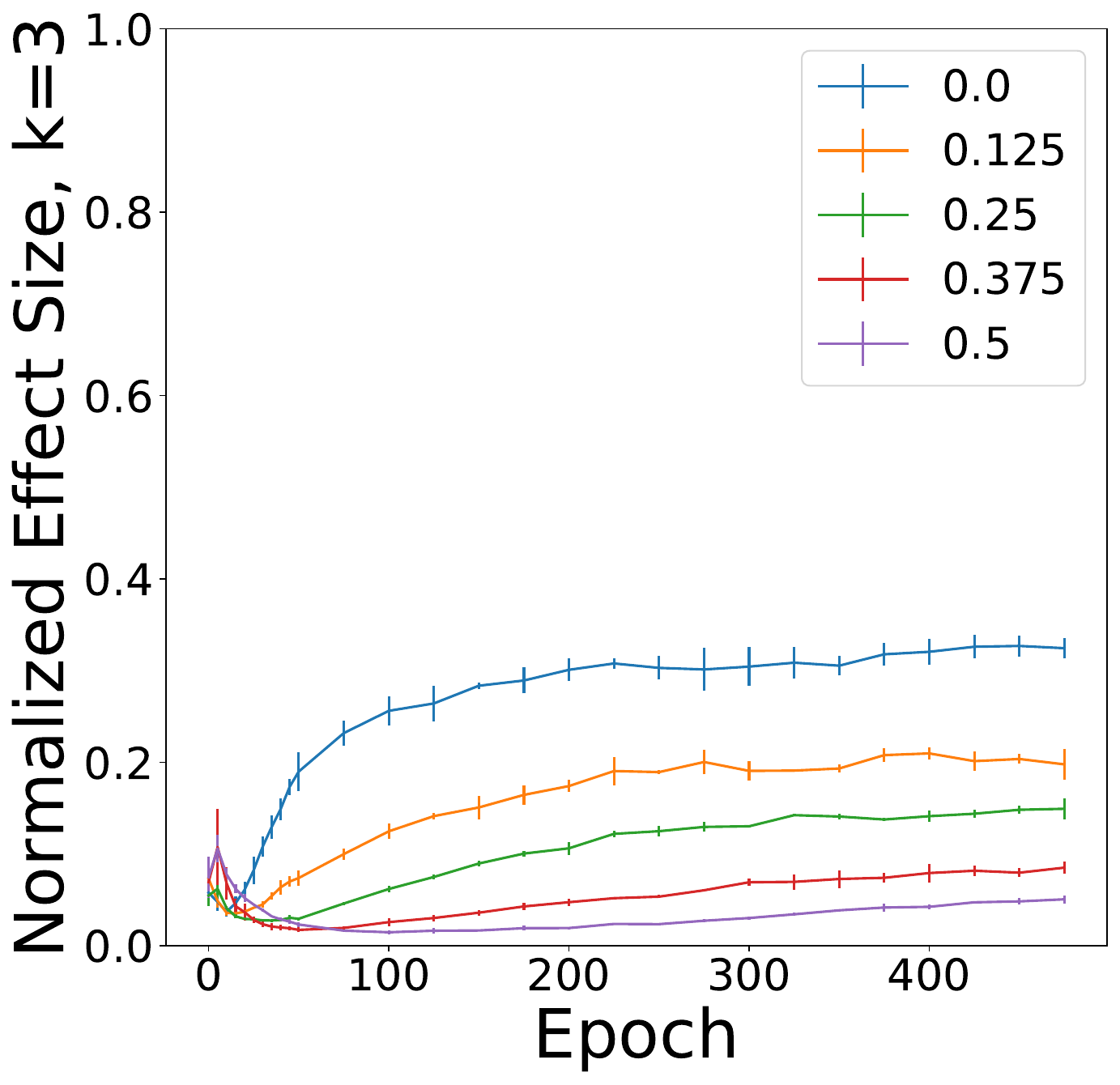}
        \caption{3-way intxs, gen 1}
    \end{subfigure}
    % ~
    % \begin{subfigure}[t]{0.23\textwidth}
    %     \centering
    %     \includegraphics[width=\textwidth]{fig/over_epochs/final_over_epochs_mains_1_mses.pdf}
    %     \caption{MSEs, gen 1}
    % \end{subfigure}
    \\
    \begin{subfigure}[t]{0.23\textwidth}
        \centering
        \includegraphics[width=\textwidth]{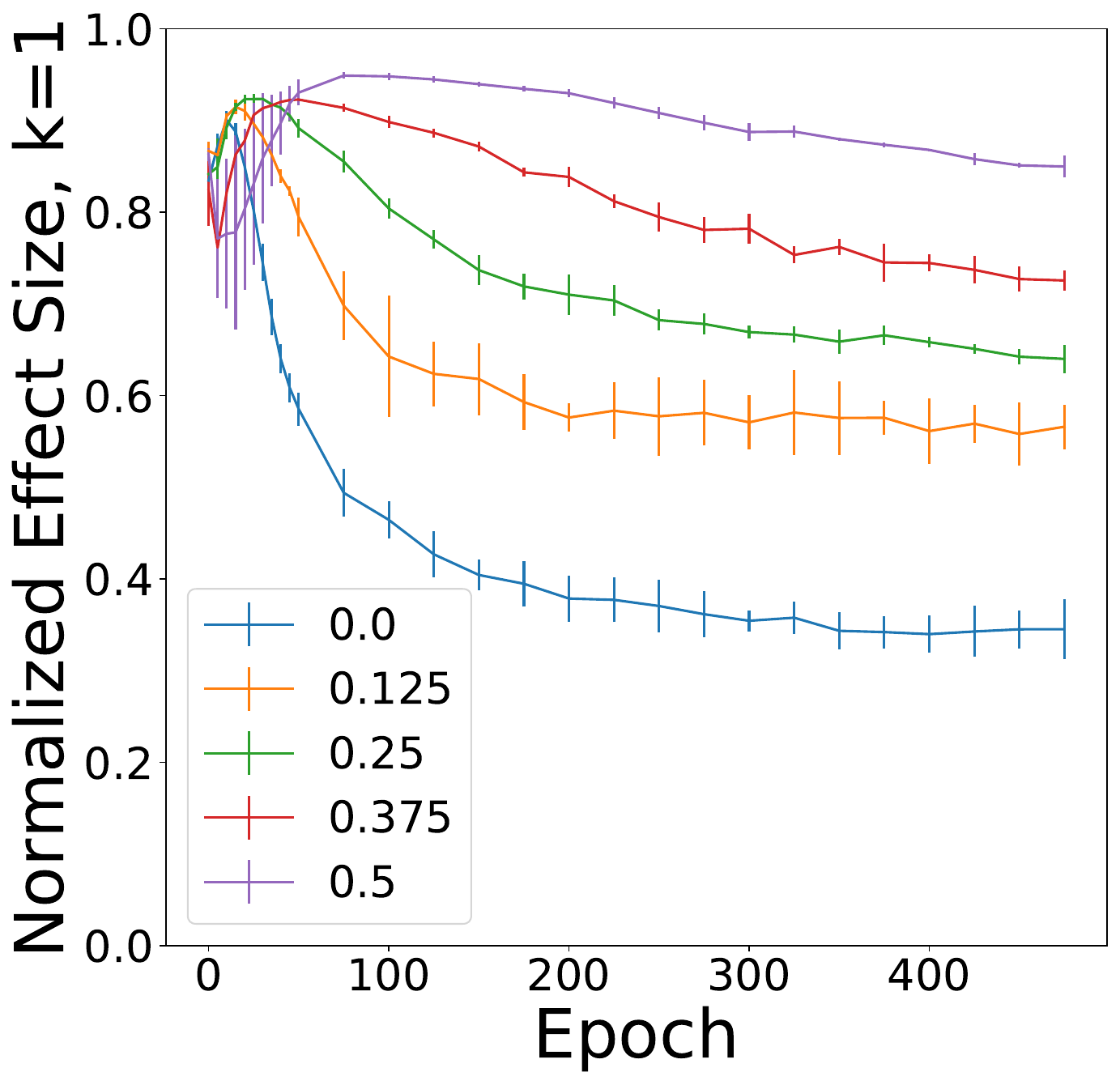}
        \caption{1-way intxs, gen 2}
    \end{subfigure}
    ~
    \begin{subfigure}[t]{0.23\textwidth}
        \centering
        \includegraphics[width=\textwidth]{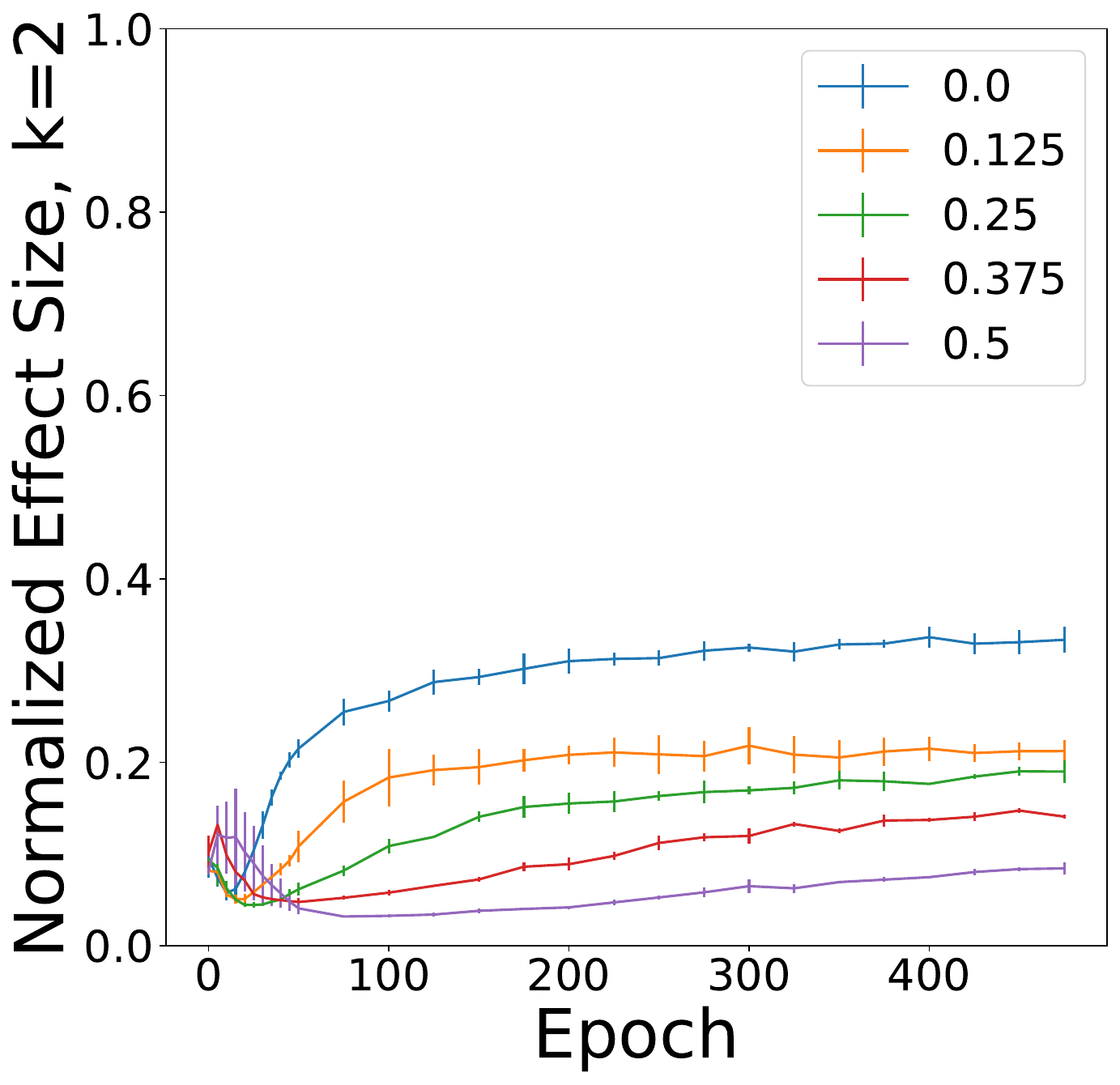}
        \caption{2-way intxs, gen 2}
    \end{subfigure}
    ~
    \begin{subfigure}[t]{0.23\textwidth}
        \centering
        \includegraphics[width=\textwidth]{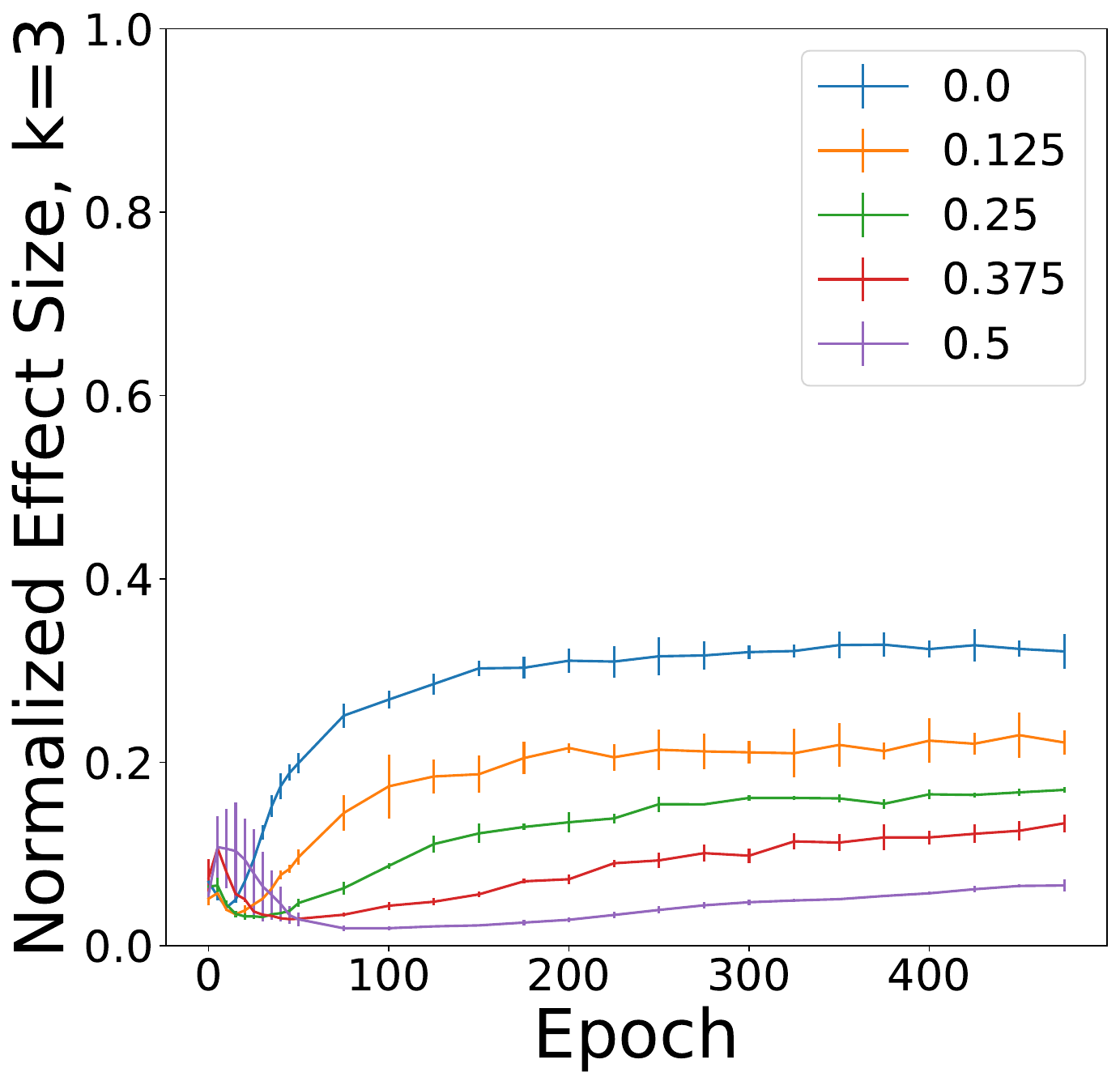}
        \caption{3-way intxs, gen 2}
    \end{subfigure}
    % ~
    % \begin{subfigure}[t]{0.23\textwidth}
    %     \centering
    %     \includegraphics[width=\textwidth]{fig/over_epochs/final_over_epochs_mains_0_mses.pdf}
    %     \caption{MSEs, gen 2}
    % \end{subfigure}
    \\
    \begin{subfigure}[t]{0.23\textwidth}
        \centering
        \includegraphics[width=\textwidth]{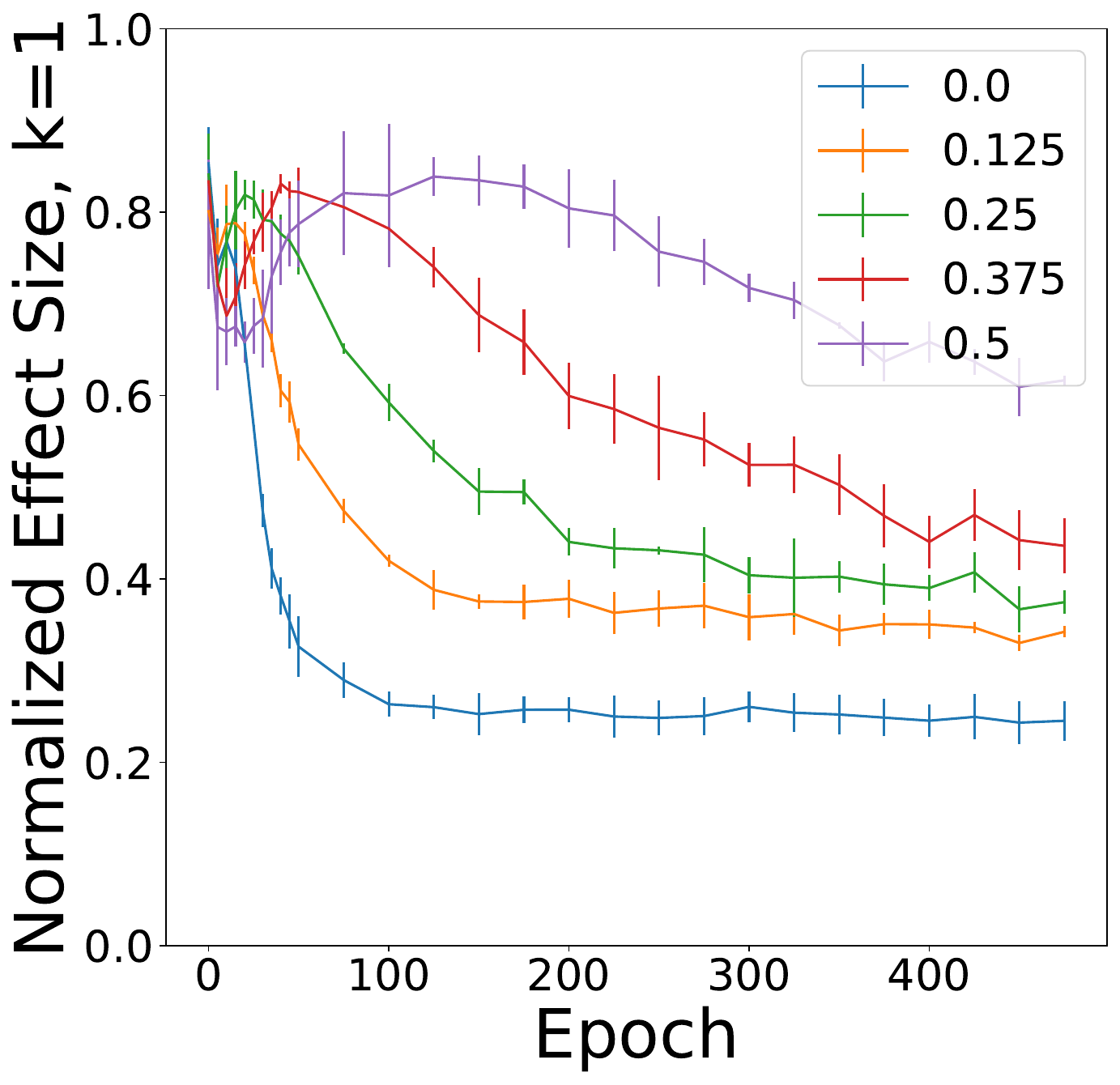}
        \caption{1-way intxs, gen 3}
    \end{subfigure}
    ~
    \begin{subfigure}[t]{0.23\textwidth}
        \centering
        \includegraphics[width=\textwidth]{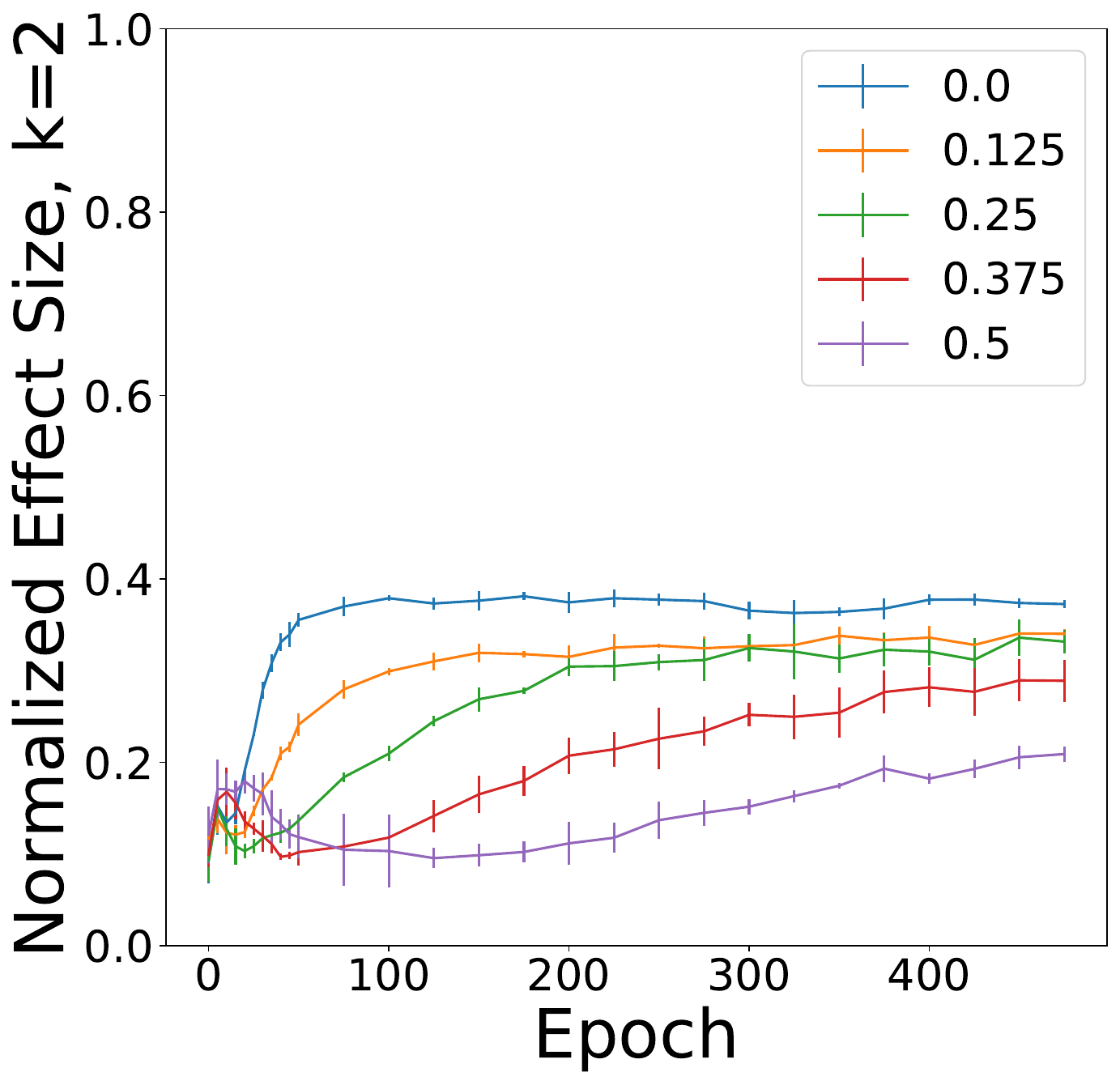}
        \caption{2-way intxs, gen 3}
    \end{subfigure}
    ~
    \begin{subfigure}[t]{0.23\textwidth}
        \centering
        \includegraphics[width=\textwidth]{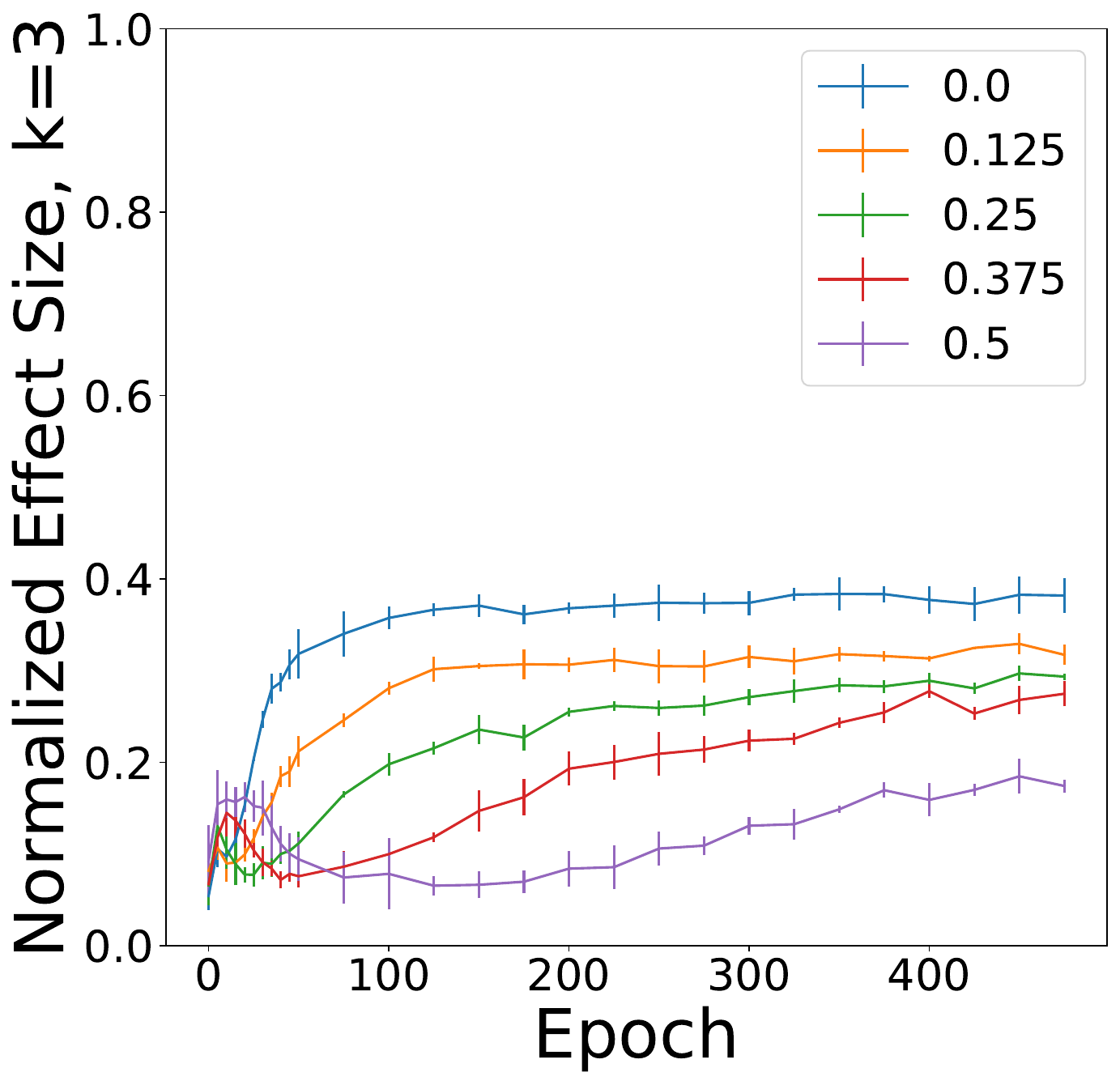}
        \caption{3-way intxs, gen 3}
    \end{subfigure}
    % ~
    % \begin{subfigure}[t]{0.23\textwidth}
    %     \centering
    %     \includegraphics[width=\textwidth]{fig/over_epochs/over_epochs_threes_mses.pdf}
    %     \caption{MSEs, gen 3}
    % \end{subfigure}
    \caption{
    Learned interaction effects of order 1, 2 and 3 (cols 1, 2, and 3 respectively) by epoch. %, and model error on train and test (col 4) vs. epochs.
    Each row corresponds to a different generator as described in Sec.~\ref{sec:other_reg:early_stopping}: the generator in the top row has only 1-way interactions, the generator in the middle row has only 2-way interactions, and the bottom row has only true 3-way interactions. Key findings are described in Sec.~\ref{sec:other_reg:early_stopping}. 
    \label{fig:dropout_over_epochs}}
\end{figure*}

\paragraph{Early Stopping}
\label{sec:other_reg:early_stopping}
The effective capacity of NNs increases during training \citep{weigend1994overfitting}, and recent work supports the view that randomly-initialized NNs start as simple functions that are made more complex through training \citep{de2018deep,nakkiran2019sgd,NIPS2018_8076}. 
Thus, it makes sense that early stopping can help select models that generalize well \citep{prechelt1998early,caruana2001overfitting}. 
To see how early stopping interplays with the Dropout-induced effective learning rates, we study the effects learned over the course of optimization. 

We generate 1500 samples of 25 input features where $X_i \sim \text{Unif}(-1, 1)$ and the target is generated according to one of three settings: (1) only main effects: $Y \sim \text{N}(\sin(X_0) + \cos(X_1), \sigma^2)$, (2) only pair effects: $Y \sim \text{N}(\sin(X_0)\cos(X_1), \sigma^2)$, and (3) only three-way effects: $Y \sim$ $\text{N}(\sin(X_0)\cos(X_1)X_2, \sigma^2)$. 
We optimize fully-connected NNs on these data 
and measure effect sizes as described in Sec.~\ref{sec:intx_nn:measuring}. 
Results are shown in Fig.~\ref{fig:dropout_over_epochs}. 
The key findings are: 1) the rightmost column shows that NNs with low rates of Dropout tend to massively overfit due to a reliance on high-order interactions; 2) the different levels of Dropout have different steady-state optima; 3) because Dropout slows the learning of high-order effects, early stopping is doubly effective in combination with Dropout.
NNs tend to learn simple functions earlier (regardless of Dropout usage), and Dropout slows the learning of high-order interactions; these factors combine to reduce the complexity of the learned function under early stopping.

\paragraph{Weight Decay}
\label{sec:other_reg:weight_decay}
\begin{figure}[t]
    \centering
    \begin{subfigure}[t]{0.3\columnwidth}
        \centering
        \includegraphics[width=\columnwidth]{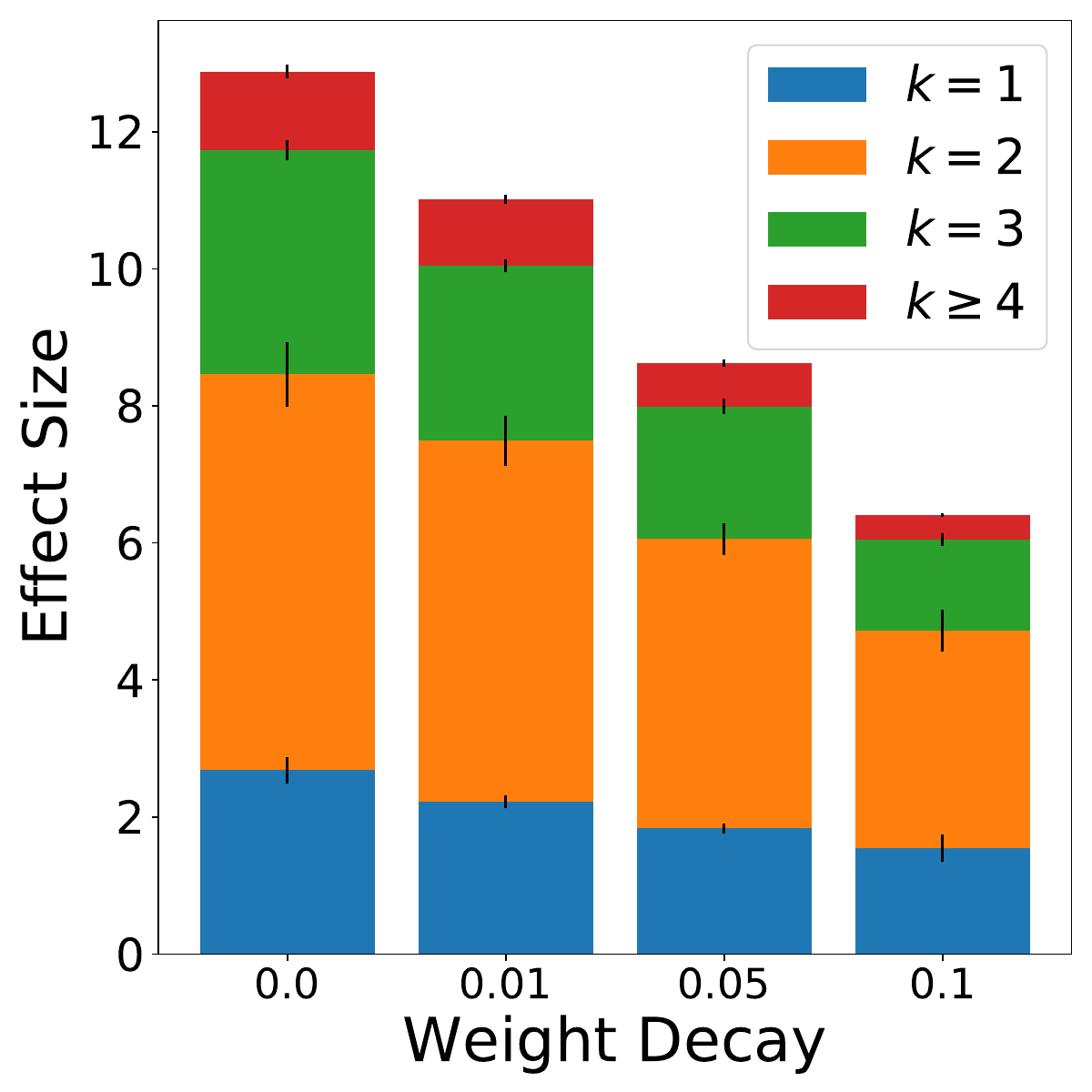}
        \caption{Total Effects
        \label{fig:weight_decay:total}}
    \end{subfigure}
    ~
    \begin{subfigure}[t]{0.3\columnwidth}
        \centering
        \includegraphics[width=\columnwidth]{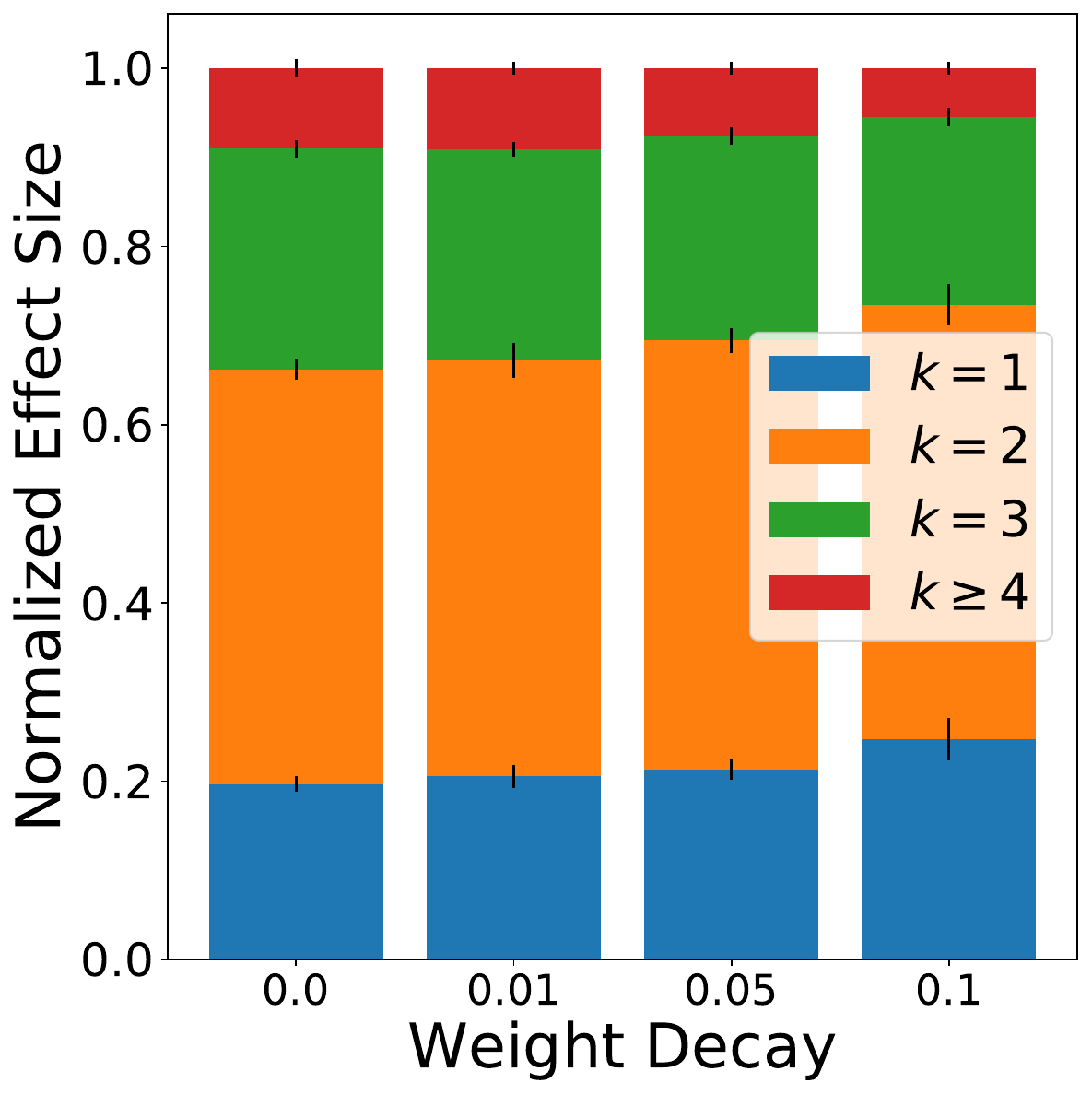}
        \caption{Normalized Effects
        \label{fig:weight_decay:norm}}
    \end{subfigure}
    ~
    \begin{subfigure}[t]{0.3\columnwidth}
        \centering
        \includegraphics[width=\columnwidth]{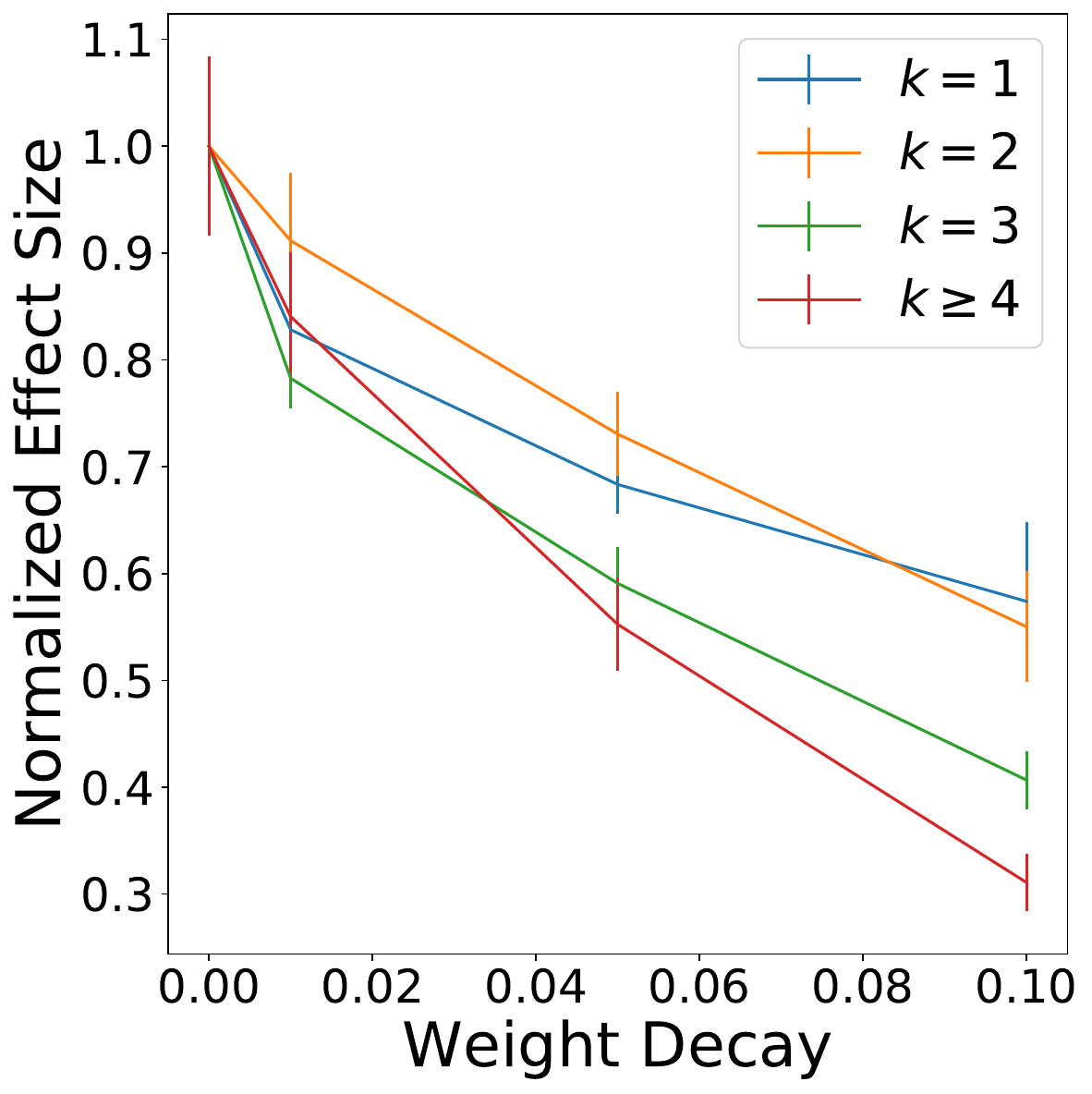}
        \caption{Rate of Decay
        \label{fig:weight_decay:rate}}
    \end{subfigure}
    \caption{Weight decay can weakly regularize against interactions; however, the regularization comparable to Dropout occurs at extremely strong weight decay for which training is unstable.
    \label{fig:weight_decay}}
\end{figure}

Another popular regularization mechanism is weight decay: placing an $\ell_2$ penalty on the weights of the network. 
We study weight decay on the same data generator as we studied Dropout in Sec.~\ref{sec:noise_expt}. 
As the results in Fig.~\ref{fig:weight_decay} show, strong weight decay (large values of $\lambda$) has a modest effect of regularizing against interaction effects. 
However, achieving the same practical benefit from weight decay as from Dropout is untenable due to the training instability that strong weight decay introduces: when weight decay was larger than 0.2, the NNs learned constant functions.  

\section{Discussion and Implications}
\label{sec:discussion}
In this paper, we examined a concrete mechanistic explanation of how Dropout works: by regularizing higher-order interactions. 
This explanation of Dropout has several implications for its use and crystallizes some of the conventional wisdom regarding how and when to use Dropout.

\subsection{Dropout For Explanations}
\begin{figure}[htb]
    \centering
    \begin{subfigure}[b]{0.45\textwidth}
        \centering
        \includegraphics[width=0.45\textwidth]{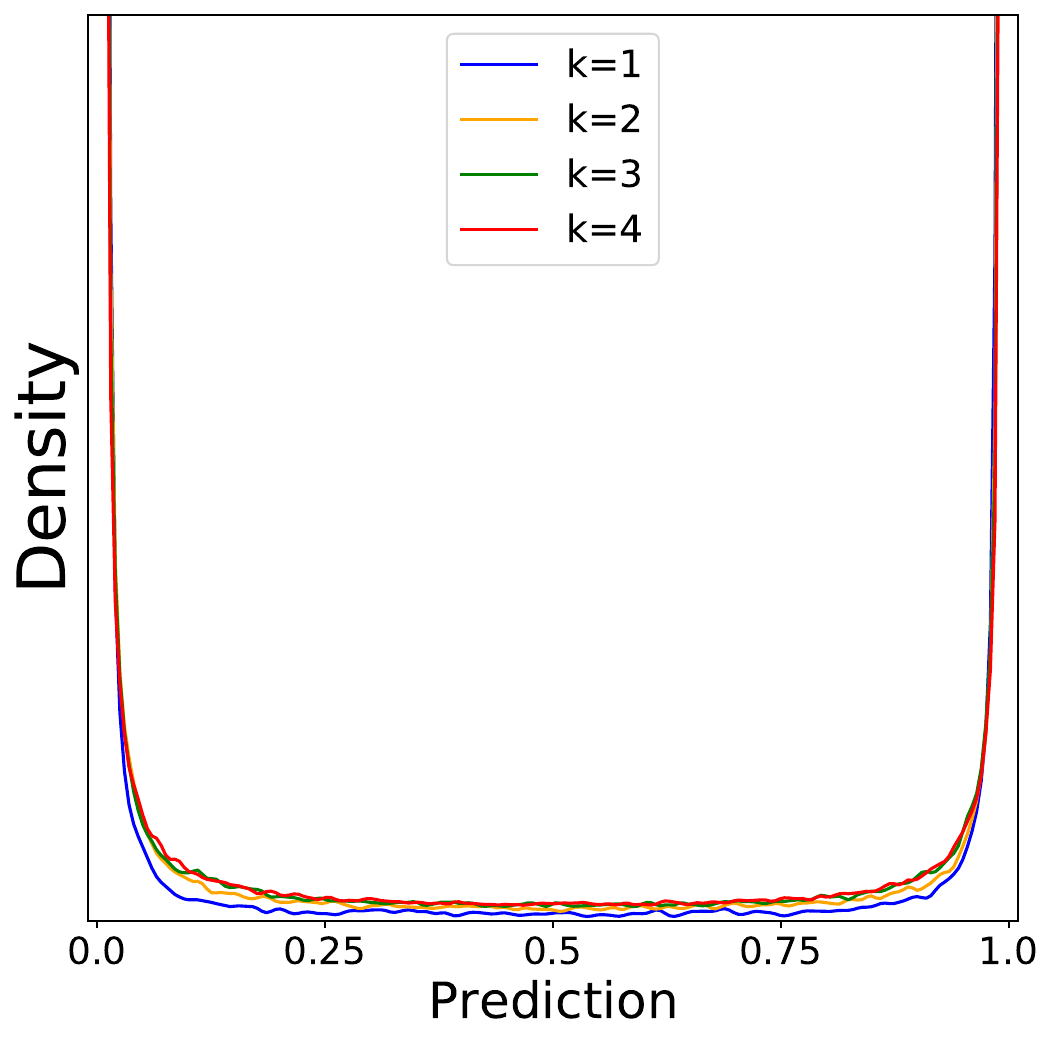}
        ~
        \includegraphics[width=0.45\textwidth]{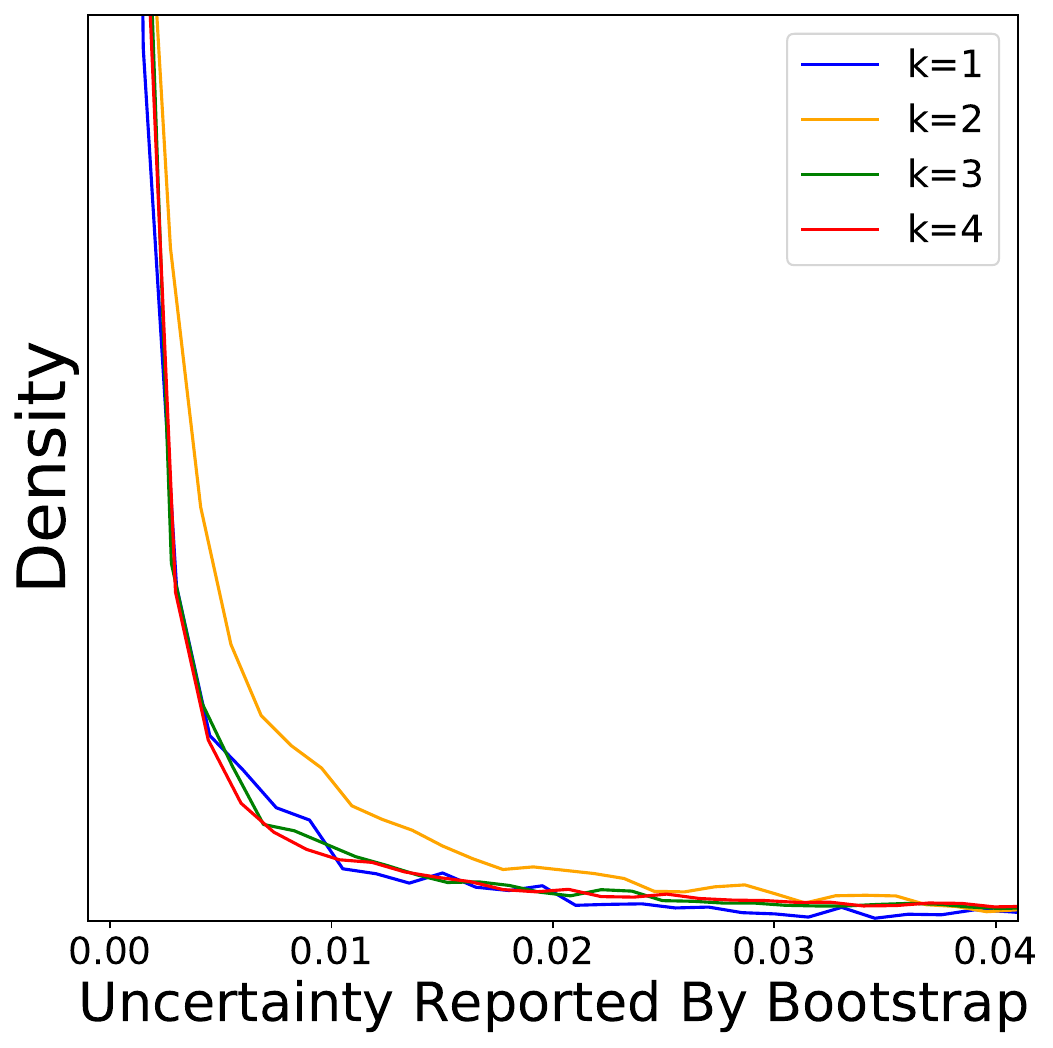}
        \caption{Histogram of predictions (left) and uncertainties (right) without Dropout}
        \label{fig:preds_no_dropout}
    \end{subfigure}
    ~
    \begin{subfigure}[b]{0.45\textwidth}
        \centering
        \includegraphics[width=0.45\textwidth]{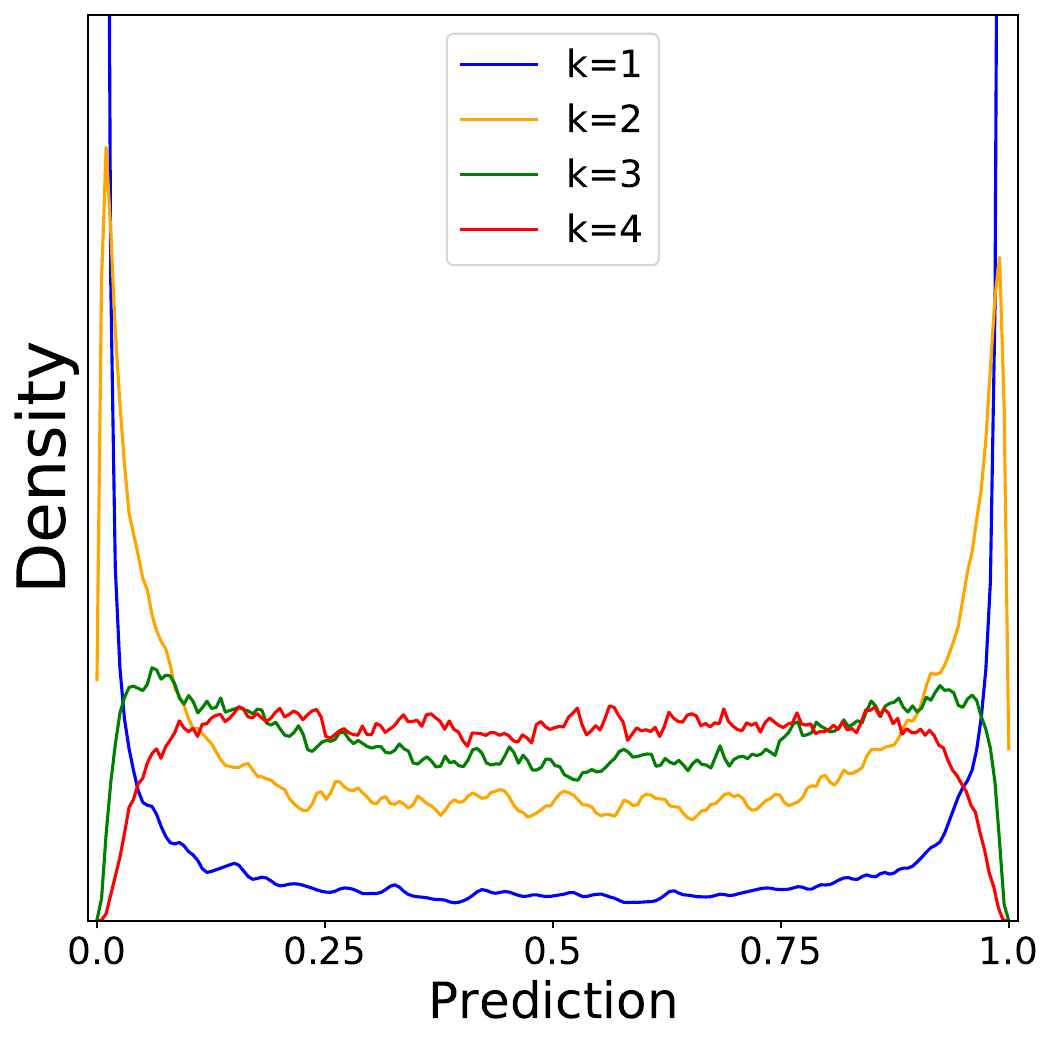}
        ~
        \includegraphics[width=0.45\textwidth]{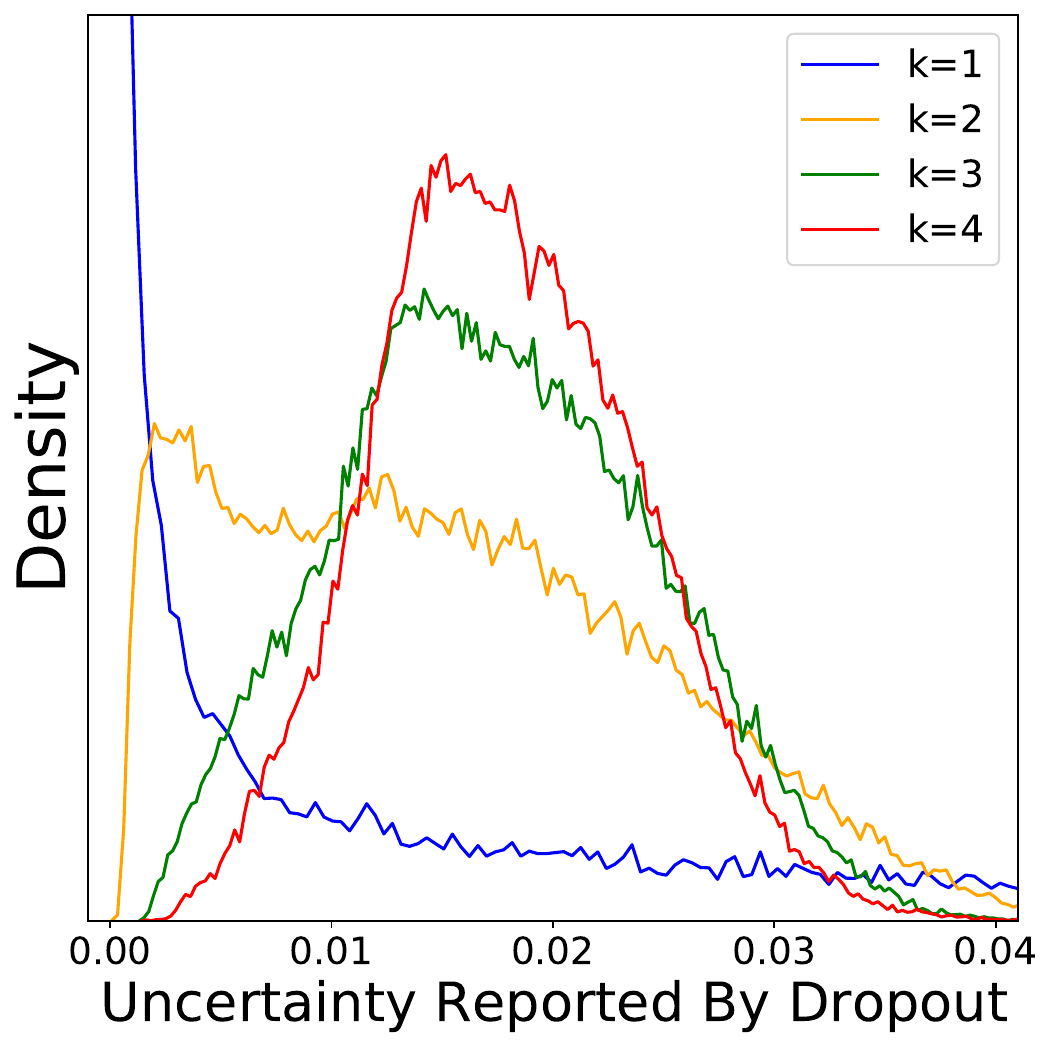}
        \caption{Histogram of predictions (left) and uncertainties (right) with Dropout at $p=0.25$}
        \label{fig:preds_dropout}
    \end{subfigure}
    \caption{Visualizing the predictions and uncertainties measured by bootstrap (a) and Dropout (b). Dropout targets higher-order interaction effects, leading to a bias in the reported uncertainties. Uncertainty is defined as the variance of the prediction over all Dropout masks. 
    \label{fig:uncertainty}}
\end{figure}

While Dropout has been used for measures of model confidence \citep{gal2016dropout,NIPS2017_6949} and to aid model interpretability \citep{chang2017dropout,chang2017interpreting}, it does not treat all effects equally. 
This bias is present both during estimation and inference. We examine the distributions of predictions and uncertainties produced by NNs under various Dropout rates (Fig.~\ref{fig:uncertainty}). 
In this experiment, we train NNs to predict a random variable which is the product of $k$ uncorrelated Bernoulli variables.  
We generate sufficient samples that all NNs learn to confidently predict the outcome for all orders (Fig.~\ref{fig:preds_no_dropout}). 
However, when Dropout is used (Fig.~\ref{fig:preds_dropout}), the models equivocate, with much greater uncertainty for the higher-order interaction effects -- the model representing an interaction of order 1 is barely affected, while the model representing an interaction of order 4 is ambivalent. % for all predictions. 
Thus, the bias of Dropout is reflected when measuring model confidence. % of predictions. 

\subsection{Setting Dropout Rate}
\label{sec:discussion:setting}
The Dropout rate should be set according to the desired magnitude of the anti-interaction regularization effect. 
If the dataset is large or sufficient augmentation can be performed, lower rates of Dropout can be used or Dropout can be omitted entirely (e.g. the New York City BikeShare dataset discussed in Section~\ref{sec:experiments:optimal_rate}). 
In addition, it is often suggested to use larger Dropout rates in deeper layers than in initial layers \citep{NIPS2013_5032}. 
This wisdom can be explained from the interaction perspective: this regularization scheme encourages NNs to do representation learning, which may require learning interactions between input features such as pixels or words, in the initial layers, while encouraging deeper layers to focus more on summing evidence from multiple sources. % rather than learning complex high-order interactions of the learned representation.

In CNNs, Dropout is typically used at lower rates than in fully-connected networks \citep{park2016analysis}. 
The convolutional architecture creates constraints that prevent arbitrary high-order interactions by restricting $N$ in ${N \choose k}$ to be a carefully selected set of local input features or hidden unit activations. 
Operators like max pooling further restrict the model's ability to learn complex interactions. 
In other words, convolutional nets create a strong architectural bias for or against different kinds of interaction effects and thus depend less on a mechanism like Dropout to blindly regularize interactions. 

\subsection{Explicitly Modeling Interaction Effects}
\label{sec:discussion:explicit}
A major challenge of estimating interaction effects is the hypothesis space which grows exponentially with the order of the interaction effect. 
If we were able to reduce the hypothesis space by specifying a small set of potential interaction effects a priori, our models could efficiently learn the correct parameters for these few interactions from data. 
Several recent works have proposed to do this by explicitly specifying the interaction effects the NNs may consider. 
Of particular note is \citep{jayakumar2020multiplicative}, which proposed to use multiplicative interactions to combine data modalities, and found that many common architectures can be seen in the lens of multiplicative interactions.

Another approach to explicitly model interaction effects is the Deep and Cross Network \citep{wang2017deep}, which uses a two-part architecture consisting of a fully-connected network and a ``cross" network in which each layer has its activation crossed with the vector of input variables, % before being transmitted to the next layer. 
increasing the interaction order at every layer. 
Interestingly, the experiments of \citep{wang2017deep} (especially Fig.~3 within) show that the best-performing architecture has only a single cross layer -- exactly what we would expect from the amount of spurious interaction effects NNs are capable of learning.

\subsection{Limitations and Broader Impacts}
This new perspective on Drooput as a regularizer of interactions effects helps to crystallize Dropout use cases and guidelines, but is in no way a full picture of NN behavior. 
There are many reasons why over-parameterized models such as deep NNs generalize to unseen data; here we have explored only one of the contributing factors of a regularizer against spurious interaction effects. 
Nevertheless, we believe that theoretical insights and concise description of NN behaviors, such as this perspective on Dropout, can provide broader impacts driven by more precise descriptions of machine learning behavior. 
Without precise theoretical understanding, significant resources must be invested to hyperparameter tuning and architecture tweaking, tending to favor the adoption of machine learning technologies in large institutions; better theoretical understandign can direct system design and reduce burdensome resource requirements.

\section{Conclusions}
\label{sec:conclusions}
In this paper, we have examined a concrete explanation of Dropout as a regularization against interaction effects. 
We have shown that the effective learning rate of interaction effects decreases exponentially with the order of the interaction effect, a crucial balance against the exponentially-growing number of potential interactions of $k$ variables. 
Input Dropout targets interactions of $k$ input variables, while Activation Dropout targets interactions of $k$ hidden units in a layer. 
Although Dropout can work in concert with weight decay and early stopping, these do not naturally achieve Dropout's regularization against high-order interactions. 
By reducing the tendency of NNs to learn spurious high-order interaction effects, Dropout helps to train models which generalize more accurately to test sets.

\subsubsection*{Acknowledgements}
We thank Chun-Hao Chang, Geoffrey Hinton, Chris Lengerich, and Ruoxi Wang for helpful discussions.
This work was started during an internship at Microsoft Research. 
BL was funded in part by the CMLH Fellowship.

%\subsubsection*{References}
%\vspace{-0.1cm}

%{\small \bibliography{dropout}}
%\bibliographystyle{abbrvnat}

\appendix
%\counterwithin{figure}{section}
%\counterwithin{table}{section}

\cleardoublepage

%\appendix
\onecolumn
\renewcommand\thefigure{\thesection.\arabic{figure}}
\setcounter{figure}{0}

\section{Interaction Effects}

\begin{figure*}[tb]
    \centering
    \begin{subfigure}{\textwidth}
        \centering
        \begin{minipage}{0.2\textwidth}
            \includegraphics[width=\textwidth]{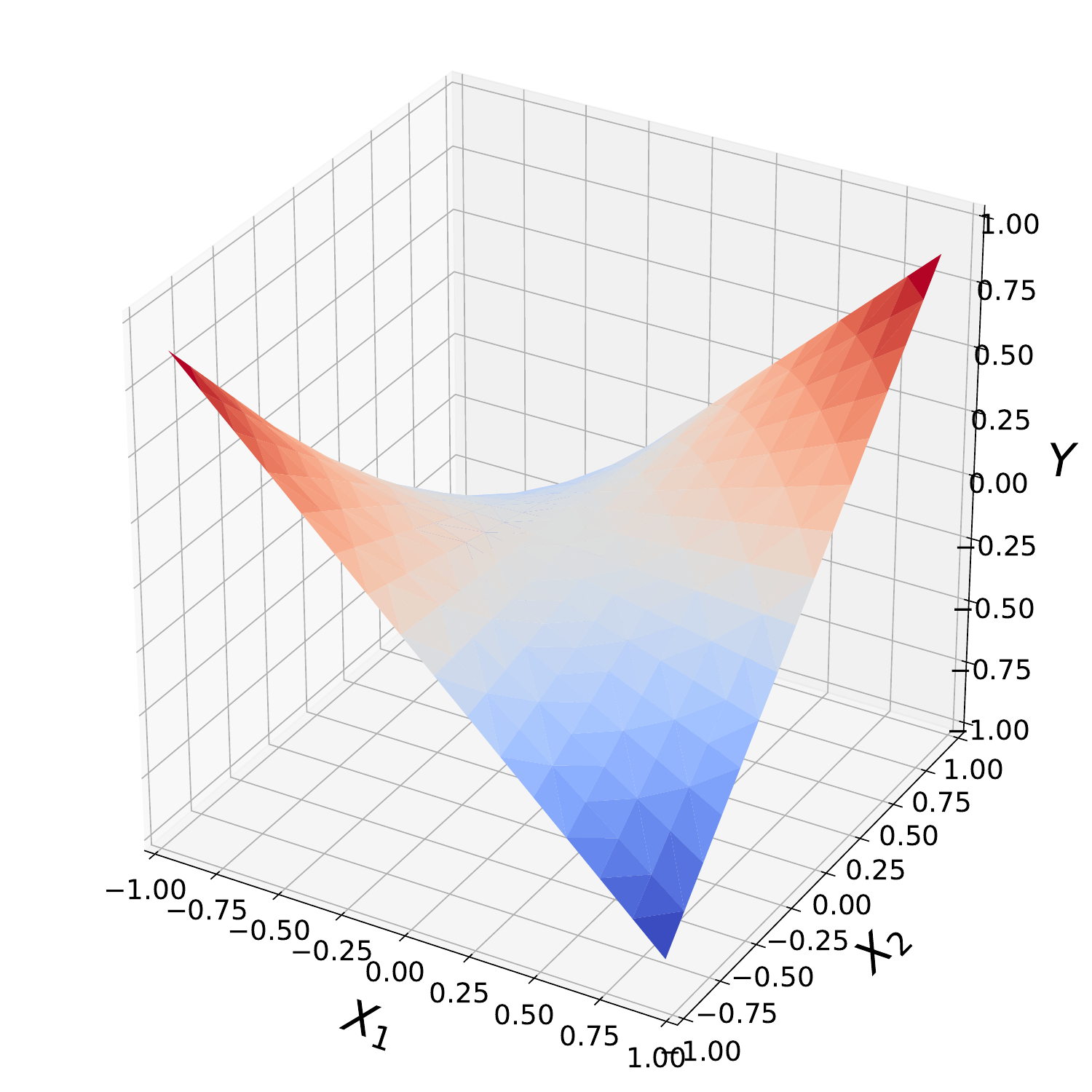}
        \end{minipage}\
        {\LARGE{=}}
        \begin{minipage}{0.2\textwidth}
            \includegraphics[width=\textwidth]{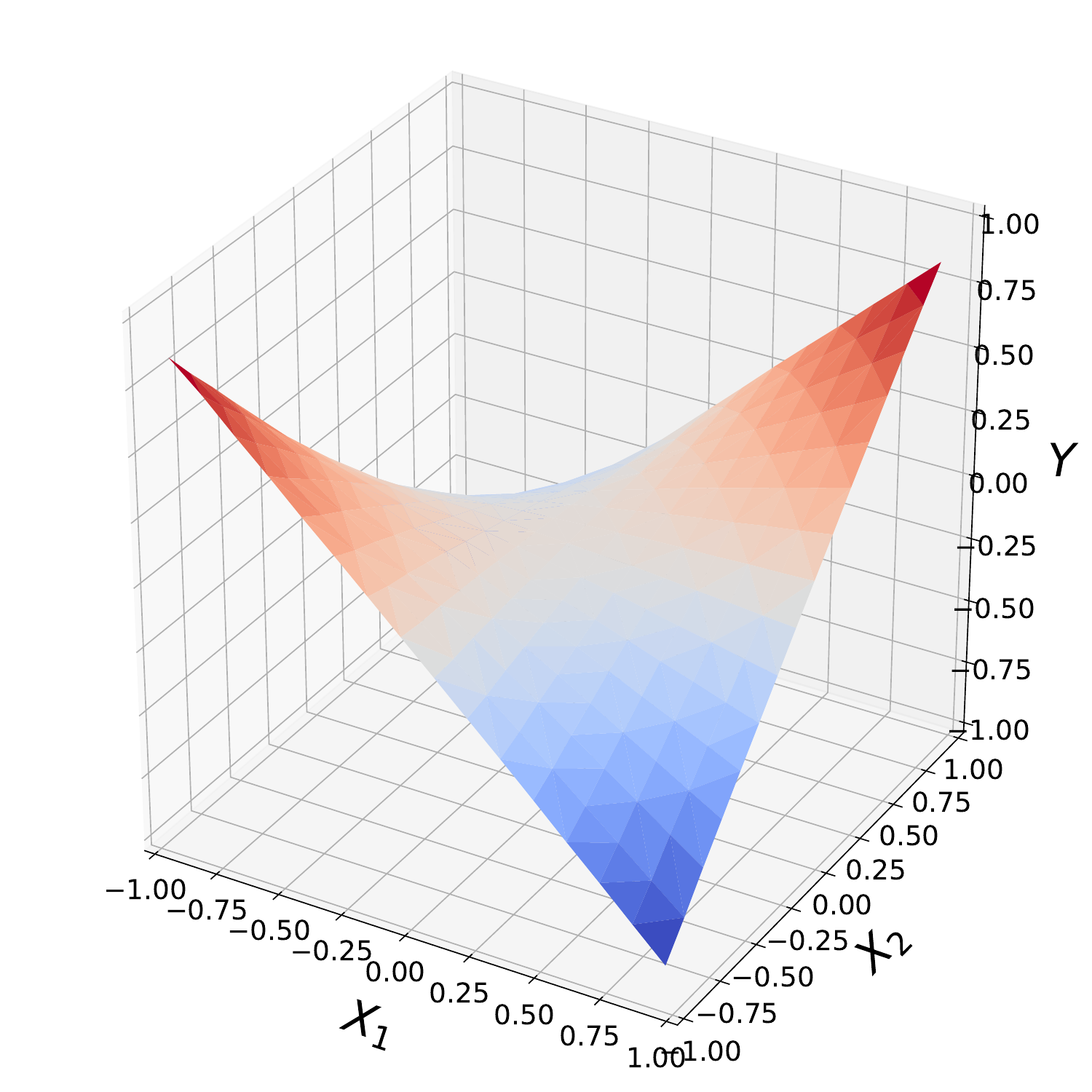}
        \end{minipage}\
        {\LARGE{+}}
        \begin{minipage}{0.2\textwidth}
            \includegraphics[width=\textwidth]{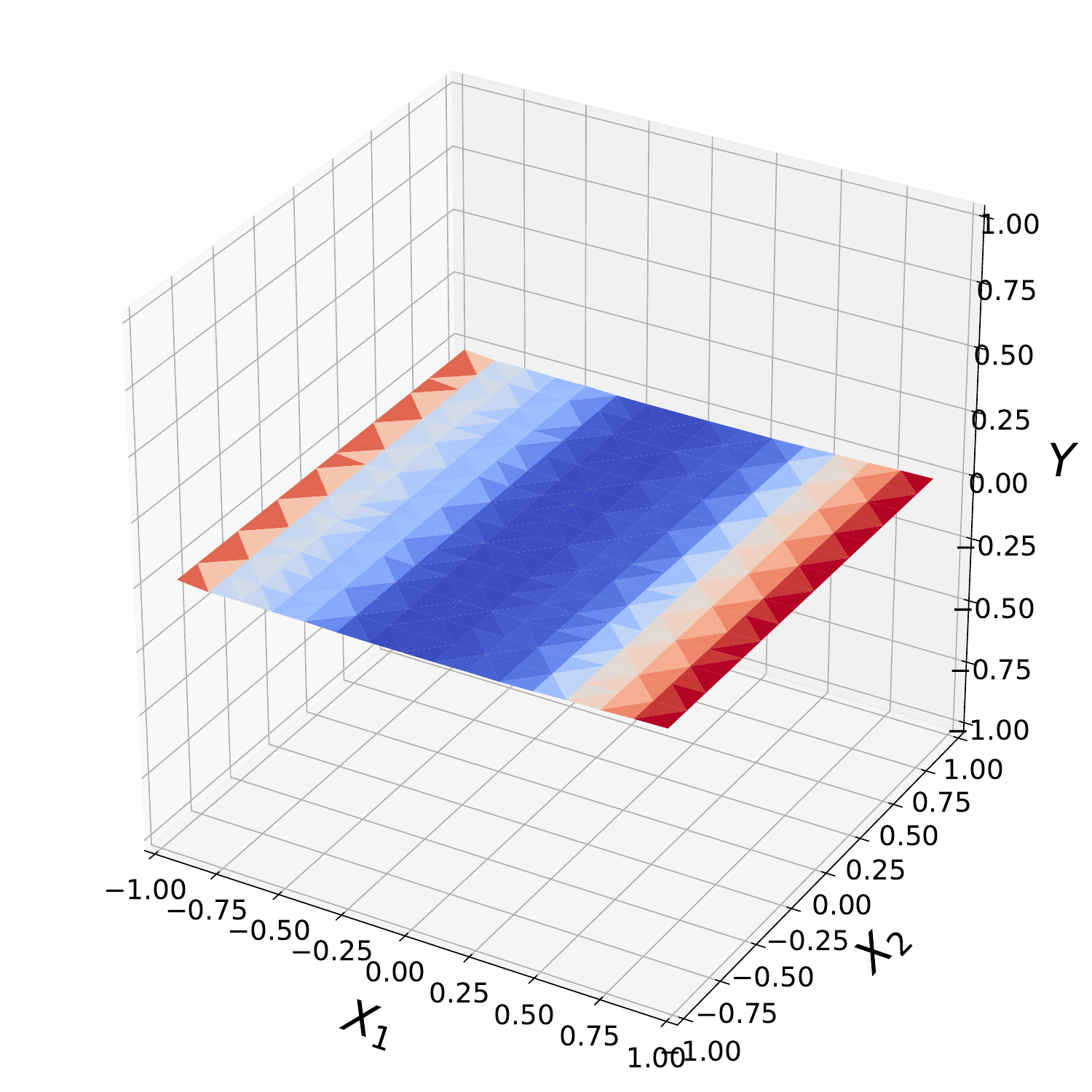}
        \end{minipage}\
        {\LARGE{+}}
        \begin{minipage}{0.2\textwidth}
            \includegraphics[width=\textwidth]{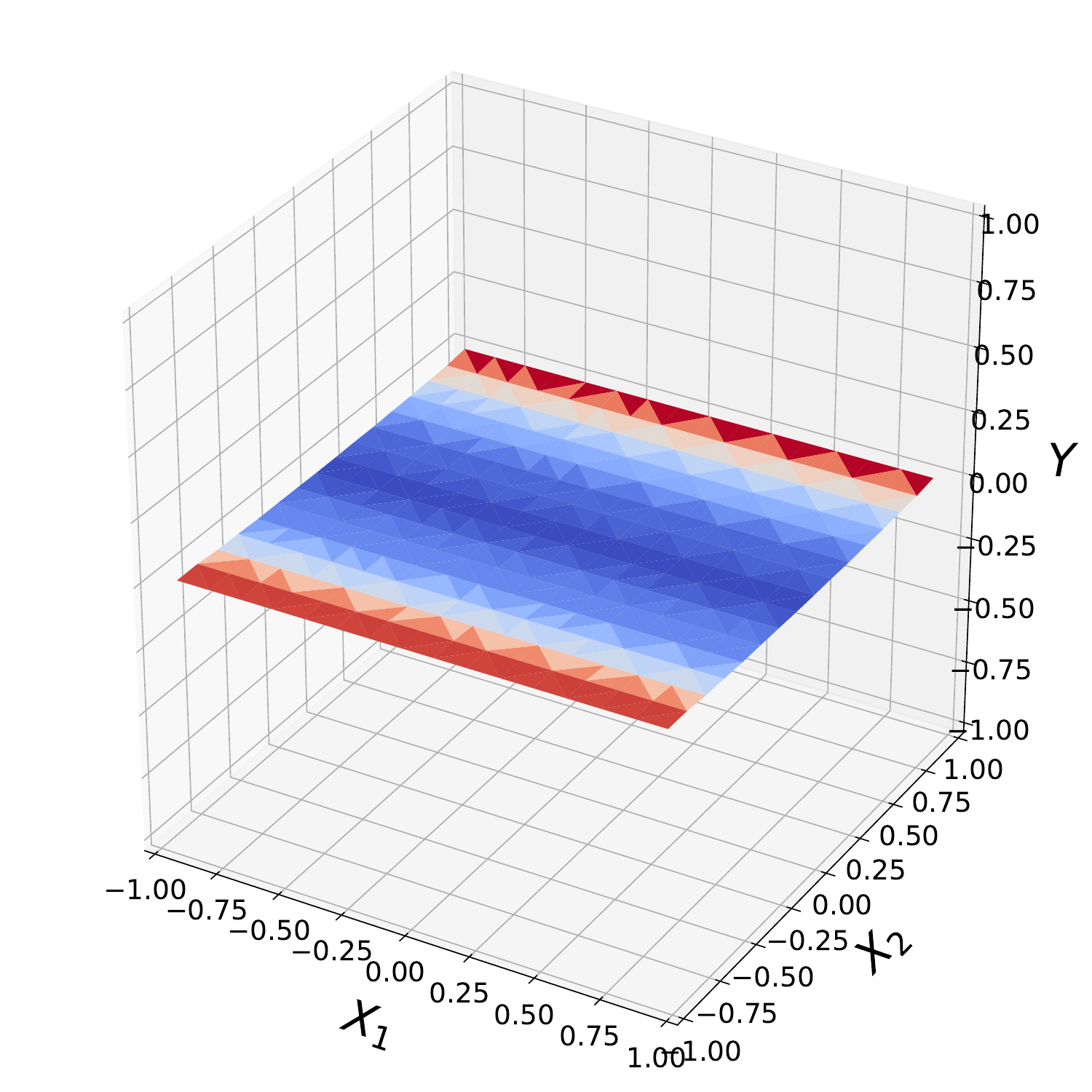}
        \end{minipage}\
        \caption{$\rho = 0.01$ \label{fig:toy_intx:rho0}}
    \end{subfigure}
    \\
    \begin{subfigure}{\textwidth}
        \centering
        \begin{minipage}{0.2\textwidth}
            \includegraphics[width=\textwidth]{fig/interactions/intx_raw.pdf}
        \end{minipage}\
        {\LARGE{=}}
        \begin{minipage}{0.2\textwidth}
            \includegraphics[width=\textwidth]{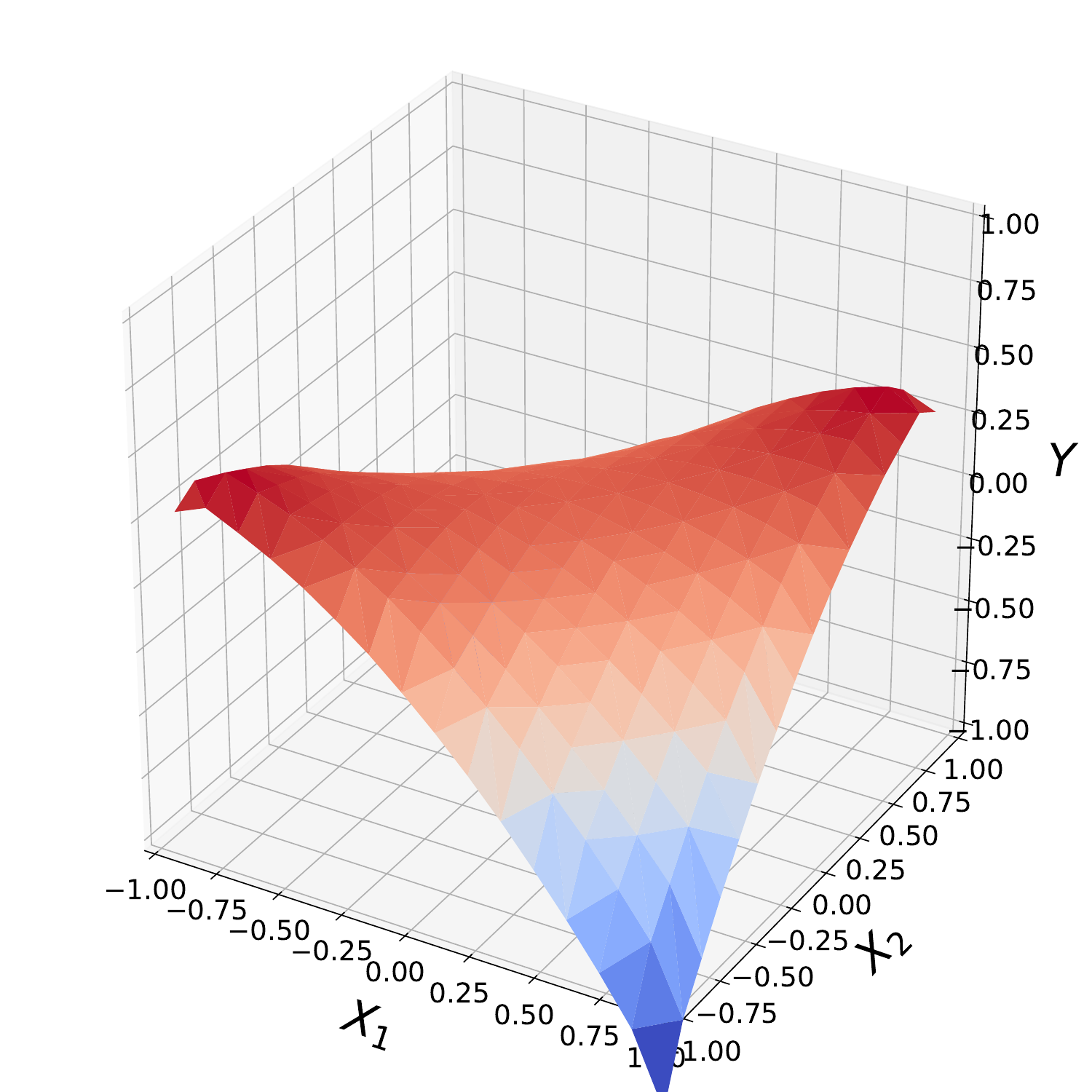}
        \end{minipage}\
        {\LARGE{+}}
        \begin{minipage}{0.2\textwidth}
            \includegraphics[width=\textwidth]{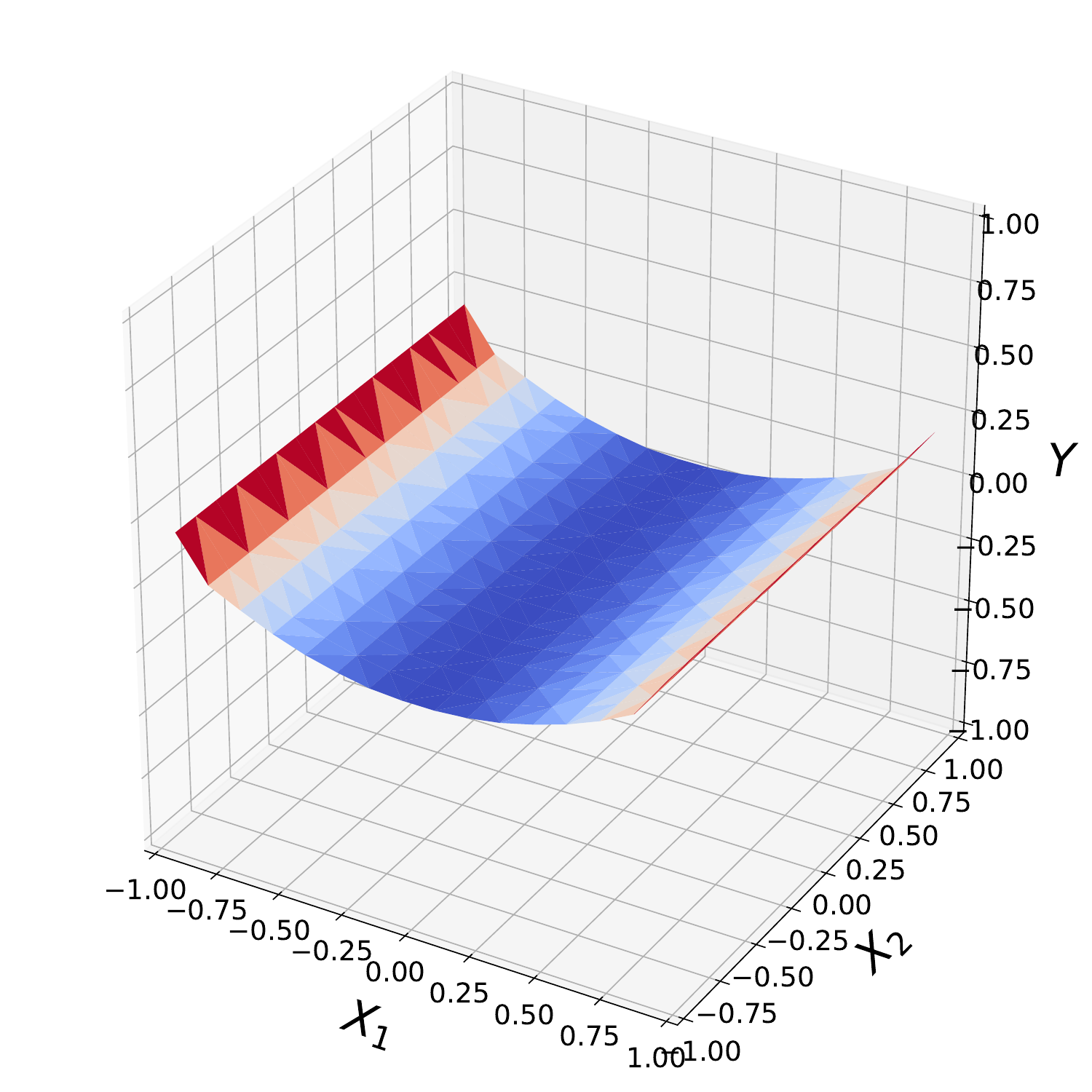}
        \end{minipage}\
        {\LARGE{+}}
        \begin{minipage}{0.2\textwidth}
            \includegraphics[width=\textwidth]{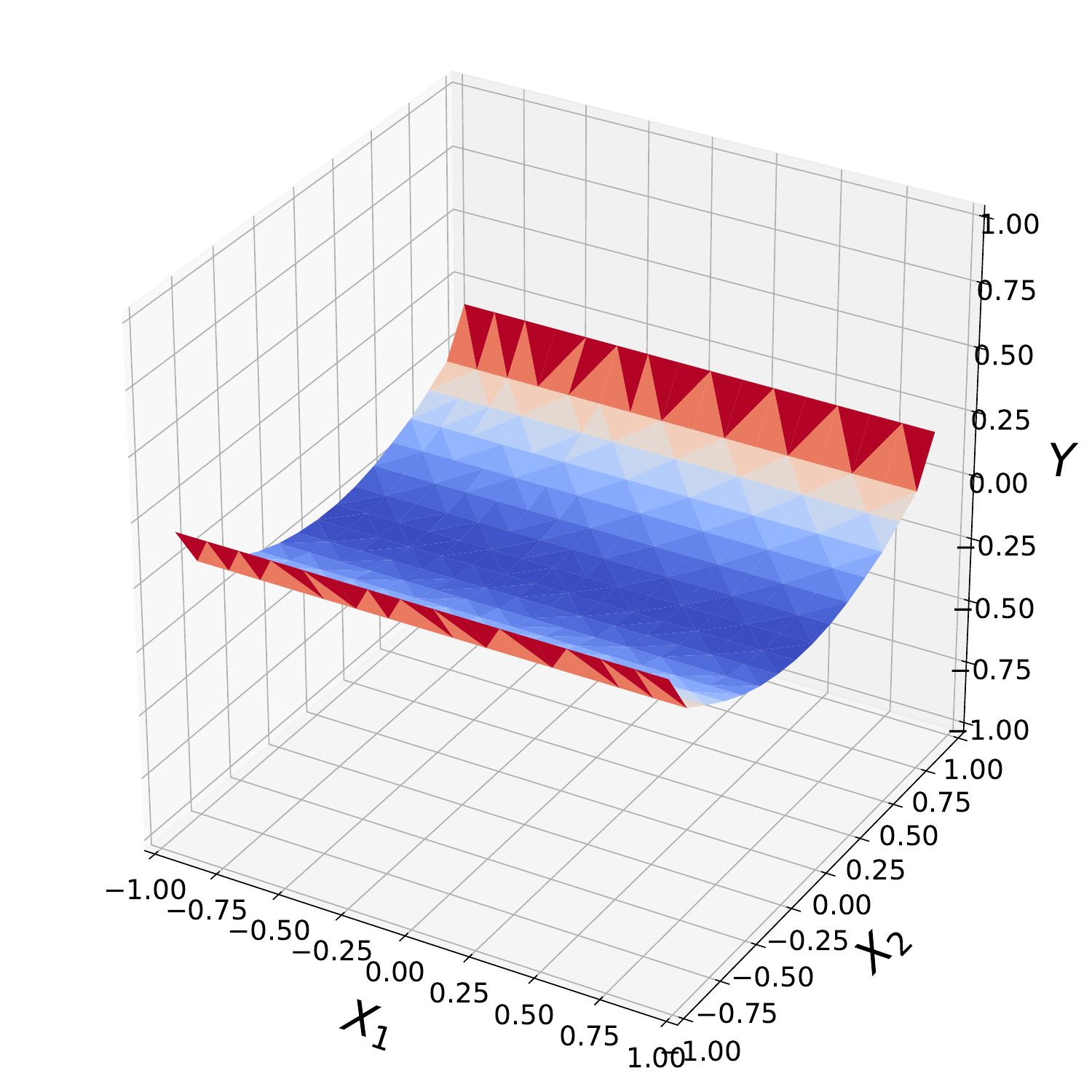}
        \end{minipage}\
        \caption{$\rho = 0.99$ \label{fig:toy_intx:rho1}}
    \end{subfigure}
    \caption{A toy example of decomposing a function into pure interaction and main effects. In each (\subref{fig:toy_intx:rho0}) and (\subref{fig:toy_intx:rho1}), there are four panes: (left) an overall function, (middle left) a pure interaction effect of $X_1$ and $X_2$, (middle right) a pure effect of $X_1$, and (right) a pure effect of $X_2$. 
    In both \subref{fig:toy_intx:rho0} and \subref{fig:toy_intx:rho1}, the overall function is $Y = X_1X_2$, but the decomposition changes based on the coefficient $\rho$ of correlation between $X_1$ and $X_2$. 
    For $X_1$ and $X_2$ uncorrelated, the multiplication is a pure interaction effect; for $X_1$ and $X_2$ correlated, much of the variance can be moved into effects of the individual variables. 
    The decomposition is unique given the joint distribution of the three variables.
    \label{fig:toy_intx}}
\end{figure*}
An example of the distribution changing the meaning of a pure interaction effect is shown in Fig.~\ref{fig:toy_intx}.

\subsection{The Unreasonable Effectiveness of Models with Few Interaction Effects}
\label{sec:intx:unreasonable}
Generalized additive models (GAMs) \cite{hastie1990generalized} are a restrictive model class which estimate functions of individual features, i.e., functions of the form $f(X_i,\ldots, X_p) = \sum_{i=1}^p g_i(X_i)$. 
There have been a large number of methods for estimating these functions, including functional forms such as splines, trees, wavelets, etc. \citep{eilers1996flexible,lou2012intelligible, wand2011penalized}. 
While vanilla GAMs describe nonlinear relationships between each feature and the label, interactions are sometimes added to further capture relationships between multiple features and the label \citep{coull2001simple,lou2013accurate,tay2019reluctant}. 

In the age of deep learning, it is surprising that GAMs with a small number of added interaction effects could be state-of-the-art on any dataset with a moderately large number of samples. 
However, successful tree-based ensembles such as XGBoost \citep{Chen2016XGBoostAS} often require only a few interaction effects to win competitions \citep{nielsen2016tree}.  
In certain cases, polynomial regression of order 2 can be competitive with fully-connected deep NNs \citep{cheng2018polynomial}, and even generalized additive models have a surprising capability to approximate deep NNs \citep{tan2018learning}. 
Similar phenomena have been observed for Gaussian Processes \citep{delbridge2019randomly} and computer vision models \citep{yin2019fourier,wang2019high,tsuzuku2019structural}.  
How are these models, which ignore the majority of interaction effects, so effective? 

\subsection{Statistical (Un)Reliability of Interaction Effects}
\label{sec:intx:statistical}
One reason why models which ignore high-order interaction effects can perform so well is the tremendous difficulty that higher-order interaction effects present to learning algorithms. 
When trying to learn high-order interaction effects, we are stuck between a rock and a hard place: the number of possible interaction effects grows exponentially (the number of $k$-order interaction effects possible from $N$ input features is $\binom{N}{k}$, while the the variance of an interaction effect grows with the interaction order \citep{leon2009sample}. 
This quandry is intensified when the effect strength decreases with interaction order, which is reasonable for real data \citep{gelman_2018}. 
It is like searching for a needle in a haystack, but as we increase $k$, the haystack gets larger and the needle gets smaller. 
For large $k$, we are increasingly likely to select spurious effects rather than the true effect -- at some point it is better to stop searching the haystack.
Viewed this way, it is less surprising that in the absence of prior knowledge of which interaction effects are true, simple models are able to outperform large models.

\subsection{Parity and Interaction Effects}

Interaction effects are intricately linked to a classically difficult function class: parity. 
In the case of two Boolean variables, a pure interaction effect is exactly a weighted XOR function and for continuous variables, pure interaction effects are a continuous analog of parity \citep{lengerich2019purifying}. 
Parity functions are notoriously difficult to learn with NNs \citep{wilamowski2003solving,selsam2018learning}. 
Does this suggest that NNs are already robust against interaction effects, and if so, why is the extra regularization of Dropout against interaction effects necessary?

It is important for us to distinguish between learning the \emph{correct} interaction effect against learning a \emph{spurious} interaction. 
Given $N$ variables, there are $O(N)$ possible main effects, $O(N^2)$ possible pairwise interactions, $O(N^3)$ possible 3-way interactions, $O(N^4)$ possible 4-way interactions, etc. 
This exponential growth in the hypothesis space of interaction terms simultaneously increases the probability that a universal approximator would estimate \emph{some} interaction effect while decreasing the probability that the same universal approximator selects the \emph{correct} interaction effect. 
For this reason, it can be possible for model classes to struggle with accurate recovery of parity functions without being inherently biased against high-order interactions. 
As shown in Figure~\ref{fig:hypo}, the exponential growth in the number of potential interaction terms is balanced by the exponential decay in learning rate induced by Dropout.
In this way, large NNs trained with Dropout can have the convenient property that they are \emph{capable} of learning high-order interactions but will put off the difficult task of learning these high-order interactions until simpler functions have been thoroughly explored.

\section{Analysis}
\subsection{Proof of Theorem 1}
\begin{proof}
Let $\E[Y|X] = F(X) = \sum_{u \in [d]}f_u(X_u)$ and $\tilde{Y} = F(X \odot M_p)$, where $M_p \sim Bernoulli(p)$ is the Input Dropout mask and $\odot$ is element-wise multiplication. Then

\begin{subequations}
\begin{align}
    \E_M[\tilde{Y}|X] &= \sum_{u \in [d]} P(X \odot M = X)f_u(X_u) + \big(1 - P(X\odot M = X)\big)\E_M[f_u(X_u \odot M^+_u)] \\
    &= \sum_{u \in [d]}(1-p)^{|u|}f_u(X_u) + \big(1 - (1-p)^{|u|}\big)\E_M[f_u(X_u \odot M^+_u)] \\
    &= \sum_{u \in [d]}(1-p)^{|u|}f_u(X_u) + \big(1 - (1-p)^{|u|}\big)\E_{v \in u}[f_u(X_{u\backslash v}, X_v=0)] \\
    &= \sum_{u \in [d]}(1-p)^{|u|}f_u(X_u) + a_u(X_u) %f_u(X_{u\backslash v}, X_v=0)] \\
    %&= \sum_{u \in [d]}(1-p)^{|u|}f_u(X_u) + \big(1 - (1-p)^{|u|}\big)\int f_u(X_{u\backslash v}, X_v=0)dX_v \\
    %&= \sum_{u \in [d]}(1-p)^{|u|}f_u(X_u) 
\end{align}
\end{subequations}
where $M^+$ is drawn uniformly from the Dropout masks with at least one zero value and 
$a_u(X_u) = \sum_{v \subseteq u}p^{|u|-|v|}(1-p)^{|v|}f_U(X_{u\backslash v}, X_v=0)$. 
Further,
\begin{subequations}
\begin{align}
    \E_{X_u}[a_u(X_u)] &= \E_{X_u}[\sum_{v \subseteq u}p^{|u|-|v|}(1-p)^{|v|}f_U(X_{u\backslash v}, X_v=0)] \\
    &= \sum_{v \subseteq u}p^{|u|-|v|}(1-p)^{|v|}\mathbb{E}_{X_u}[f_u(X_{u\backslash v}, X_v=0)] \\
    &= 0
    % \E_{X_u}[a_u(X_u)] &= \E_{X_u}[2^{-|u|}\sum_{v \in u}f_u(X_{u \backslash v}, X_v=0)] \\
    % &= 2^{-|u|}\sum_{v \in u}\E_{X_u}[f_u(X_{u \backslash v}, X_v=0)] \\
    % &= 0
\end{align}
\end{subequations}
where the final equality holds by orthogonality of the fANOVA. 
\end{proof}

\subsection{Proof of Corollary 1.1}
\begin{proof}
Let $f_u(X_u)$ be a multiplicative effect. Then
\begin{align}
    f_u(X_{u\backslash v}, X_v=0)=0 \quad \forall~v~\in~u
\end{align}
and hence $a_u(X_u) = 0~\forall~u$.
\end{proof}

\subsection{Proof of Corollary 1.2}
\begin{proof}
Let $F(X) = G(X, Y, \theta)$ for fixed $Y, \theta$. Then $\E[F(X \odot M_p)] = \sum_{u \in [d]}(1-p)^{|u|}f_u(X) + a_u(X_u)$ by Theorem 1.
\end{proof}

\subsection{Proof of Corollary 1.3}
\begin{proof}
Let $F(A_i) = G_i(A_i, Y, \theta)$ for fixed $Y, \theta$. Then $\E[F(A_i \odot M_p)] = \sum_{u \in [d]}(1-p)^{|u|}f_u(X) + a_u(X_u)$ by Theorem 1.
\end{proof}

\section{Additional Experimental Details}
\label{sec:additional_experiments}

\begin{figure*}[tp]
    \centering
    \begin{subfigure}[t]{0.3\textwidth}
        \centering
        \includegraphics[width=\columnwidth]{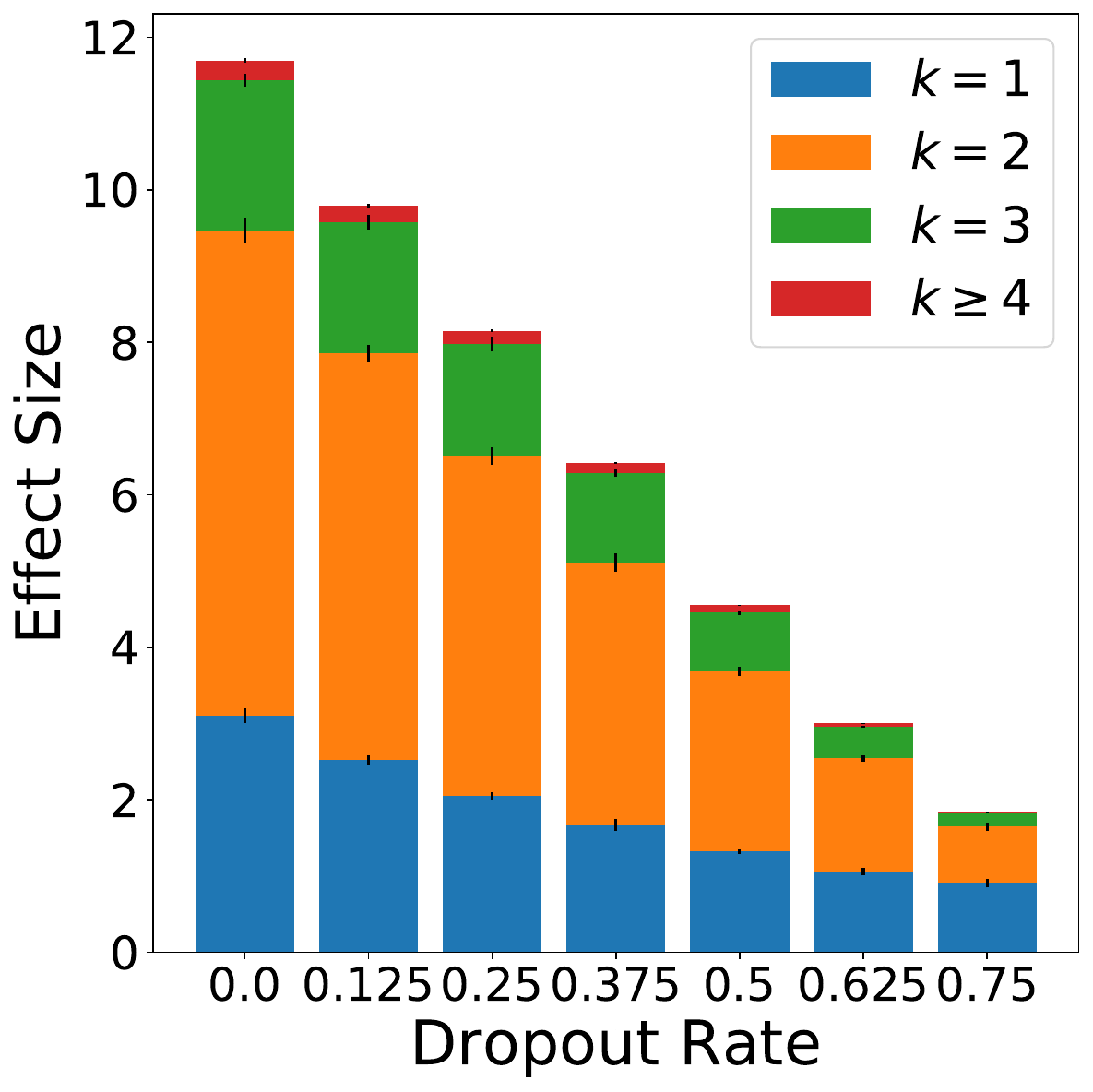}
        \caption{Total Effect: Activation Dropout 
        \label{fig:converged_128:total_activation}}
    \end{subfigure}
    ~
    \begin{subfigure}[t]{0.3\textwidth}
        \centering
        \includegraphics[width=\columnwidth]{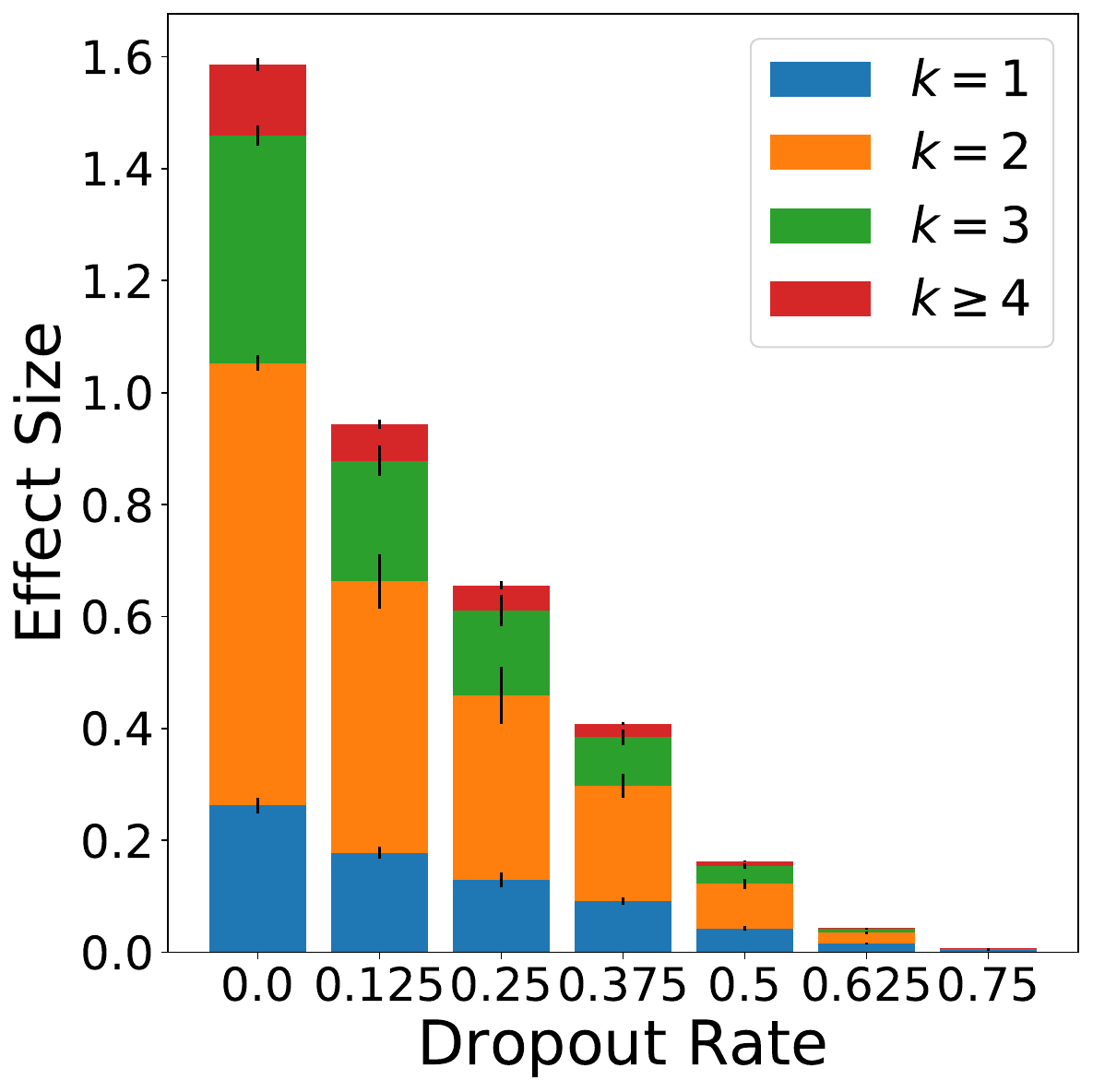}
        \caption{Total Effect: Input Dropout
        \label{fig:converged_128:total_input}}
    \end{subfigure}
    ~
    \begin{subfigure}[t]{0.3\textwidth}
        \centering
        \includegraphics[width=\columnwidth]{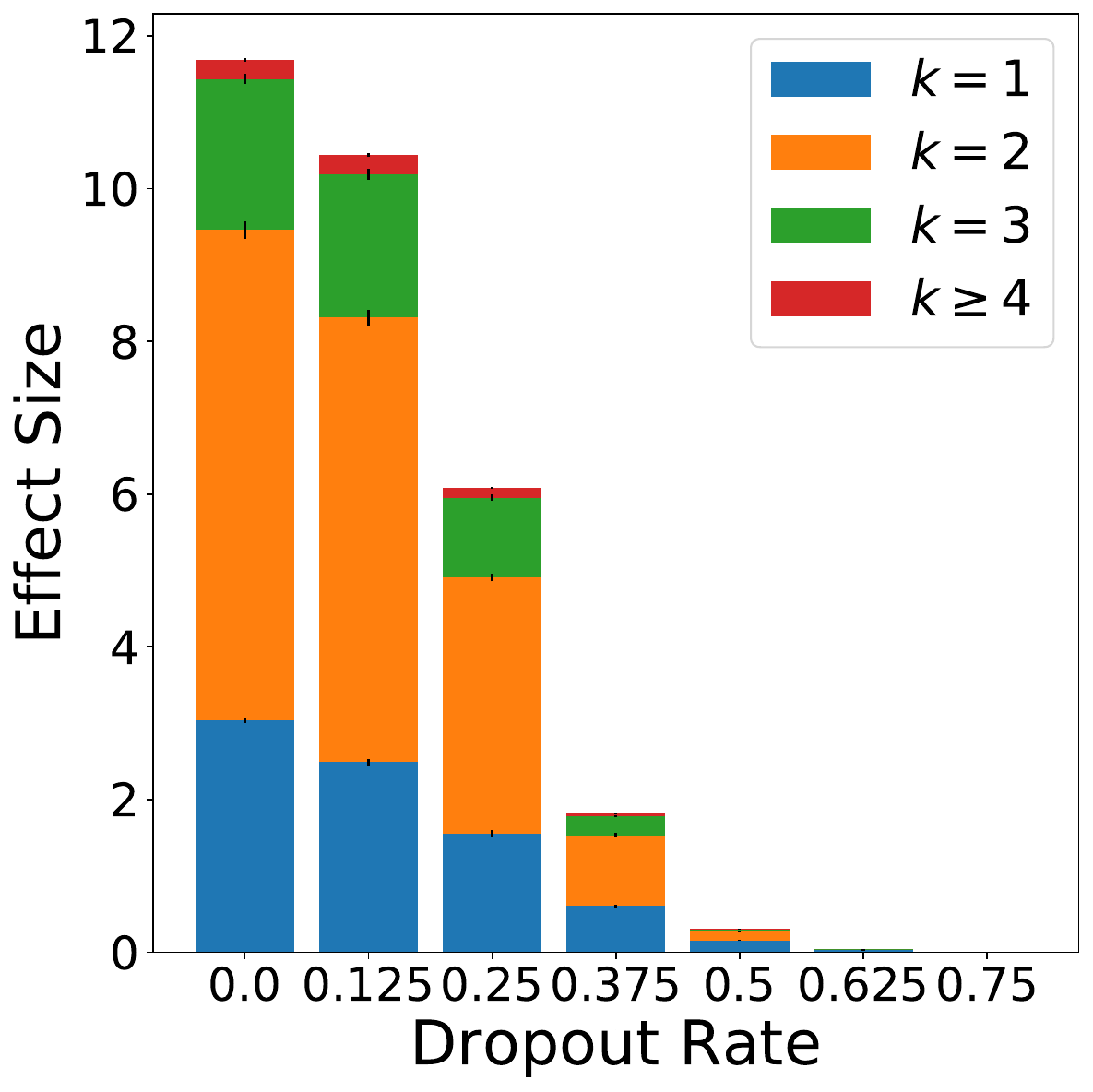}
        \caption{Total Effect: Input + Activation Dropout
        \label{fig:converged_128:total_both}}
    \end{subfigure}
    \\
    \begin{subfigure}[t]{0.3\textwidth}
    \centering
        \includegraphics[width=\columnwidth]{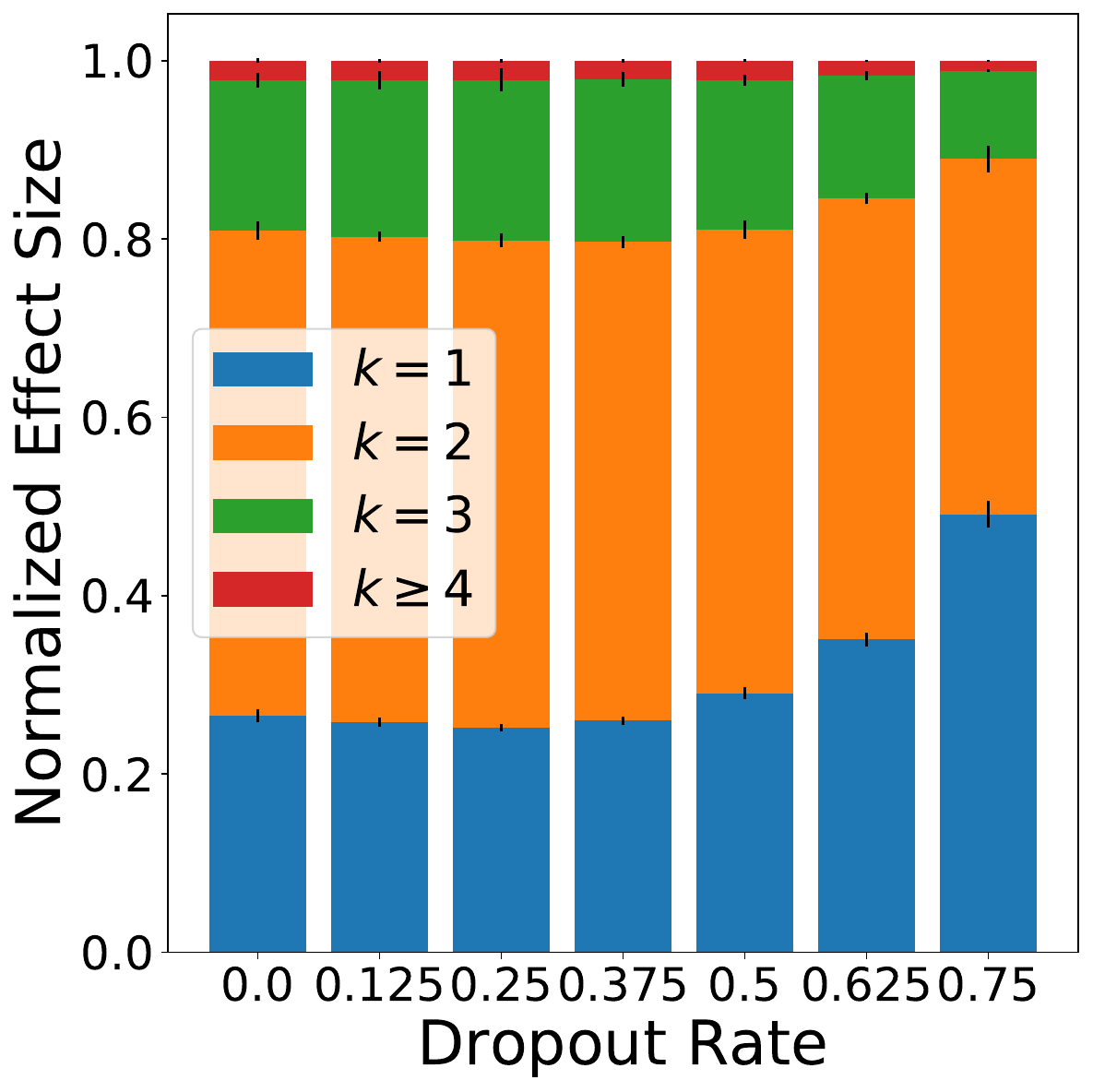}
        \caption{Normalized Effect: Activation Dropout
        \label{fig:converged_128:norm_activation}}
    \end{subfigure}
    ~
    \begin{subfigure}[t]{0.3\textwidth}
    \centering
        \includegraphics[width=\columnwidth]{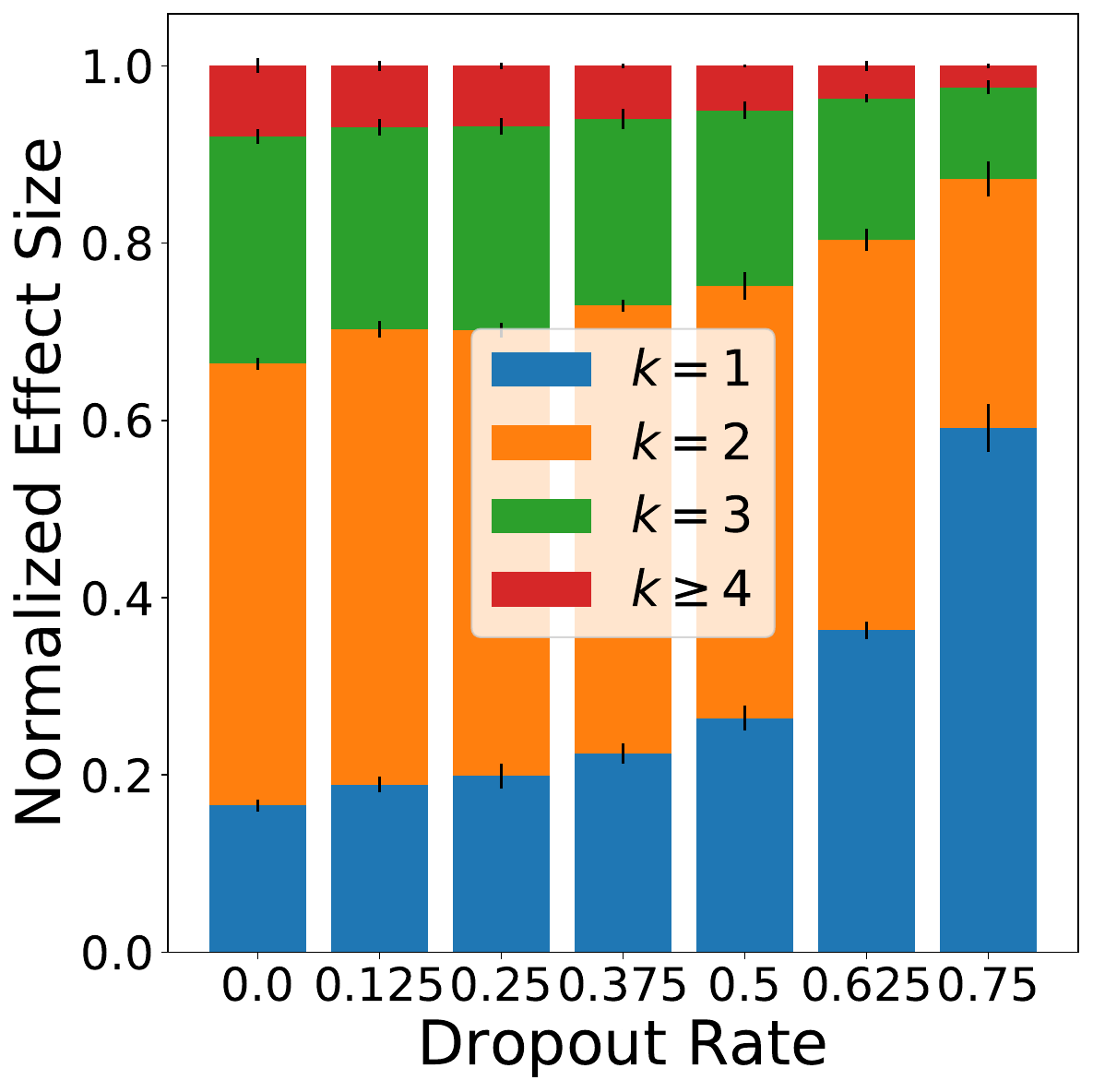}
        \caption{Normalized Effect: Input Dropout
        \label{fig:converged_128:norm_input}}
    \end{subfigure}
    ~
    \begin{subfigure}[t]{0.3\textwidth}
    \centering
        \includegraphics[width=\columnwidth]{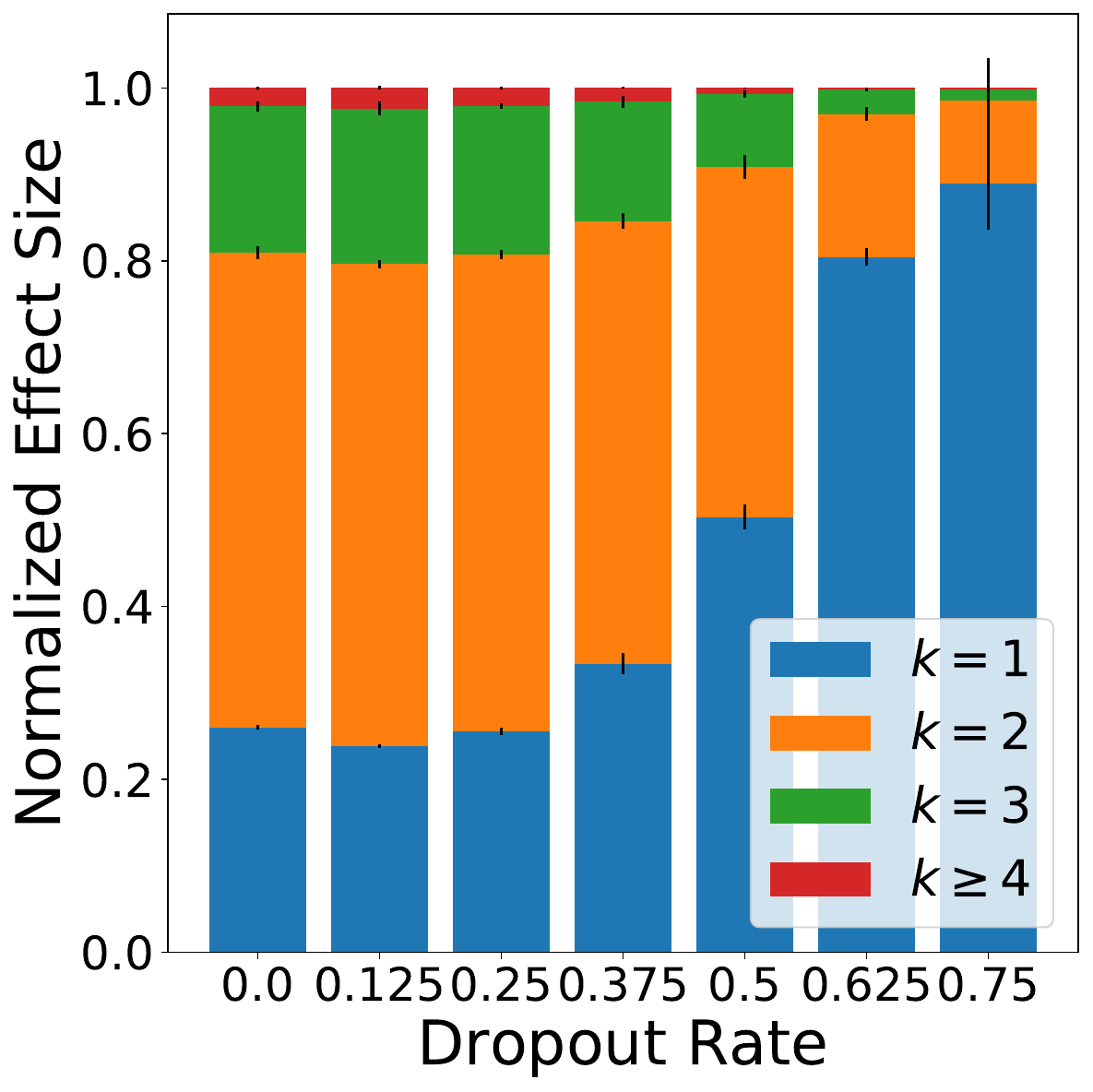}
        \caption{Normalized Effect: Input + Activation
        \label{fig:converged_128:norm_both}}
    \end{subfigure}
    \\
    \begin{subfigure}[t]{0.3\textwidth}
        \centering
        \includegraphics[width=\columnwidth]{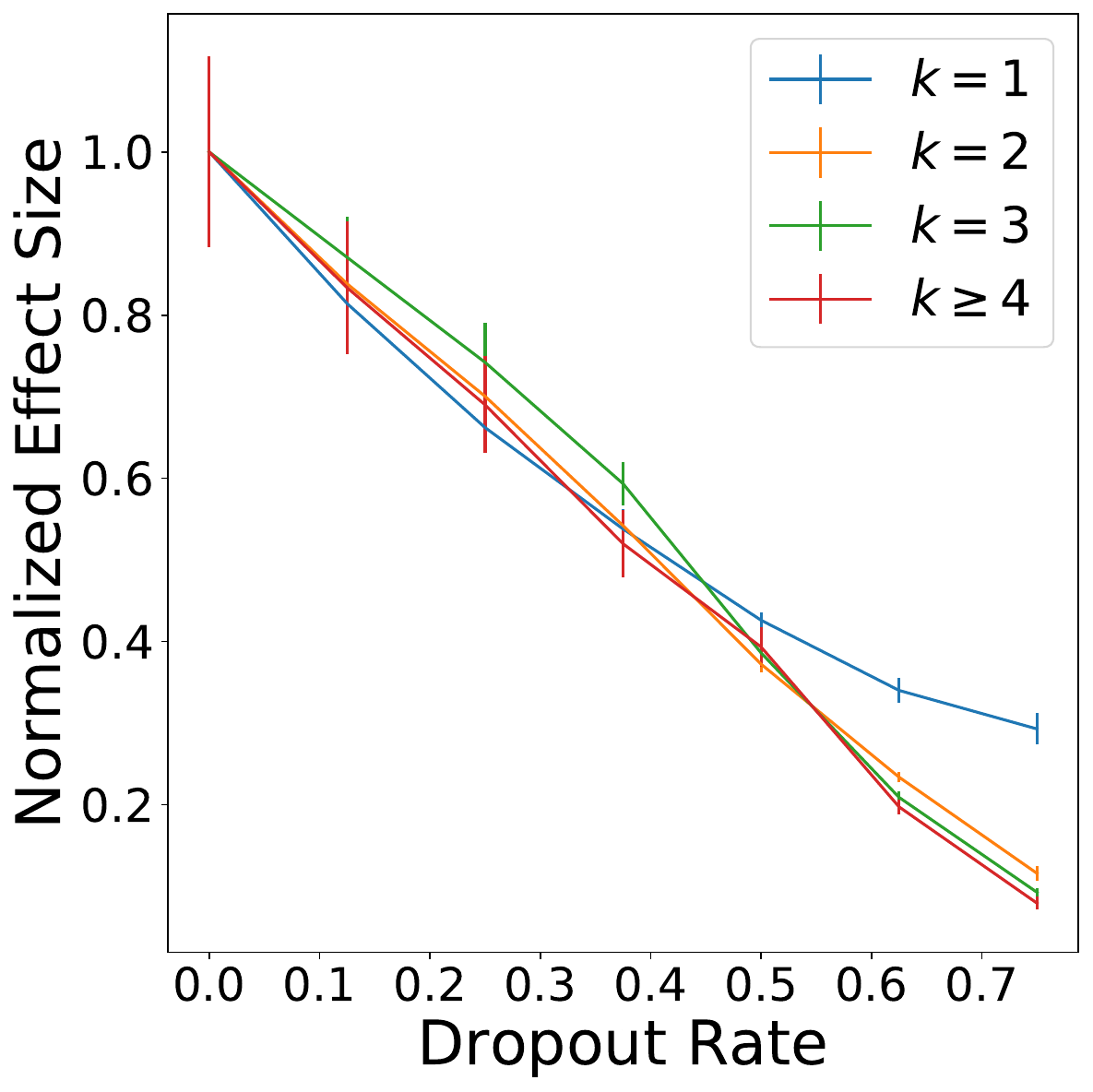}
        \caption{Shrinkage: Activation Dropout
        \label{fig:converged_128:decay_activation}}
    \end{subfigure}
    ~
    \begin{subfigure}[t]{0.3\textwidth}
        \centering
        \includegraphics[width=\columnwidth]{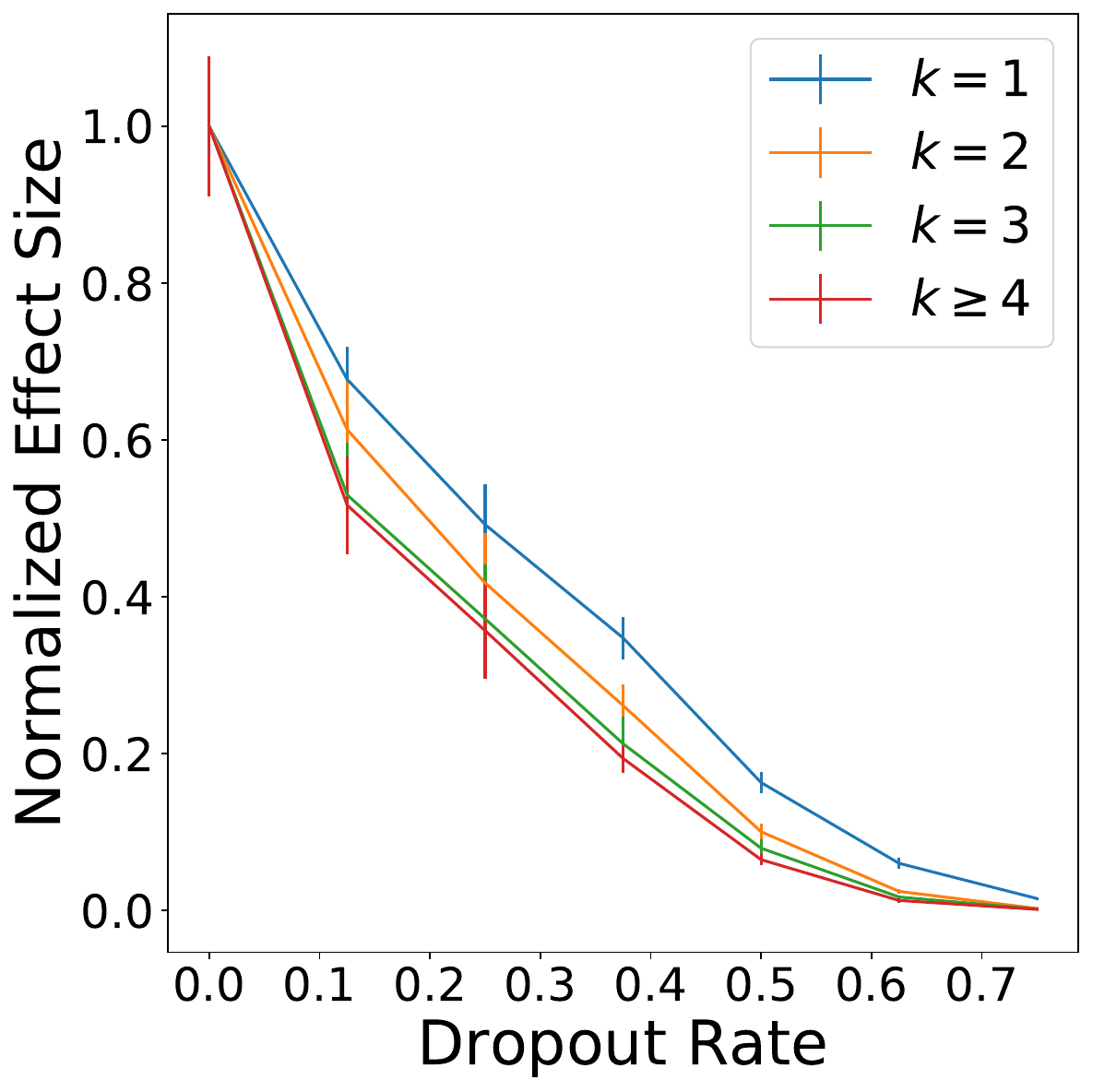}
        \caption{Shrinkage: Input Dropout
        \label{fig:converged_128:decay_input}}
    \end{subfigure}
    ~
    \begin{subfigure}[t]{0.3\textwidth}
        \centering
        \includegraphics[width=\columnwidth]{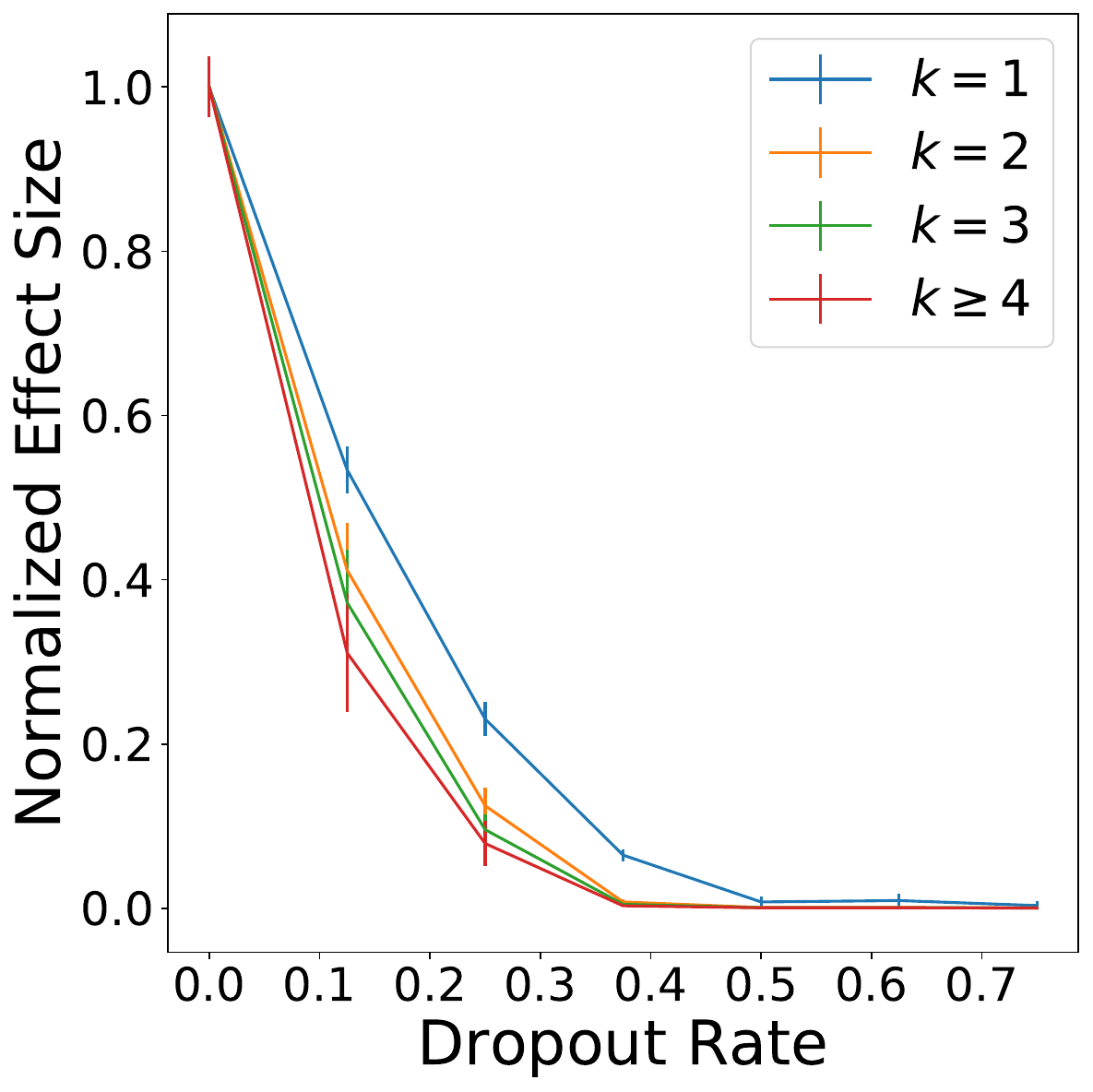}
        \caption{Shrinkage: Input + Activation Dropout
        \label{fig:converged_128:decay_both}}
    \end{subfigure}
    \caption{In this experiment, we train fully-connected neural networks on a dataset of pure noise (details in Sec.~\ref{sec:noise_expt}). 
    Displayed values are the (mean $\pm$ std. over 10 initializations) of the proportion of the trained model's variance explained by each order of interaction effect. 
    All neural networks in this figure have 128 units in each hidden layer (compared to 32 units per layer in Figure~\ref{fig:converged_32}), and we see that Activation Dropout has only a small impact, while Input Dropout significantly reduces the estimated effect sizes of the high-order interactions. As expected, increasing the size of the hidden layers from 32 in Figure~\ref{fig:converged_32} to 128 in this Figure decreases the impact of Activation Dropout on high-order interactions, but does not reduce the effectiveness of Input Dropout. 
    \label{fig:converged_128}}
\end{figure*}

\paragraph{Pure Noise Data}
Figure~\ref{fig:converged_128} shows the results of various Dropout rates on a NN with 128 hidden units in each layer. 
These results are analogous to the results shown in Fig.~\ref{fig:converged_32} of the main text for a NN with 32 hidden units in each layer.

\paragraph{Modified 20-NewsGroups}
Table~\ref{tab:newsgroups} displays the results of various Dropout Rates on the Modified 20-NewsGroups datasets described in Section~\ref{sec:experiments:optimal_rate}.

\begin{table}[htb]
    \centering
    \begin{tabular}{c|c|c|c|c|c|c}
        $k$ & \multicolumn{6}{c}{Dropout Rate} \\
        &  $0.0$ & $0.125$ & $0.25$ & $0.375$ & $0.5$ & $0.625$ \\
        \midrule
        %None & \\
        1 & $0.52 \pm 0.01$ & $0.54 \pm 0.01$ & $0.54 \pm 0.03$ & $\bf{0.57 \pm 0.02}$ & $0.55 \pm 0.02$ & $0.47 \pm 0.02$\\
        2 & $0.39 \pm 0.01$ & $0.38 \pm 0.03$ & $\bf{0.40 \pm 0.02}$ & $\bf{0.40 \pm 0.01}$ & $0.38 \pm 0.01$ & $0.27 \pm 0.02$\\
        3 & $0.39 \pm 0.01$ & $\bf{0.41 \pm 0.01}$ & $\bf{0.41 \pm 0.01}$ & $0.40 \pm 0.02$ & $0.40 \pm 0.02$ & $0.27 \pm 0.04$ \\
        %4 & $0.33 \pm 0.02$ & $0.36 \pm 0.01$ & $0.38 \pm 0.02$ & $0.37 \pm 0.02$ & $0.31 \pm 0.02$ & $0.24 \pm 0.03$\\
    \end{tabular}
    \caption{Test accuracies of the models trained on the modified 20-Newgroups datasets (Sec.~\ref{sec:experiments:optimal_rate}). Reported values are (mean $\pm$ std) of the test accuracies over 5 experiments, with the best setting in each row bolded. Each row indicates $k$, the order of the added interaction effect. As $k$ is increased, lower levels of Dropout tend to outperform. Different modifications of the dataset change the difficulty of the task, so the accuracy values are not comparable across rows.}
    \label{tab:newsgroups}
\end{table}

\begin{figure*}[t]
    \centering
    \begin{subfigure}[t]{0.23\textwidth}
        \centering
        \includegraphics[width=\textwidth]{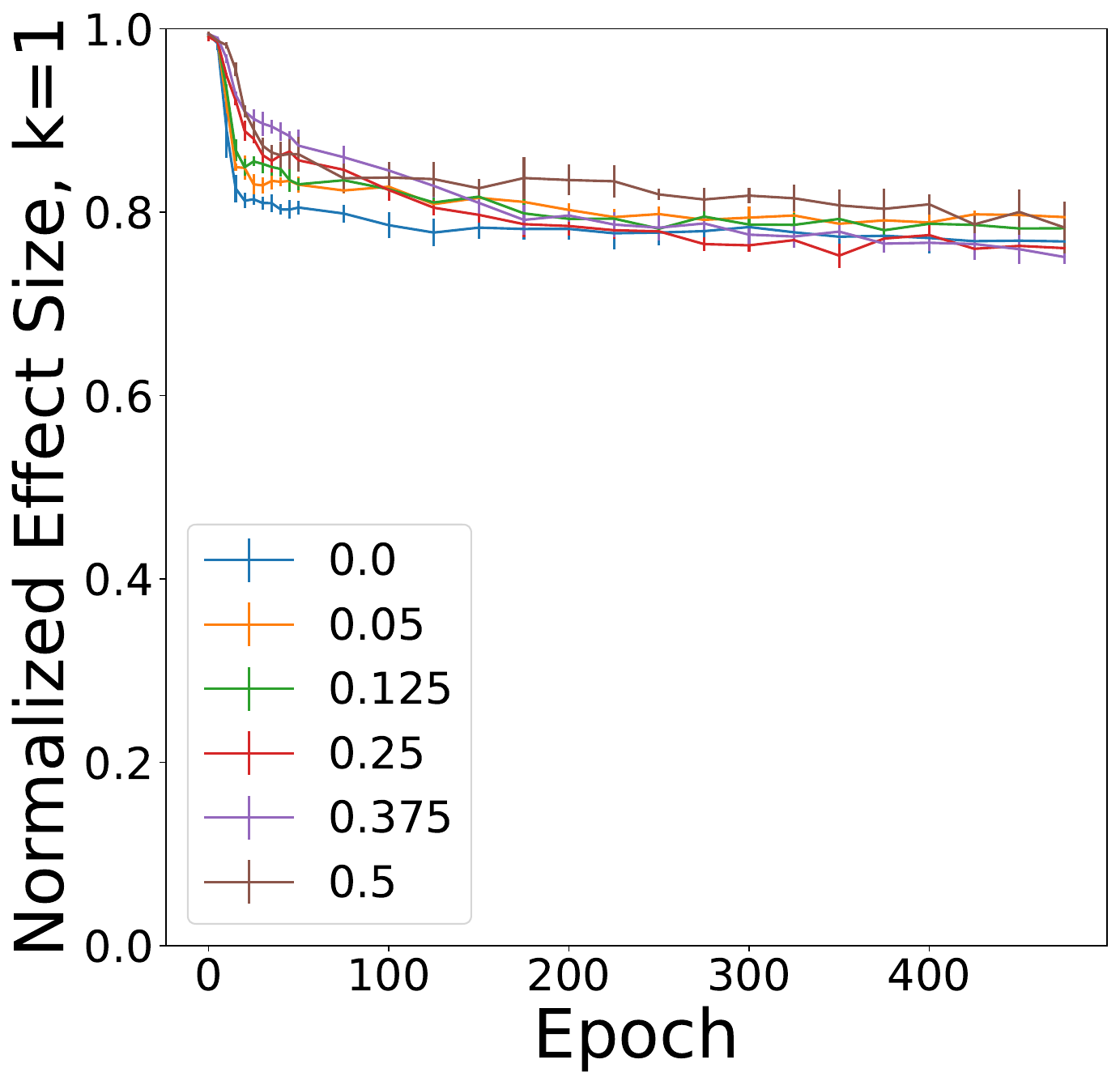}
        \caption{1-way interactions}
    \end{subfigure}
    ~
    \begin{subfigure}[t]{0.23\textwidth}
        \centering
        \includegraphics[width=\textwidth]{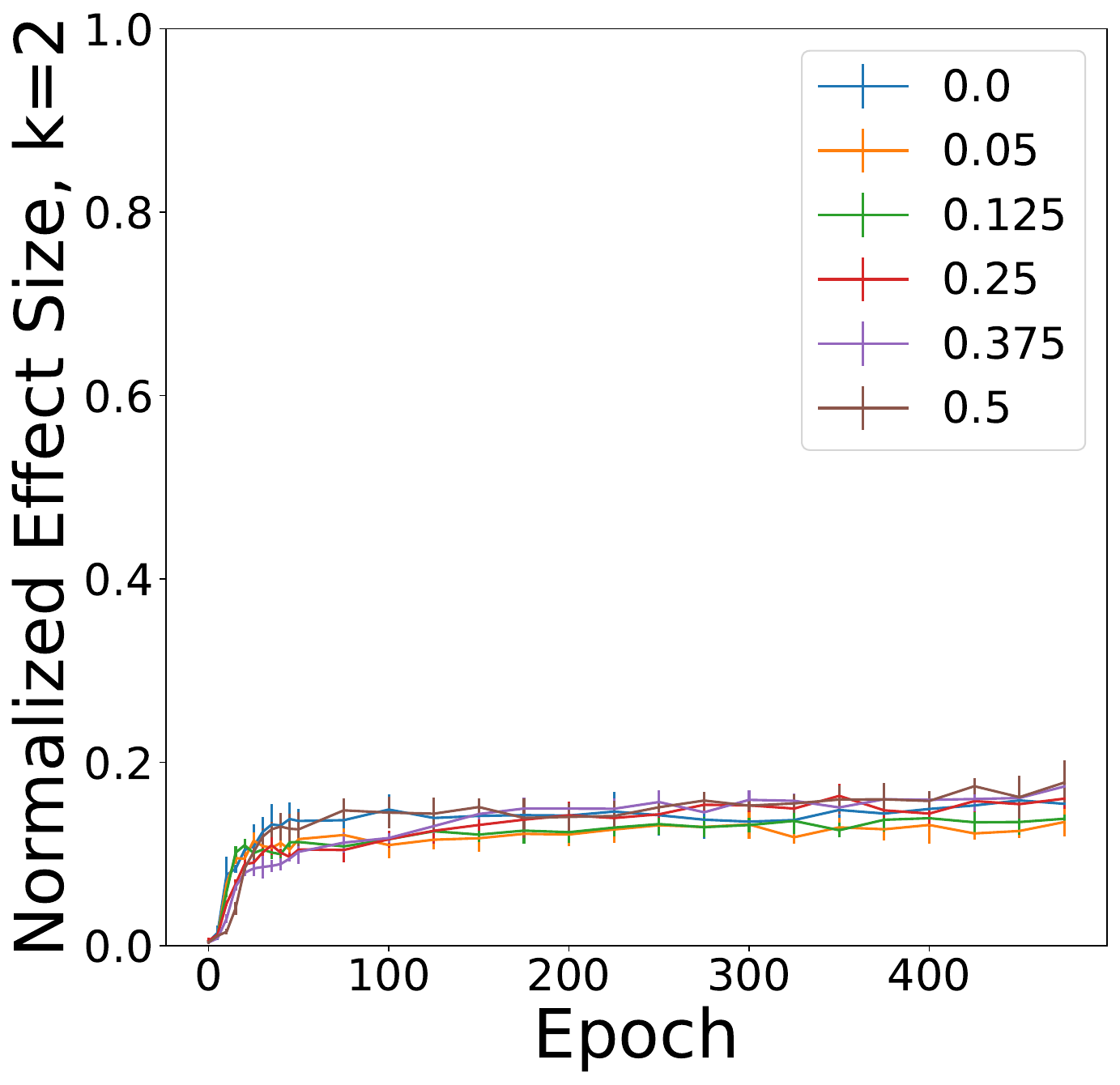}
        \caption{2-way interactions}
    \end{subfigure}
    ~
    \begin{subfigure}[t]{0.23\textwidth}
        \centering
        \includegraphics[width=\textwidth]{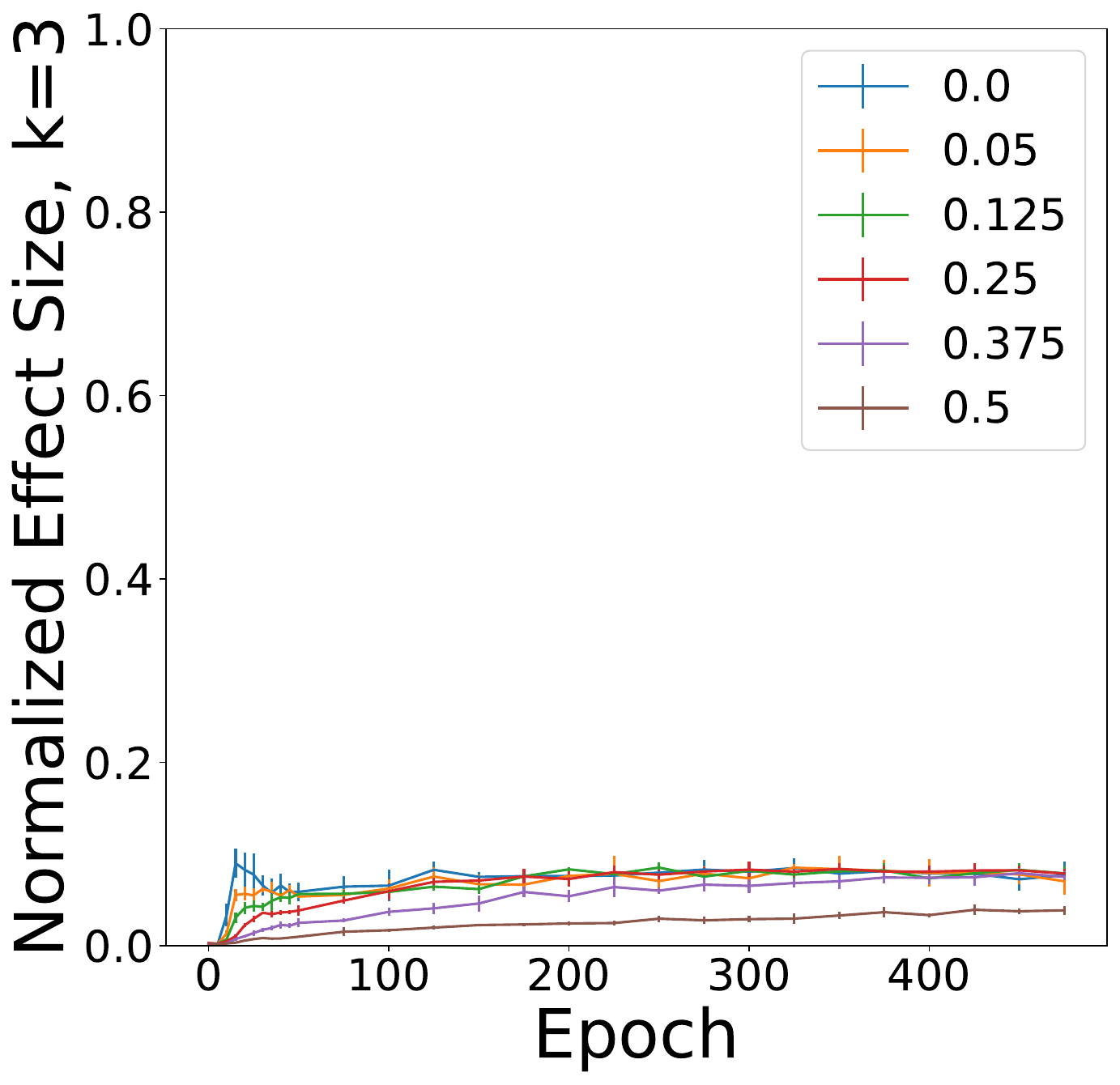}
        \caption{3-way interactions}
    \end{subfigure}
    ~
    \begin{subfigure}[t]{0.23\textwidth}
        \centering
        \includegraphics[width=\textwidth]{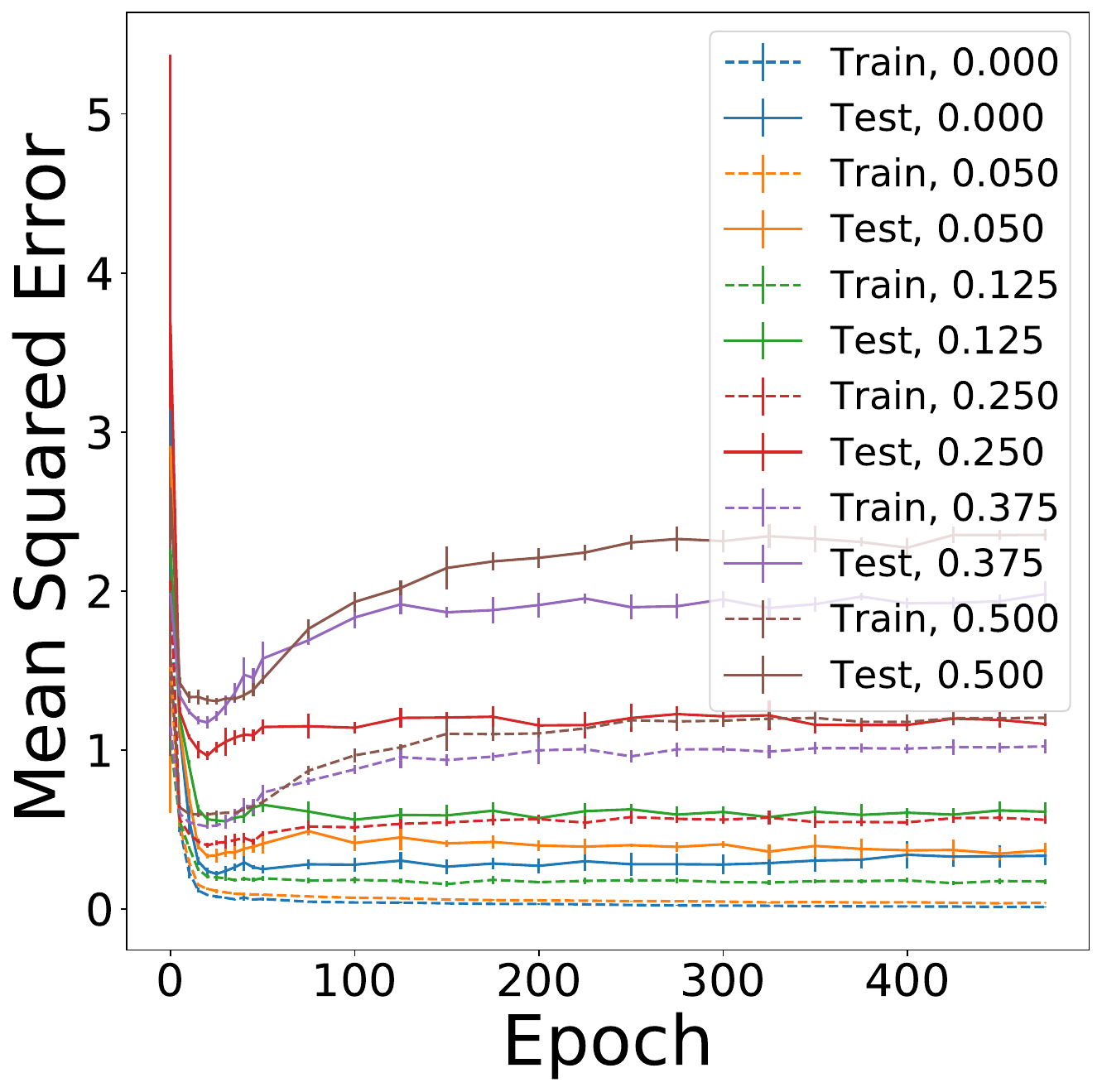}
        \caption{MSEs}
    \end{subfigure}
    \caption{Learned interaction effects and model errors over epochs training on the BikeShare Dataset. In this dataset, there are true interaction effects of orders 2 and 3, so the models with high Dropout rates generalize \emph{worse} than the models with low Dropout rates. This behavior is expected under our perspective of Dropout as an interaction regularizer, but unexpected under the perspective of Dropout as a generic model regularizer. 
    \label{fig:dropout_bikeshare}}
\end{figure*}

\paragraph{BikeShare}
Figure~\ref{fig:dropout_bikeshare} displays results of various Dropout rates on a NN trained on the New York City Bikeshare dataset. 
Because this dataset contains real interaction effects \citep{tan2018learning}, the optimal Dropout rate for generalizing to the test set is actually 0.

\end{document}